\documentclass{article}

\usepackage[round,authoryear,sort]{natbib}  
\usepackage{enumitem}
\usepackage{amsfonts}
\usepackage{amssymb}
\usepackage{amsmath}

\usepackage{ragged2e}  
\usepackage{microtype}
\usepackage{graphicx}
\usepackage{subcaption}
\usepackage{adjustbox}
\usepackage{caption}
\usepackage{booktabs} 

\usepackage{multirow}
\usepackage{array}
\usepackage{makecell}

\usepackage[preprint]{icml2026}

\usepackage[dvipsnames]{xcolor}
\usepackage{etoc}
\newcolumntype{C}[1]{>{\centering\arraybackslash}p{#1}}

\usepackage{hyperref}
\usepackage[capitalize,noabbrev]{cleveref}
\hypersetup{
  colorlinks,
  citecolor=BlueViolet,  
  linkcolor=Bittersweet,     
  urlcolor=cyan}
\usepackage{etoolbox}

\usepackage[textsize=tiny]{todonotes}

\usepackage{tabularx,threeparttable,multirow}
\newcolumntype{Y}{>{\raggedright\arraybackslash}X}
\usepackage{enumitem}
\newenvironment{compactitem}{
  \begin{itemize}[itemsep=3pt,leftmargin=20pt]
}{
  \end{itemize}
}

\icmltitlerunning{Active Learning with Task--Driven Representations for Messy Pools}

\begin{document}
\twocolumn[
  \icmltitle{Active Learning with Task--Driven Representations for Messy Pools}

  \icmlsetsymbol{equal}{*}
  \begin{icmlauthorlist}
    \icmlauthor{Kianoosh Ashouritaklimi}{yyy}
    \icmlauthor{Tom Rainforth}{yyy}
  \end{icmlauthorlist}

  \icmlaffiliation{yyy}{Department of Statistics, University of Oxford}
  \icmlcorrespondingauthor{Kianoosh Ashouritaklimi}{kianoosh.ashouritaklimi@stats.ox.ac.uk}
  \icmlkeywords{Machine Learning, ICML}
  \vskip 0.3in
]

\printAffiliationsAndNotice{}  

\begin{abstract}
Active learning has the potential to be especially useful for messy, uncurated pools where data points vary in relevance to the target task. 
However, state-of-the-art approaches to this problem currently rely on using fixed, unsupervised representations of the pool, focusing on modifying the acquisition function instead. 
We show that this \emph{model} setup can undermine their effectiveness at dealing with messy pools, as such representations can fail to capture important information relevant to the task. 
To address this, we propose using \emph{task--driven representations} that are periodically updated during the active learning process using the previously collected labels. We introduce two specific strategies for learning these representations, one based on directly learning semi--supervised representations and the other based on supervised fine--tuning of an initial unsupervised representation.
We find that both significantly improve empirical performance over using unsupervised or pretrained representations.
\end{abstract}

\section{Introduction}
\label{sec:intro}

Active learning~\citep{mackayALInformation,confSampling} is a framework for selecting the best datapoints to label during the training of a predictive model. It has the potential to be especially useful in the setting of messy, uncurated pools commonly encountered with real--world data, where the unlabelled data points have widely varying relevance to our task of interest~\citep{epig,semi_epig,messyPool3, ClusterMargin}.
For example, our pool may contain many examples of classes that we are not interested in predicting or the subtle variations we are trying to pick up on may be dwarfed by irrelevant features. 

 \looseness=-1
Unfortunately, such messiness can undermine existing active learning pipelines.
In particular, simple predictive uncertainty measures will often be highest for irrelevant points in the pool~\citep{epig}.
Previous work has tried to address this by developing acquisition functions that are more robust to forms of messiness like class imbalance and irrelevant data points~\citep{GALAXY, SIMILAR, zhang2023algorithm, nuggehalli2023direct, deepOpenActiveLearning}. 

 \looseness=-1
The success of active learning methods, though, is critically dependent not only on our acquisition strategy, but our model choice as well. 
In particular, there is a growing body of evidence that it is imperative to use \emph{semi--supervised} models to incorporate the rich information available in the unlabelled data, for the sake of both direct prediction and guiding acquisition~\citep{SSLAL2, TypiClust, SSLAL3, SSLAL5, sener2017active, yehuda2022active, luth2023navigating, chan2021marginal, bhatnagar2020pal, ebrahimi2020minimax, gao2020consistency}.
By contrast, the aforementioned approaches for dealing with messy pools have all focused mainly on fully supervised models.
\citet{semi_epig} recently provided a first notable exception to this by showing that their prediction-orientated acquisition strategy can be successfully combined with \emph{unsupervised} representations to yield state-of-the-art active learning performance for messy pools.

In this work, we show that effectively dealing with messy pools requires careful specific considerations in how our model is constructed, not just our acquisition strategy.
In particular, we highlight that the current use of unsupervised representations can itself undermine our ability to effectively deal with messy pools:  the task-agnostic nature of such representations mean that they can fail to capture all the information relevant to our task. As a result, the representations can fail to capture the right notion of similarity between inputs for our task, leading to inaccurate predictive correlations and, ultimately, suboptimal acquisitions \citep{grosseMIG, NeuralTestBED}.

We suggest to address this issue by using \textbf{task--driven representations}. Namely, we argue that updating our representations at regular intervals during the active learning process allows us to guide the representations towards capturing task--relevant information. 
This, in turn, enables our model to better learn the relevant similarities between inputs, improves uncertainty estimation, and ultimately leads to better acquisition decisions.

\looseness=-1
To this end, we introduce two concrete strategies for learning these task--driven representations.
The first is to periodically retrain the representation using a semi--supervised objective based on the original pool data and the labels collected thus far.
Specifically, we build on the CCVAE approach of~\citet{ccvae} to learn representations where a subset of the latents are guided by a downstream classifier to capture information in the labels.
The second is a more lightweight approach where we periodically perform a simple supervised fine--tuning of the original unsupervised representation.
Empirically, we find that both approaches lead to more effective acquisitions and significantly enhance model performance. 

In summary, our contributions are: 
\begin{compactitem}
    \item Showing active learning with unsupervised representations can break down with messy pools
    (\S\ref{sec:shortfals}).

    \item Suggesting the use of task--driven representations by periodically updating our representations throughout the active learning process (\S\ref{approach}).

    \item Introducing two strategies for learning task--driven representations (\S\ref{approach}).

    \item Showing our approaches improve performance compared with current state--of--the--art (\S\ref{experiments}).
\end{compactitem}

\section{Background}
\label{sec:background}
\subsection{Problem Formulation}
\label{background:problem}

Active learning (AL) provides a principled approach to adaptively selecting datapoints to label when training a predictive model. We consider pool--based AL where we have a small initial labelled dataset $\mathcal{D}_l=\{(x^l_i,y^l_i)\}_{i=1}^M$ and a larger unlabelled pool $\mathcal{D}_u=\{(x^u_i)\}_{i=1}^N$, with inputs $x \in \mathcal{X}$ and outputs $y \in \mathcal{Y}$. 
The objective is to iteratively choose (a subset of) points from $\mathcal{D}_u$ for labeling, with the aim of producing the best model with the fewest labels.

Though our approach applies more generally, for simplicity we assume a classification setting and that our model is probabilistic with updatable parameters $\phi$, such that $p_{\phi}(y \vert x) = \mathbb{E}_{p_\phi(\theta)}\left[p_{\phi}(y \vert x, \theta)\right]$, for some stochastic parameters $\theta$.
We will further assume that data is treated to be i.i.d.~conditional on $\theta$ 
such that $p_{\phi}(y_1, y_2 \vert x_1, x_2) = \mathbb{E}_{p_\phi(\theta)}\left[p_{\phi}(y_1 \vert x_1, \theta)p_{\phi}(y_2 \vert x_2, \theta)\right]$. 
\subsection{Active Learning with Messy Pools}
\vspace{-4pt}

In real--world data, the unlabelled data points in our pool can have widely varying relevance to our task of interest. Pools of web--scraped audio, images and text are common examples of this. Active
learning ought to be particularly helpful in dealing with this messiness, by identifying only the most useful inputs to label.
However, it can cause problems for many standard active learning pipelines, as predictive uncertainty is often highest on these irrelevant data points~\citep{epig,semi_epig}.

Previous work in this setting has primarily focused on dealing with such messiness by designing appropriate acquisition functions. Notable works include \textbf{SIMILAR} \citep{SIMILAR} which uses submodular information measures as acquisition functions to deal with pools involving class imbalance and redundant classes; \textbf{GALAXY} \citep{GALAXY} which proposes a graph--based acquisition function that has shown state--of--the--art performance on pools with redundant and imbalanced classes; and~\textbf{DIRECT} \citep{nuggehalli2023direct} which uses a boundary-aware, one-dimensional acquisition strategy to deal with both class imbalance and label noise.

The acquisition strategy we will primarily utilise in this work is \textbf{EPIG} \citep{epig}, which uses a \emph{prediction--oriented} acquisition strategy to deal with imbalanced and redundant classes.
Specifically, EPIG introduces a target input distribution $p_{*}(x_{*})$ and then considers the expected uncertainty reduction~\citep{rainforth2024modern,lindley1956measure} in hypothetical future predictions $y_*|x_*$:
\begin{equation*}
\resizebox{1.\columnwidth}{!}{$\displaystyle
\text{EPIG}(x) \!=\! \mathbb{E}_{p_*(x_*)p_\phi(y \vert x)}\left[\text{H}[p_\phi(y_* \vert x_*)] \!-\! \text{H}[p_{\phi}(y_* \vert x_*, x, y)] \right]
$}
\end{equation*}
\looseness=-1
where $\text{H}$ refers to Shannon entropy \citep{theory_communication}.
By focusing on a particular target prediction task, EPIG allows acquisitions to be focused on data points that will aid downstream performance and hopefully avoid irrelevant points in the pool.

\vspace{-3pt}
\subsection{Semi--Supervised Active Learning}
\label{background:SSL-BAL}
\vspace{-3pt}

\looseness=-1
Previous work on active learning with messy pools has primarily used fully supervised models \citep{SIMILAR, GALAXY, nuggehalli2023direct, epig, deepOpenActiveLearning}. This is in spite of a growing line of work which shows that it is typically imperative to use semi--supervised models for most active learning problems \citep{SSLAL2, TypiClust, SSLAL3, SSLAL5}: by incorporating the rich information available in the unlabelled data, semi--supervised approaches can improve immediate predictive performance and also the quality of uncertainty estimates that drive acquisitions.

\looseness=-1
Recently, \cite{semi_epig} 
observed reliable gains over random acquisition, in messy and non-messy settings,
by first learning unsupervised representations from the unlabelled data and then performing active learning on top of those fixed representations with a fully supervised prediction head. Concretely, they decompose the predictive model $p_\phi(y|x)$ into a fixed deterministic encoder $g: X \rightarrow \mathbb{R}^d$ and a stochastic prediction head $p_\phi(y|z,\theta_h)$, where $z = g(x)$ and $\theta_h \sim p_{\phi}(\theta_h)$. The overall predictive model is then given by
$p_\phi(y|x) = \mathbb{E}_{p_{\phi}(\theta_h)}\left[p_\phi(y|g(x), \theta_h)\right]$. By fixing the encoder while updating only the prediction head between active learning iterations, they are able to leverage large encoders pretrained on the pool that capture much of the information needed for the downstream task in a lower-dimensional latent space \citep{big_self_ssl, SimpleSimCLR}, while using smaller prediction heads that improve the computational efficiency of the active learning and the quality of the updates.
\begin{figure}[t] 
    \centering 
    \begin{subfigure}[b]{0.5\columnwidth}
        \includegraphics[width=\linewidth]{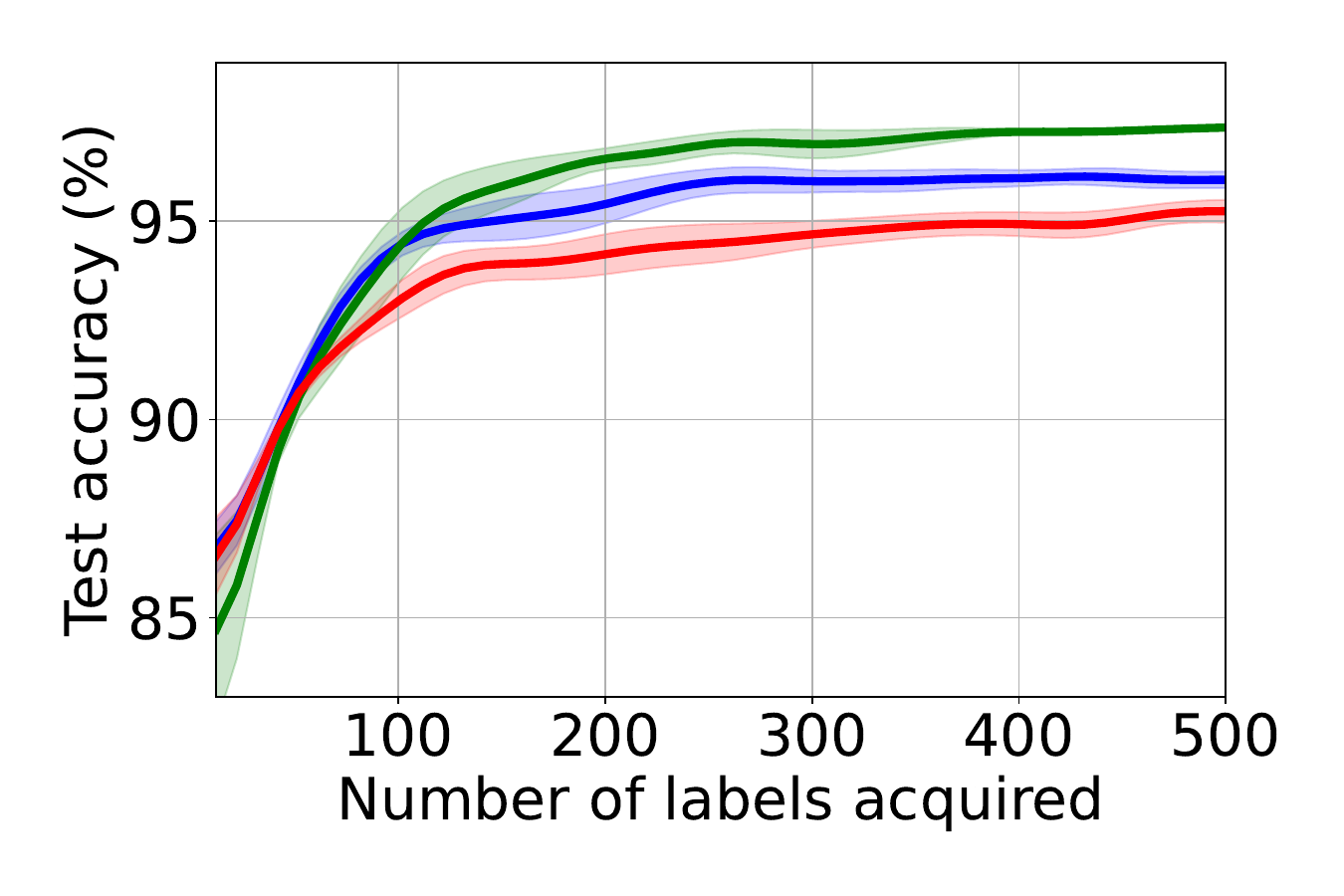}
        \caption{\textbf{F+MNIST}}
        \label{fig:problem_fmnist_epig}
    \end{subfigure}
    \hspace{-0.3cm}    
    \begin{subfigure}[b]{0.5\columnwidth}
        \includegraphics[width=\linewidth]{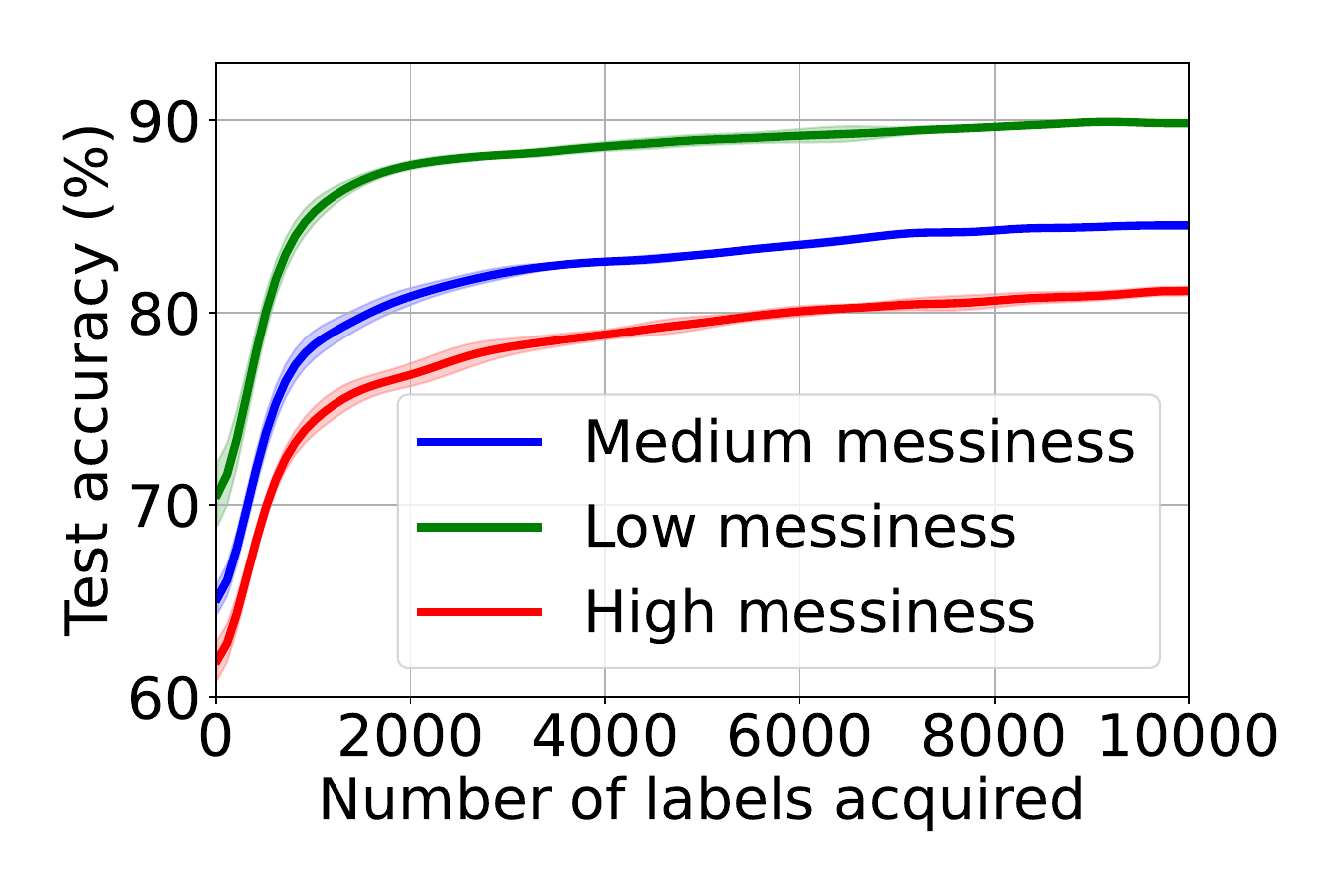}
        \caption{\textbf{CIFAR-10+100}}
        \label{fig:problem_cifar_epig}
    \end{subfigure}
    \caption{Test accuracy for EPIG with unsupervised representations on \textbf{F+MNIST} (left) and \textbf{CIFAR-10+100} (right) (see \S\ref{experiments})
    under increasing levels of pool ``messiness'', namely decreasing the number of pool samples which are of the classes of interest. All experiments were run for 4 seeds, solid line shows mean and shading $\pm 1$ standard error.}
    \label{fig:problem}
\end{figure}

\section{Shortfalls of Unsupervised Representations for Messy Pools}
\label{sec:shortfals}
\looseness=-1
We now explain why dealing with messy pools requires careful consideration of the underlying \emph{model} setup, not just the choice of acquisition function.
Namely, we explain how using unsupervised representations, as per~\citep{semi_epig}, can itself break down with messy pools.

\looseness=-1
To this end, we first highlight three key features that can occur with real--world pools and be problematic for active learning approaches: \textit{i)} \textbf{class imbalance}, an uneven distribution of classes; \textit{ii)} \textbf{redundant classes}, where the pool contains data points that are not one of the classes we wish to predict
(e.g.~our pool is images of animals but our task is specifically classifying dog breeds); and \textit{iii)} \textbf{redundant information}, where there is significant information in the data that is irrelevant to what we are trying to predict 
(e.g.~detecting legions in MRI brain data, where variations in equipment used to make scans causes large irrelevant variations in data points).
These ``messiness'' characteristics are present in a wide range of scenarios ranging from web-scraped data to natural data \citep{DeepALRealisticPool, DeepALWILD, messyData1, messyData2} and also encompass most forms of data messiness considered in prior work \citep{GALAXY, SIMILAR, deepOpenActiveLearning, zhang2023algorithm, nuggehalli2023direct}.

\looseness=-1
A weakness of unsupervised representations here is that as our data becomes increasingly messy, the representations may fail to capture all the information relevant for our task. This has been observed outside of the active learning context for various unsupervised representation learning methods \citep{UnsupervisedPretrainUncurated, DivideContrastUncurated, RobustnessRepresentationDistShiftUncurated, FixAStep}. At a high level, it comes from unsupervised representations being \textit{task--agnostic}: as the pool becomes messier, the \textit{task--specific} information becomes smaller compared to the task--irrelevant information, and the representation increasingly focuses on the latter. Even if the task--specific information has not been lost completely by the unsupervised representation, its dilution will generally still increase the difficulty of learning an effective prediction head ~\citep{whendoescontrastive, FixAStep}.

\looseness=-1
A direct consequence of this is that it will inevitably hurt the performance of active learning algorithms. We observe this in Figure \ref{fig:problem} for the case of EPIG, where we perform active learning on a balanced pool, but with unsupervised representations that were pre--trained on pools of increasing messiness. This is expected, as here selecting the most informative data points relies on the model's ability to make accurate similarity judgments in the latent space \citep{semi_epig}. Capturing these similarities is essential for establishing the predictive correlations that drive effective exploration and exploitation in active learning \citep{grosseMIG, NeuralTestBED, HigherOrderPred}. However, these similarities are \emph{task--dependent} and break down with messier pools as our representations fail to include relevant task--specific information.

\section{Using Task--Driven Representations}
\label{approach}
Motivated by the issues discussed in  \S\ref{sec:shortfals}, we propose to instead use \emph{task-driven} representations for active learning with messy pools. Our suggested approach builds on the semi--supervised approach of \citet{semi_epig} described in \S\ref{sec:background}. However, instead of using a fixed unsupervised encoder, we regularly update it as we acquire more labels using a semi--supervised representation learning technique. That is, our predictive model is given by $p_\phi(y\vert{x}) = \mathbb{E}_{p(\theta_h)}\left[p_\phi(y\vert g(x), \theta_h)\right],$ where $p_\phi(y\vert z, \theta_h)$ is our prediction head with stochastic parameters $\theta_h \sim p_{\phi}(\theta_h)$ and $z=g(x)$ are our representations as before, but $g: X \rightarrow \mathbb{R}^d$ 
is now a semi--supervised encoder that utilises both the unlabelled data \emph{and} acquired labels.

\looseness=-1
There are a variety of different methods one could use to learn this {task-driven} semi--supervised encoder (e.g. \cite{SSLRL1, SSLRL3, big_self_ssl, PAWS, RoPAWS}).
Something they generally have in common is that they utilise a ``guidance classifier'',  $c: \mathbb{R}^d \to [0, 1]^{|\mathcal{Y}|}$, that maps representations to class probabilities.
This classifier will either be learned alongside the encoder itself, typically by maximising an objective that accounts for both fidelity of the representation across all the data and the performance of the classifier on the labelled data, or will be used as part of a fine--tuning procedure to update a pretrained unsupervised encoder.
The aim of this is to guide the representations to be task--driven, such that they retain the information required for both effective downstream prediction and label acquisition. To complement our task--driven representations, we use the EPIG acquisition function for our approach, which benefits from treating the guidance classifier and prediction head as distinct components, as we explain later.

\looseness=-1
The best setup to use for training/updating  the encoder will inevitably vary between problems. Below, we outline two possible concrete approaches.  The first is inspired by the CCVAE approach of~\citet{ccvae} and involves fully retraining the encoder using a semi--supervised variational objective that encourages the label information to concentrate in a small subset of the learned latents. This has the advantage of providing a low-dimensional representation that is strongly tailored to the task, but the retraining can be expensive. The second is a more lightweight approach that simply uses supervised fine--tuning of the original unsupervised representation. This has the advantage of speed and simplicity, but may make it harder to balance pool and label information in the representation.
\subsection{A Split Representation Approach}
\label{sub:split_representation_approach}
\looseness=-1
The characteristic capturing variational auto-encoder (CCVAE) approach of~\citet{ccvae} is a 
semi--supervised representation learning method that aims to capture label--specific information in the representations it learns through careful structuring and guidance of the latent space.
Specifically, they partition the latent representations as $z = z_c \cup z_{\char`\\c}$, where only $z_c$ is taken as input to the guidance classifier(s), while the whole $z$ is used in the unsupervised part of the training objective (namely reconstruction when using VAEs).
This encourages a disentanglement of the information in the representation, with $z_c$ containing the information relevant for classification.
Unlike the original CCVAE approach, we will focus on the single output setting with no further partitioning of $z_c$.
We also note that while, in the interest of simplicity, we focus on using VAE--based representations~\citep{VAE} in the following and in our experiments as per the original CCVAE approach, this general \emph{split representation} approach can be used for learning task--driven representations more generally: we simply need to update the unsupervised component of the training objective (i.e.~$\mathcal{L}(\lambda,\psi;x)$) and our architectures appropriately.

\looseness=-1
This split representation perspective is attractive for our purposes because it first allows for relatively strong pressure to be applied to $z_c$ to be highly predictive of $y$.  This means that we can use a relatively simple prediction head (taking as inputs the lower--dimensional $z_c$) in our active learning loop that will hopefully have reliable reducible uncertainty estimates and be quick to update.  
Second, by also having an explicit representation for ostensibly non--label--relevant information, in the form of $z_{\char`\\c}$, we are well placed to perform diagnostic checks for needing to update the encoder, e.g.~by comparing the accuracy of the prediction head to a classifier trained with the full $z$.
Finally, we found this to empirically give better downstream predictions when using VAE--based representations than approaches where the classifier is used to guide the entire representation, e.g.~\citet{SSLRL1}. We now describe other key algorithmic decisions, with full details provided in the Appendix.

\looseness=-1
\noindent \textbf{Encoder training}~
Unlike in the original CCVAE, we have no need to perform generations or interventions with our representation.
We therefore eschew the introduction of an additional conditional generative model on $z_c|y$ and directly train the encoder and downstream classifier in an end--to--end manner 
using both the labelled and unlabelled data. Specifically, we maximise the following objective, corresponding to Equation (2) in~\citet{ccvae},
\begin{align}
\label{eqn:ccvae}
&\mathcal{J}(\lambda,\psi,\omega)=
\sum_{x \in \mathcal{D}_{\mathrm{pool}}}\mathcal{L}(\lambda,\psi;x) \\
&+\sum_{(x,y)\in \mathcal{D}_{\mathrm{labelled}}}\mathcal{L}(\lambda,\psi;x)
+
\alpha \, \mathbb{E}_{q_\lambda(z \mid x)}\!\left[\left\{c_{\omega}(z_c)\right\}_y\right] \nonumber
\end{align}
\looseness=-1
where $\mathcal{L}(\lambda,\psi;x)= 
\mathbb{E}_{q_\lambda(z|x)} \left[\log \left(p_\psi(x|z) p(z)/q_\lambda(z|x)\right)\right]$ is the standard VAE objective \citep{VAE}, $q_\lambda(z|x)$ is the VAE encoder with parameters $\lambda$ (and we take  $g(x) = \mathbb{E}_{q_\lambda(z|x)}[z]$), $p_\psi(x|z)$ is the VAE decoder with parameters $\psi$, $p(z)$ is a fixed isotropic Gaussian prior, $c_\omega$ is the downstream classifier with parameters $\omega$, $\mathcal{D}_{\mathrm{pool}}$ is the unlabelled pool data, $\mathcal{D}_{\mathrm{labelled}}$ is the labelled data gathered thusfar, and $\alpha$ is a hyperparameter controlling the label pressure on $z_c$.

\looseness=-1
Following~\citet{ccvae}, we perform the optimization using stochastic gradient ascent where updates with the labelled and unlabelled data are conducted in separate batches.
As semi--supervised encoders typically struggle with class imbalance and the low--data regimes considered in active learning \citep{realisticSSL, openSSL, safeSSL}, we further perform simple data augmentations on our labelled set and upsample minority classes. To deal with the redundant classes in our pool, we follow \citet{epig, semi_epig} by labelling them as a single ``redundant'' category and retaining them in our labelled set, noting that these labels still contain useful information for future acquisition by marking points as not being one of the target classes \citep{yang2023not}.

\looseness=-1
\noindent \textbf{Classifier and prediction head}~~
While on the face of it the guidance classifier, $c_\omega$, and the prediction head, $p_\phi(y\vert z, \theta_h)$, are both simply predictors for the output given the representations, 
the differing needs of representation learning and label acquisition means 
their roles in our pipeline, and thus desirable characteristics, differ significantly.
We, therefore, generally recommend that they are chosen separately. The guidance classifier must be differentiable but need not be probabilistic (indeed it will generally want to be deterministic to make the encoder training easier).  It is typically beneficial for it to have limited capacity and be smoothly varying in its inputs, as this forces the encoder to learn a $z_c$ from which it is easy to make predictions.  In our experiments, we use a simple neural network with one hidden layer of $128$ units.

\looseness=-1
The prediction head, on the other hand, needs to be probabilistic with well--calibrated reducible uncertainty estimates.  It will be updated at every iteration so it should ideally be cheap to update, and it should not require careful hyperparamter tuning or access to validation data.  In our experiments, we use Random Forests~\citep{breiman2001random}, due to their fast training and strong ``out--of--the--box'' performance, and ablate with their different prediction heads in the Appendix.

\looseness=-1
\noindent \textbf{Encoder retraining}~~
We retrain our encoder regularly after every $k$ acquired labels. We recommend using larger values of $k$ ($\gtrsim 25$ in our experiments) to keep computational costs low and because very small choices of $k$, and in particular taking $k=1$, has the potential to harm performance, by creating a disconnect between the update strategy assumed by the acquisition function  (which is based only on the prediction head) and the actual updates performed.

\subsection{A Representation Fine--Tuning Approach}
\label{sub:finetuning_approach}

\looseness=-1
A potential weakness of the previous approach is the cost of retraining the representation at regular intervals.
While this may be perfectly acceptable in some settings, noting that active learning is usually only applied when the cost of labelling significantly outweighs updating the model, there may be cases where it is not viable, such as when we have very large and complex pools, or we are using a separate pre--trained encoder instead of one learned from the pool data.
We, therefore, now consider a more lightweight approach that simply uses the labels to fine--tune the representation. 

\looseness=-1
The approach naturally starts with some initial representation defined by an encoder $g: X \to \mathbb{R}^d$.
This can either be trained directly on the pool data using any powerful unsupervised representation learning technique---such as those based on contrastive learning (e.g. SimCLR \citep{SimpleSimCLR}), clustering (e.g. SwAV \citep{swav}, DeepCluster \citep{deepcluster}) or masked autoencoders (e.g. MAE \citep{MAE})---or it can be taken as a fixed pretained encoder trained on some other data, such as ESM-3 for protein sequences~\citep{hayes2025simulating}. In the latter case, the representation does not necessarily need to have been trained in an unsupervised manner itself, but it will inevitably not have information about the labels of the task at hand as these are yet to be collected.
For our experiments we will use an encoder trained on the pool data using SimCLRv2~\citep{big_self_ssl}.

\looseness=-1
Once initialised, we extend $g$ by adding a guidance classifier $c_\omega: \mathbb{R}^d \to [0, 1]^{|\mathcal{Y}|}$ to its final layer, resulting in the model $g \circ c: X \to [0, 1]^{|\mathcal{Y}|}$, which we train in a fully supervised manner using the acquired labels following~\citet{big_self_ssl}. The final representations are taken from $g$ after this fine--tuning as before.
For $c_\omega$, we follow~\citet{big_self_ssl} and use what is effectively a 1-layer neural network with a modest number of hidden units. For the same reasons discussed in \S\ref{sub:split_representation_approach}, we do not use $c_\omega$ for our prediction head, again using a random forest in our experiments instead. Further algorithmic details on the precise approach used for the experiments and ablations with different prediction heads are given in the Appendix.

\begin{table*}[t]
\captionsetup{font=footnotesize}
\caption{Final test accuracy of different active learning methods on the \textbf{F+MNIST}, \textbf{CIFAR-10+100} and \textbf{CheXpert} datasets. We report the mean $\pm1$ standard error over 4 seeds.
The full active learning curves are given in the Appendix.}
\label{tab:comparisons}
\centering
\footnotesize
\begin{tabular}{c|c|c|c} 
\hline
\textbf{Method} & \textbf{F+MNIST} & \textbf{CIFAR-10+100} & \textbf{CheXpert} \\
\hline
\textbf{SIMILAR \citep{SIMILAR}} & $93.82 \pm 0.18$ & $30.87 \pm 1.57$ & $71.94 \pm 0.47$ \\
\textbf{GALAXY \citep{GALAXY}} & $84.74 \pm 0.74$ & $55.28 \pm 0.42$ & $78.76 \pm 0.35$ \\
\textbf{Cluster Margin \citep{ClusterMargin}} & $94.24 \pm 0.17$ & $32.79 \pm 0.45$ & $75.41 \pm 0.29$ \\
\textbf{US+EPIG (SimCLRv2, \cite{semi_epig})} & $98.53 \pm 0.12$ & $76.19 \pm 0.42$ & $77.84 \pm 0.28$ \\
\textbf{US+EPIG (VAE, \cite{semi_epig})} & $94.50 \pm 0.34$ & $30.47 \pm 1.17$ & $66.69 \pm 0.56$ \\
\textbf{US Random (SimCLRv2)} & $92.76 \pm 2.37$ & $74.60 \pm 0.35$ & $74.70 \pm 0.07$ \\
\textbf{US Random (VAE)} & $86.50 \pm 2.24$ & $27.90 \pm 0.87$ & $65.60 \pm 0.18$ \\
\hline
\hline
\textbf{TD-FT Random} & $96.23 \pm 0.37$ & $77.14 \pm 0.42$ & $81.67 \pm 0.40$ \\
\textbf{TD-SPLIT Random} & $88.19 \pm 2.62$ & $54.90 \pm 2.31$ & $75.79 \pm 0.75$ \\
\textbf{TD-SPLIT (Ours)} & $\mathbf{98.46 \pm 0.17}$ & $59.84 \pm 1.25$ & $76.47 \pm 0.27$ \\
\textbf{TD-FT (Ours)} & $\mathbf{99.56 \pm 0.10}$ & $\mathbf{80.90 \pm 0.75}$ & $\mathbf{83.23 \pm 0.38}$ \\
\end{tabular}
\end{table*}

\vspace{-3pt}
\section{Related Work}
\label{related_work}
\vspace{-3pt}
Previous works on active learning with messy, uncurated pools have primarily focused on developing new acquisition strategies \citep{GALAXY, SIMILAR, zhang2023algorithm, nuggehalli2023direct, russakovsky2015imagenet}, neglecting the importance of using a model that incorporates information from the unlabelled data.
Indeed, they have mainly used fully supervised models, with some approaches initialising their model with weights pre--trained on ImageNet in a fully supervised fashion or from foundations models such as CLIP \citep{radford2021learning, deepOpenActiveLearning, nuggehalli2023direct}. We compare with these alternatives in \S\ref{sec:foundation_models} and find they significantly underperform compared to using representations trained on the pool.

\looseness=-1
On the other hand, various work have considered using semi--supervised models in active learning \citep{SSLAL2, ALPretrainedLLM, SSLAL4, SSLAL5, sener2017active, yehuda2022active, luth2023navigating}, with several works showing supervised models significantly lagging behind semi--supervised ones in downstream performance \citep{semi_epig, TypiClust, yehuda2022active}. In general, there have been questions raised about the benefits of active learning when semi--supervised models are used: many acquisition strategies designed for fully supervised models have been shown to no longer provide reliable gains over random acquisition in the semi-supervised setting~\citep{sener2017active, yehuda2022active, luth2023navigating, chan2021marginal, bhatnagar2020pal, ebrahimi2020minimax, gao2020consistency}. 
However, \citet{semi_epig} recently showed that EPIG does reliably outperform random acquisition, as well as various other acquisition strategies, in this setting, using their unsupervised representation learning approach.

\vspace{-3pt}
\section{Experiments}
\vspace{-3pt}
\label{experiments}
\looseness=-1
We refer to the approaches introduced in \S\ref{sub:split_representation_approach}, \ref{sub:finetuning_approach} as \textbf{TD-SPLIT} and \textbf{TD-FT} respectively. To validate them, we first compare with
various baselines that have been specifically designed for, or shown strong performance on, the messy pool scenarios we are interested in (c.f.~\S\ref{sec:shortfals}).
As part of this, we include different variants of the unsupervised representation (\textbf{US}) approach of~\citet{semi_epig}.
All experiments were run on an NVIDIA H100 80GB GPU. 

\looseness=-1
\textbf{Datasets:} To construct datasets with redundant classes and class imbalance, we combined existing benchmarks. Our \textbf{F+MNIST} dataset uses the digits ``5'' and ``6'' from MNIST \citep{deng2012mnist} as the target classes for active learning, while the entire FashionMNIST \citep{xiao2017fashion} dataset is included as redundant data. \textbf{CIFAR-10+100} uses the first five classes of CIFAR-10 \citep{CIFAR10} as its target classes, with the full CIFAR-100 dataset \citep{CIFAR10} serving as the redundant data.

\looseness=-1
We further use the \textbf{CheXpert}~\citep{CheXpert} dataset as an example of a real--world dataset with redundant information and existing class imbalance.  \textbf{CheXpert} comprises of chest X--rays taken from a variety of patients from different angles. We consider the binary classification task of identifying \emph{pleural effusion}, i.e.~fluid in the corner of the lungs. The redundant features here are the many anatomical and acquisition--related variations (e.g., bones, implants, soft tissue, scan artefacts) that are irrelevant to the  diagnosis and generally represent larger image variations than the target--relevant information \citep{ccvae}.

\looseness=-1
\textbf{Representation learning:} For all datasets, we use VAE~\citep{VAE} and SimCLRv2~\citep{big_self_ssl} encoders, pairing them with \textbf{TD--SPLIT} and \textbf{TD--FT} respectively. We also include unsupervised variants following~\citet{semi_epig} (noting that they also consider the same two encoder strategies).
For all approaches, we adopt the learning rate, batch size, and optimizer from the original papers and train for 500 epochs.

\looseness=-1
\textbf{Models:}
For \textbf{F+MNIST} and \textbf{CheXpert}, we use the Burgess encoder \citep{burgess2017understanding} (and decoder) for our VAE--based representations and otherwise use a ResNet18 encoder \citep{resnet18}. For \textbf{CIFAR-10+100}, we replace the ResNet18 with a ResNet50 and the Burgess encoder with a ResNet--VAE \citep{kingma2016improved}.

\looseness=-1
\textbf{Active learning:}
We run active learning for a budget of 500, 6000, and 10,000 labels for \textbf{F+MNIST}, \textbf{CheXpert} and \textbf{CIFAR--10+100} respectively. We use batch acquisitions for all datasets, with a batch size of 10 for \textbf{F+MNIST} and 100 for the others. We use the 
``power'' batch acquisition strategy from~\cite{SupervisedAL6} with $\beta=4$ for \textbf{F+MNIST} and $\beta=8$ otherwise. We make this choice as this strategy is both highly scalable and has been shown to give performance comparable to more sophisticated batch acquisition strategies. We re--train our semi--supervised encoders every 5 acquisition rounds and ablate with different re--training periods in the Appendix.
\begin{figure*}[!t]
\centering
\begingroup
\captionsetup{font=small,skip=2pt}
\captionsetup[sub]{font=footnotesize,skip=2pt}
\resizebox{1\textwidth}{!}{
\begin{minipage}{\textwidth}
\centering
\includegraphics[width=0.8\textwidth]{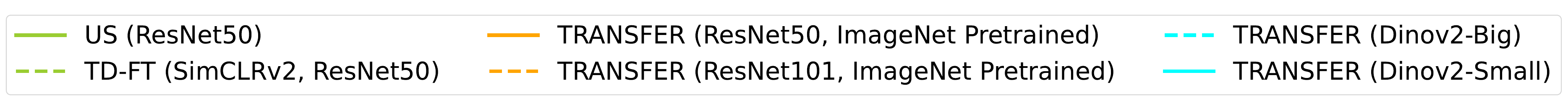}

\begin{subfigure}[b]{0.32\textwidth}
  \centering\includegraphics[width=\linewidth]{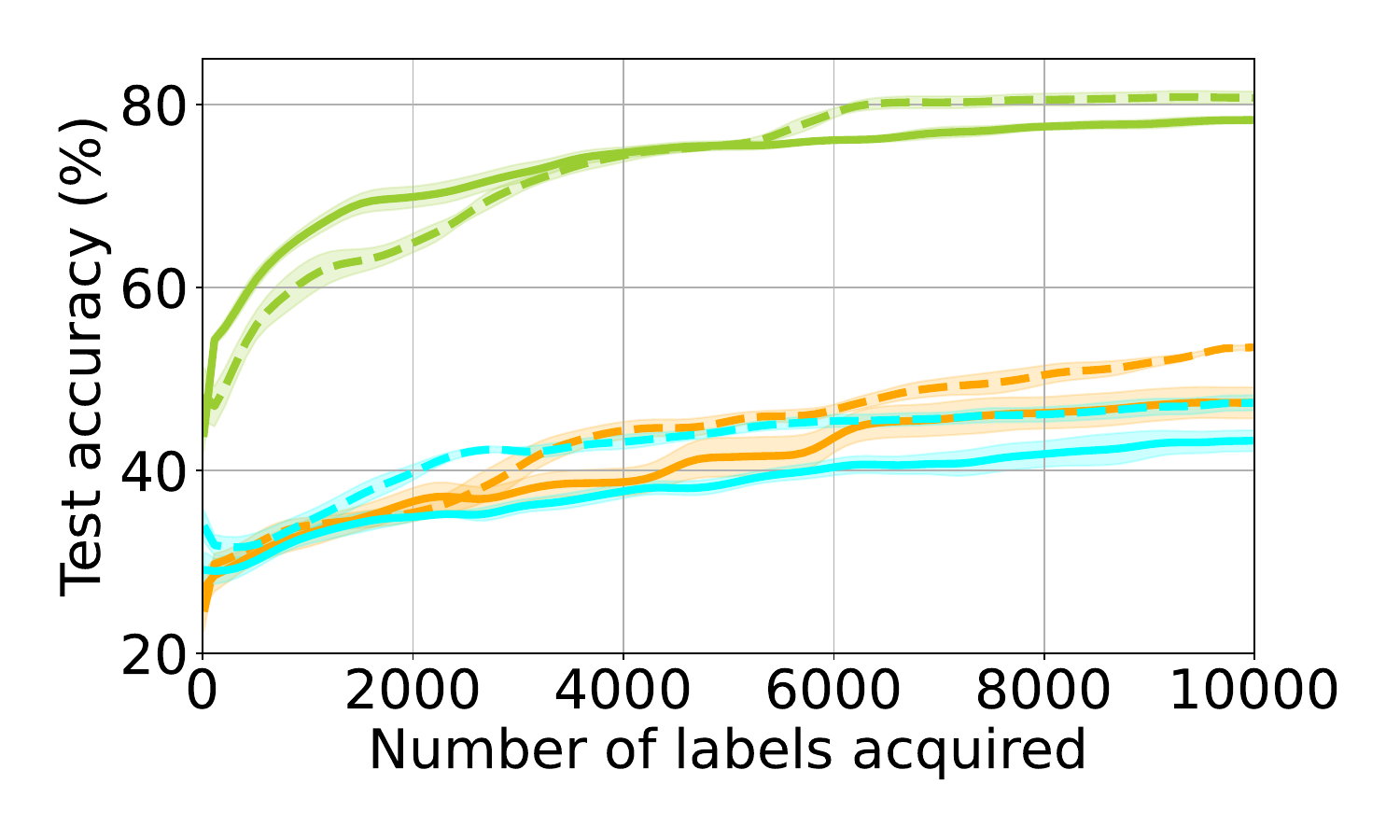}
  \caption{EPIG}\label{fig:cifar_epig}
\end{subfigure}\hfill
\begin{subfigure}[b]{0.32\textwidth}
  \centering\includegraphics[width=\linewidth]{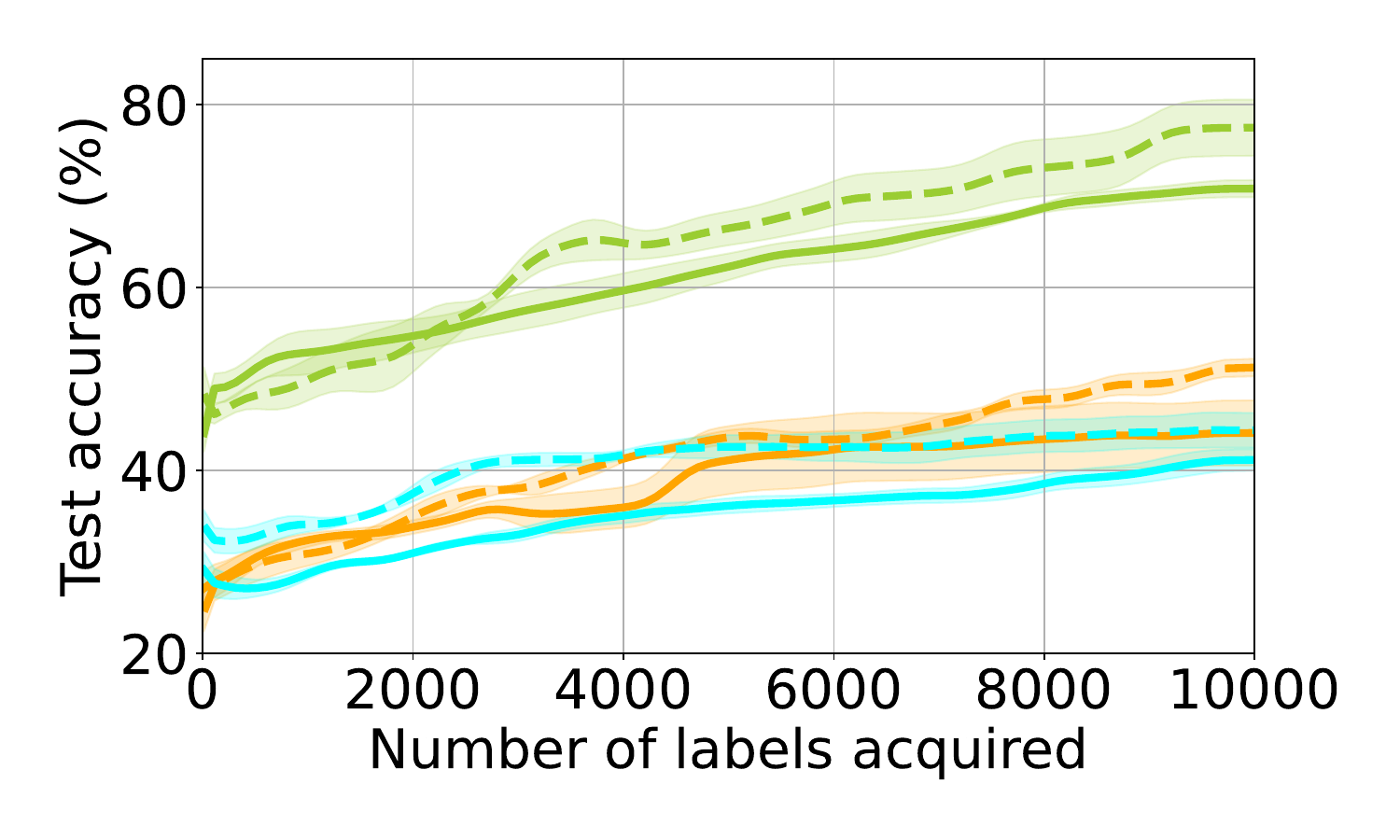}
  \caption{BALD}\label{fig:cifar_bald}
\end{subfigure}\hfill
\begin{subfigure}[b]{0.32\textwidth}
  \centering\includegraphics[width=\linewidth]{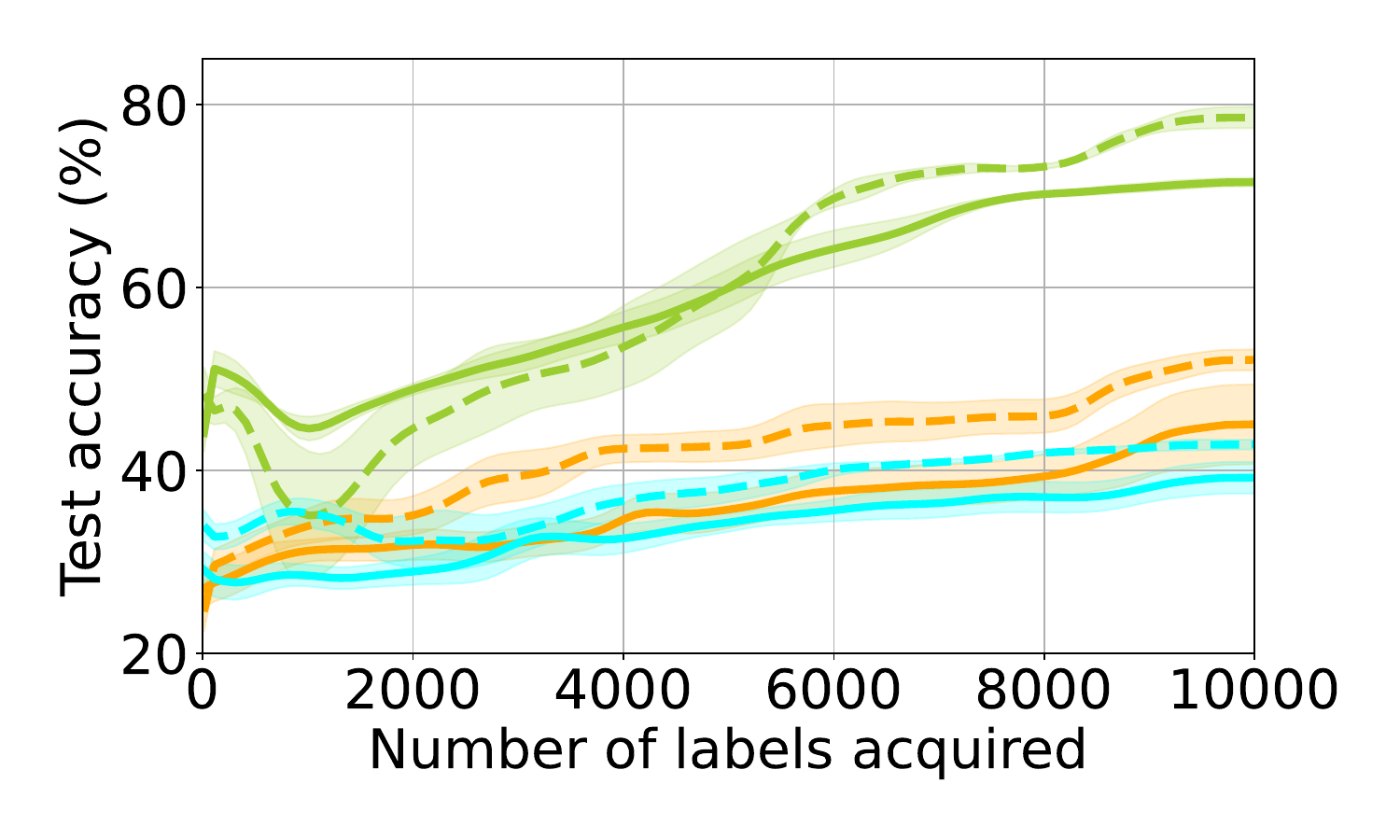}
  \caption{CS}\label{fig:cifar_cs}
\end{subfigure}
\end{minipage}
} 
\caption{Test accuracy on \textbf{CIFAR-10+100} using the \textbf{US} approach, our task-driven \textbf{TD-FT} approach, and a \textbf{TRANSFER} learning approach.
All experiments were run for 4 seeds. Solid line shows mean and shading $\pm 1$ standard error.}
\label{fig:cifar-10+100}
\endgroup
\vspace{-8pt}
\end{figure*}

\subsection{Comparison with Existing Approaches}
\label{sub:comparisons}

\looseness=-1
To highlight the effectiveness of our approach, we first compare against baselines designed for (or shown good performance on) problems with messy pools: \textbf{SIMILAR} \citep{SIMILAR}, \textbf{Cluster Margin} \citep{ClusterMargin}, and \textbf{GALAXY} \citep{GALAXY}, and the approach in \citet{semi_epig} which we refer to as \textbf{US+EPIG}.
We also compare with random acquisition using the \textbf{US}, \textbf{TD-SPLIT}, and \textbf{TD-FT} approaches.
We implement the baselines as in the original papers.

From Table \ref{tab:comparisons}, we see that our \textbf{TD-FT} approach outperforms all our baselines on all the datasets. The importance of our model setup with task-driven representations
is further highlighted by the fact that even with a simple random acquisition strategy, this model generally outperforms all the existing baselines, providing an emphatic demonstration that acquisition strategy is not the only thing to consider when dealing with messy pools.
Moreover, we also find that we can significantly boost the performance of the baselines by integrating them within our approach, though they still fall short of our \textbf{TD-FT} approach (see Appendix).

Our \textbf{TD-SPLIT} method also shows strong performance relative to the baselines, albeit generally underperforming \textbf{TD-FT}. 
This is because it is using a much more lightweight encoder: it still consistently and comprehensively outperforms the analogous \textbf{US+EPIG} approach with VAE--based representations, thus again emphasising the importance of using task--driven representations.

\begin{figure*}[t!]
\centering
\begingroup
\captionsetup{font=small,skip=2pt}
\captionsetup[sub]{font=footnotesize,skip=2pt}
\resizebox{0.93\textwidth}{!}{
\begin{minipage}{\textwidth}
\begin{subfigure}[b]{0.32\textwidth}
  \centering\includegraphics[width=\linewidth]{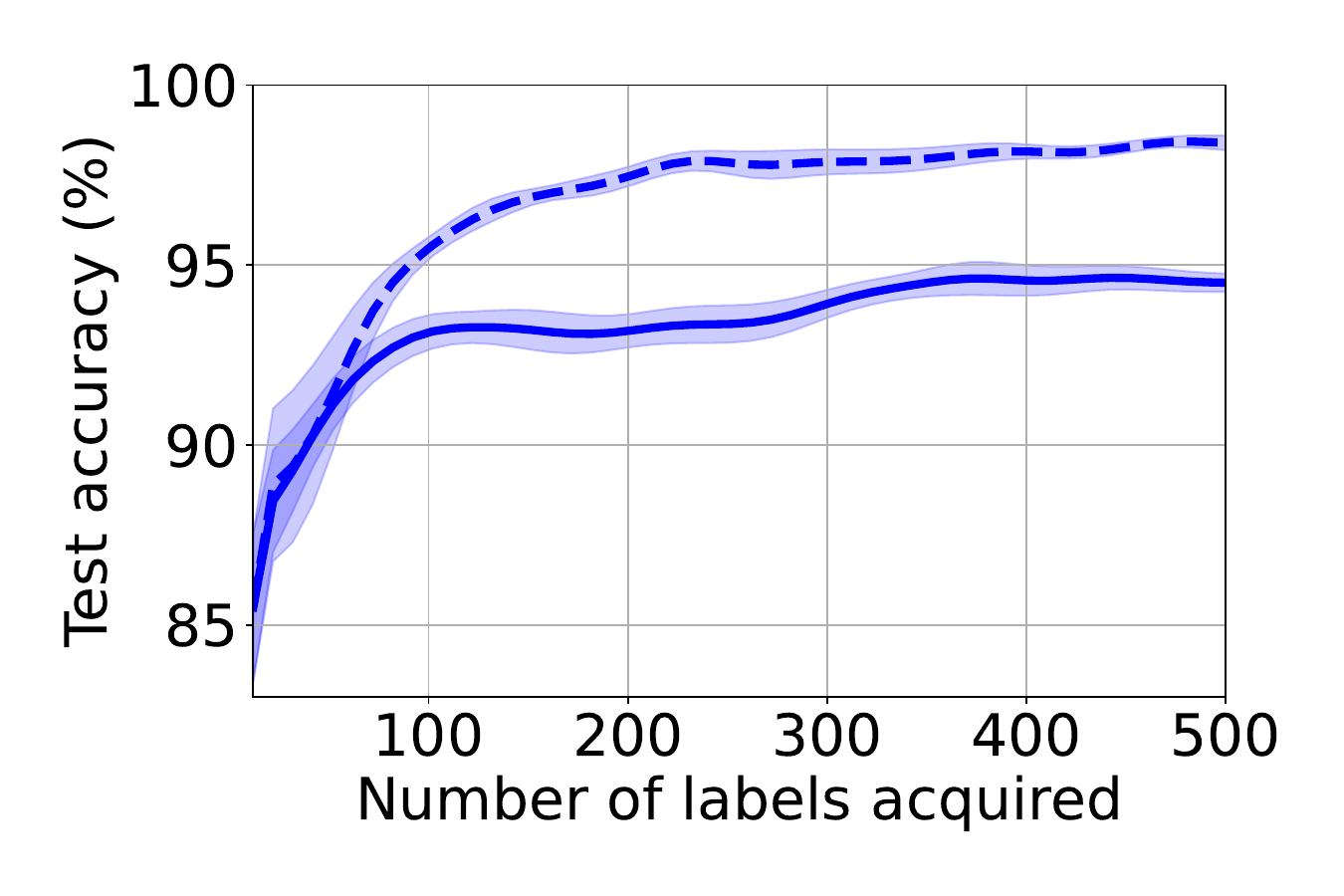}
  \caption{EPIG}\label{fig:fmnsit_vae_epig}
\end{subfigure}\hfill
\begin{subfigure}[b]{0.32\textwidth}
  \centering\includegraphics[width=\linewidth]{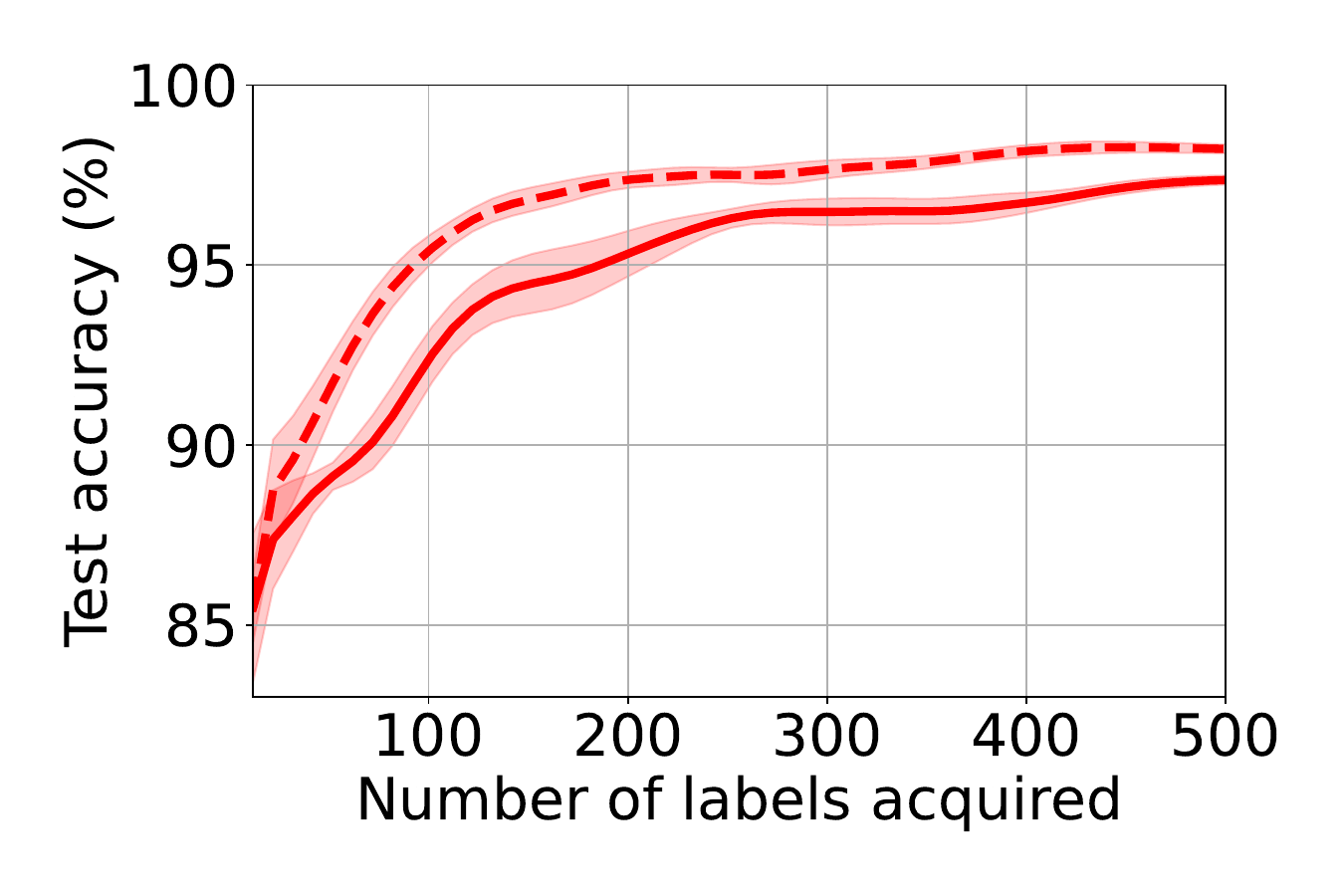}
  \caption{BALD}\label{fig:fmnsit_vae_bald}
\end{subfigure}\hfill
\begin{subfigure}[b]{0.32\textwidth}
  \centering\includegraphics[width=\linewidth]{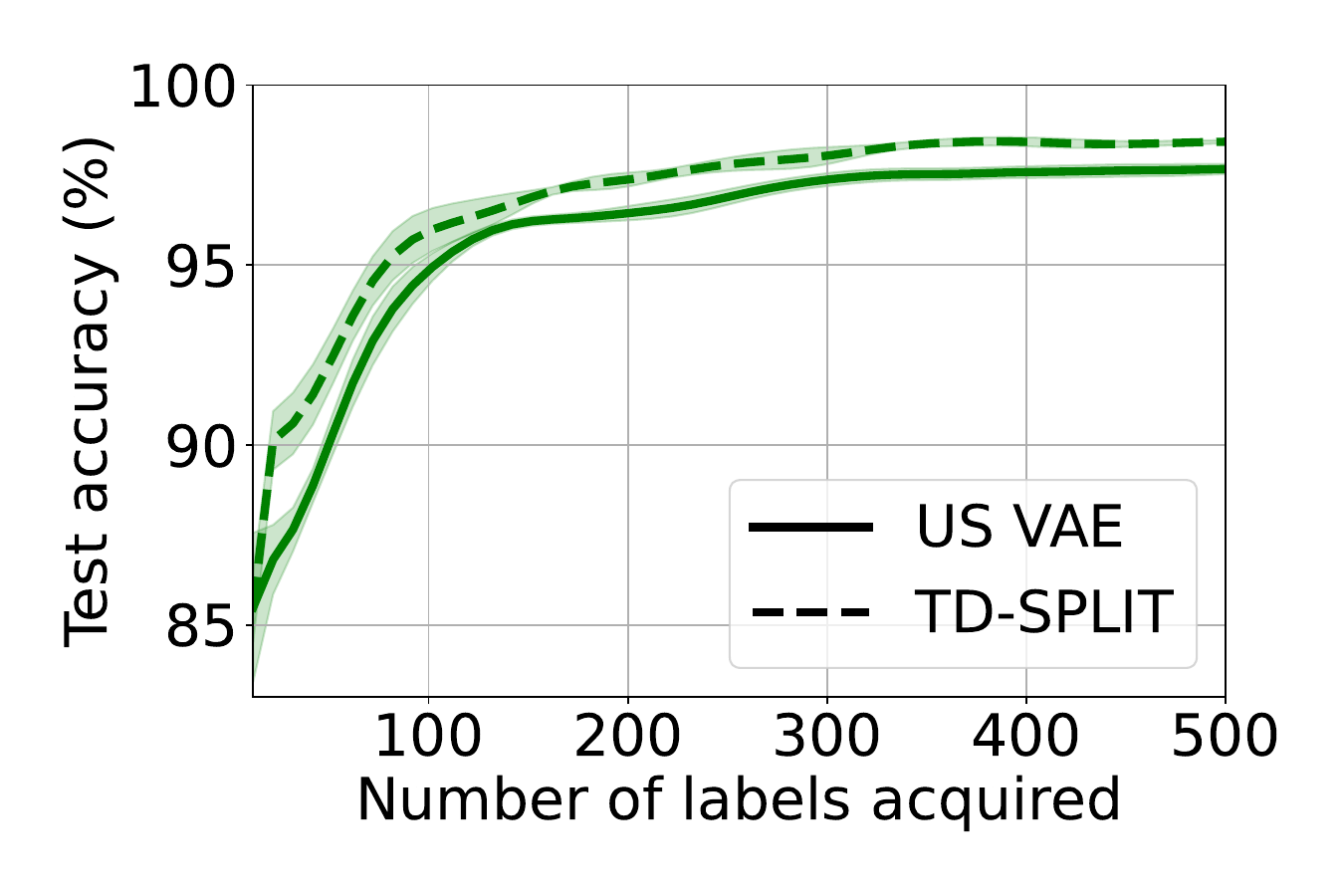}
  \caption{CS}\label{fig:fmnsit_vae_cs}
\end{subfigure}
\begin{subfigure}[b]{0.32\textwidth}
  \centering\includegraphics[width=\linewidth]{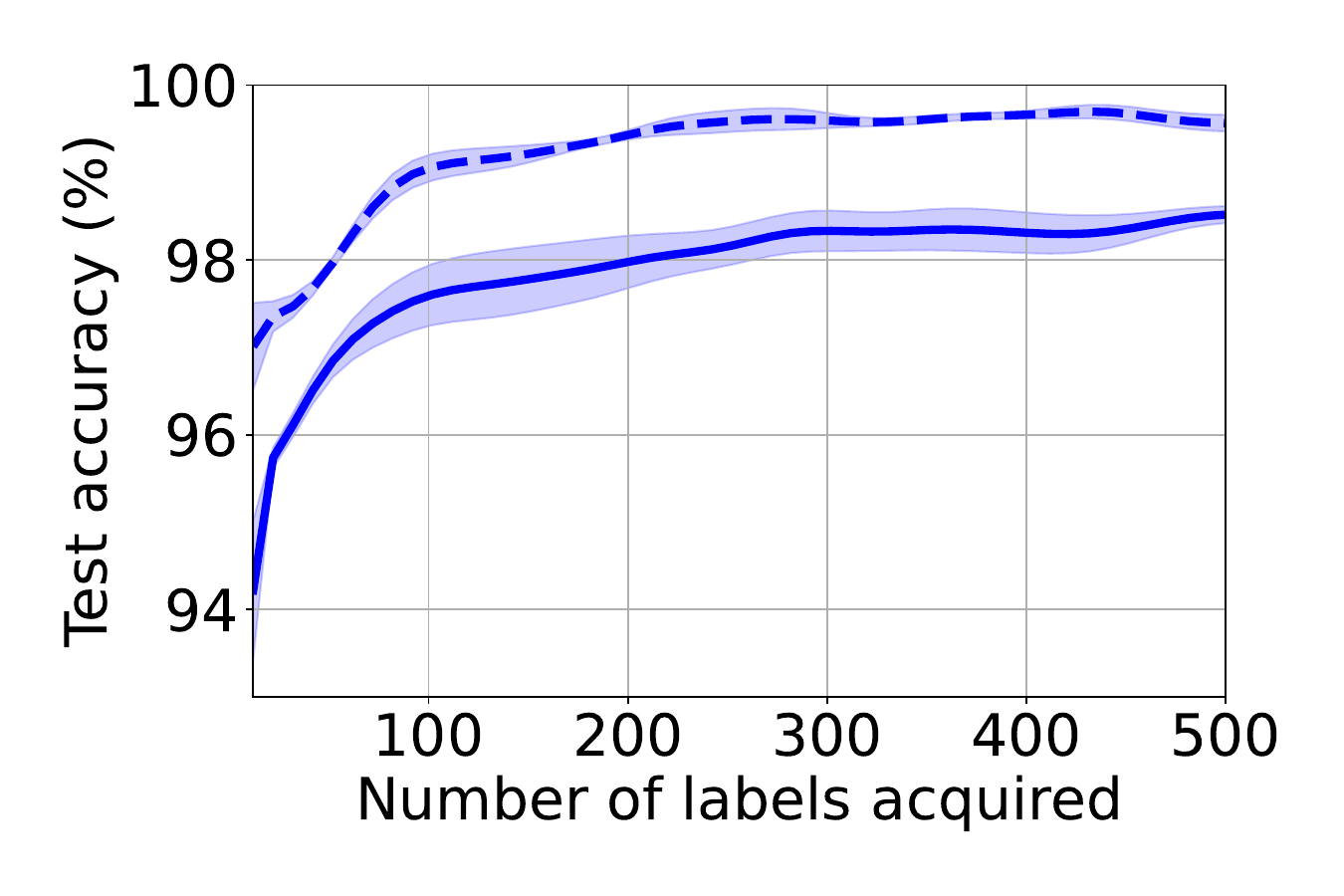}
  \caption{EPIG}\label{fig:fmnsit_simclr_epig}
\end{subfigure}\hfill
\begin{subfigure}[b]{0.32\textwidth}
  \centering\includegraphics[width=\linewidth]{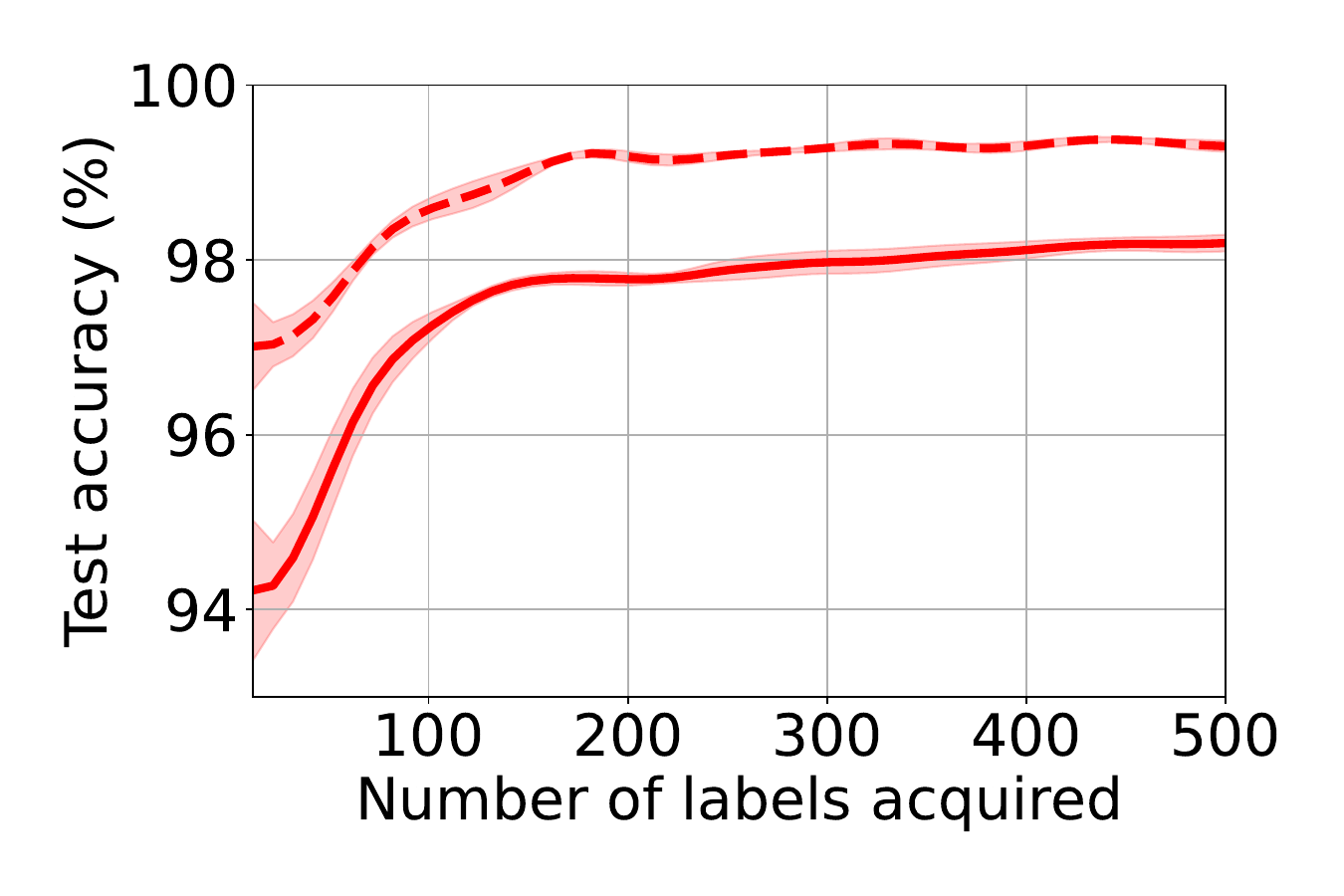}
  \caption{BALD}\label{fig:fmnsit_simclr_bald}
\end{subfigure}\hfill
\begin{subfigure}[b]{0.32\textwidth}
  \centering\includegraphics[width=\linewidth]{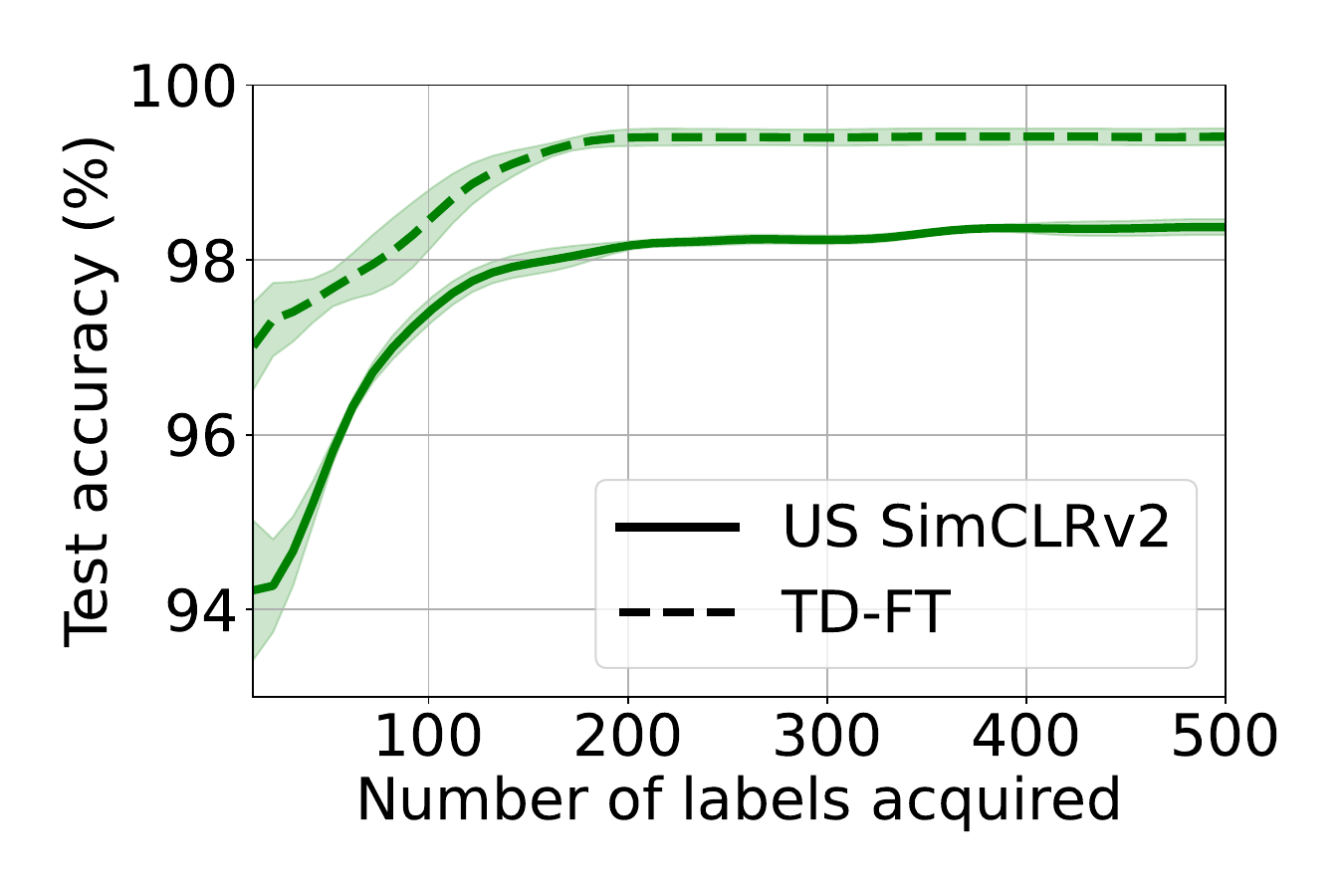}
  \caption{CS}\label{fig:fmnsit_simclr_cs}
\end{subfigure}
\end{minipage}
}

\caption{Test accuracy on \textbf{F+MNIST} using the \textbf{US} approach and our task--driven approach.
Top and bottom row respectively show the results using VAE--based encoders and SimCLRv2 encoders. 
Solid line shows mean and shading $\pm 1$ standard error across 4 seeds.
}
\label{fig:fmnist}
\endgroup
\end{figure*}

\begin{figure*}[t!]
\centering
\begingroup
\captionsetup{font=small,skip=2pt}
\captionsetup[sub]{font=footnotesize,skip=2pt}
\resizebox{0.93\textwidth}{!}{
\begin{minipage}{\textwidth}
\begin{subfigure}[b]{0.32\textwidth}
  \centering\includegraphics[width=\linewidth]{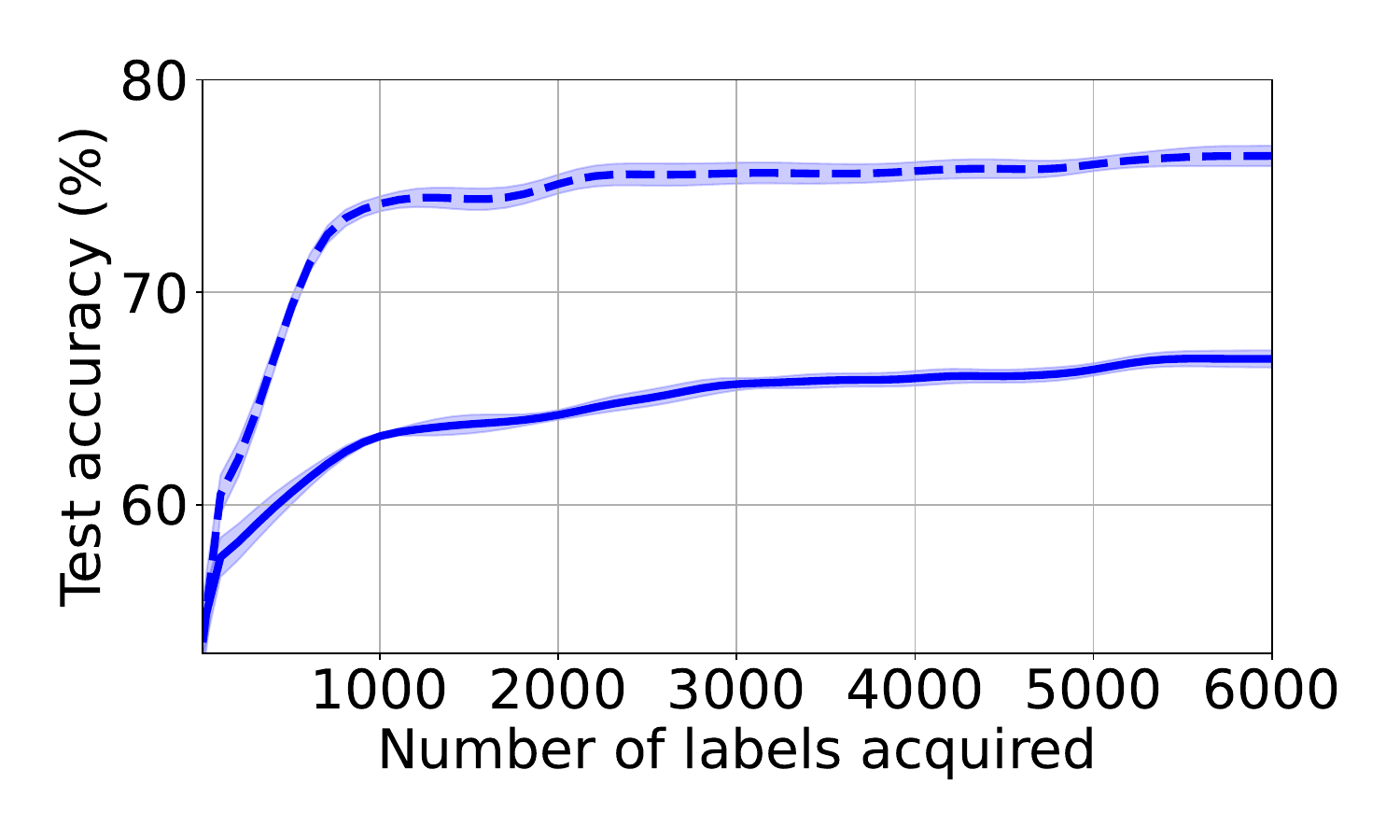}
  \caption{EPIG}\label{fig:sub1}
\end{subfigure}\hfill
\begin{subfigure}[b]{0.32\textwidth}
  \centering\includegraphics[width=\linewidth]{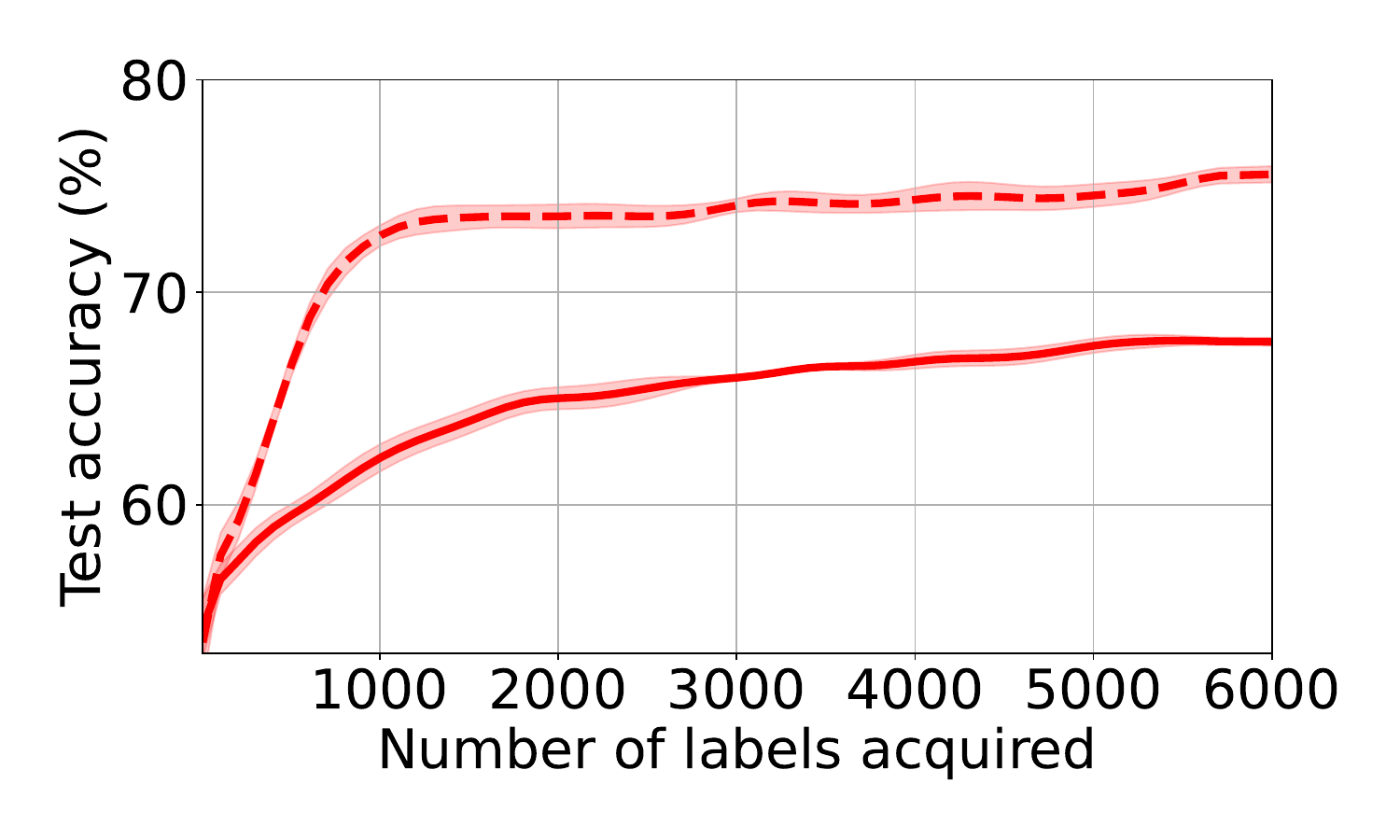}
  \caption{BALD}\label{fig:sub2}
\end{subfigure}\hfill
\begin{subfigure}[b]{0.32\textwidth}
  \centering\includegraphics[width=\linewidth]{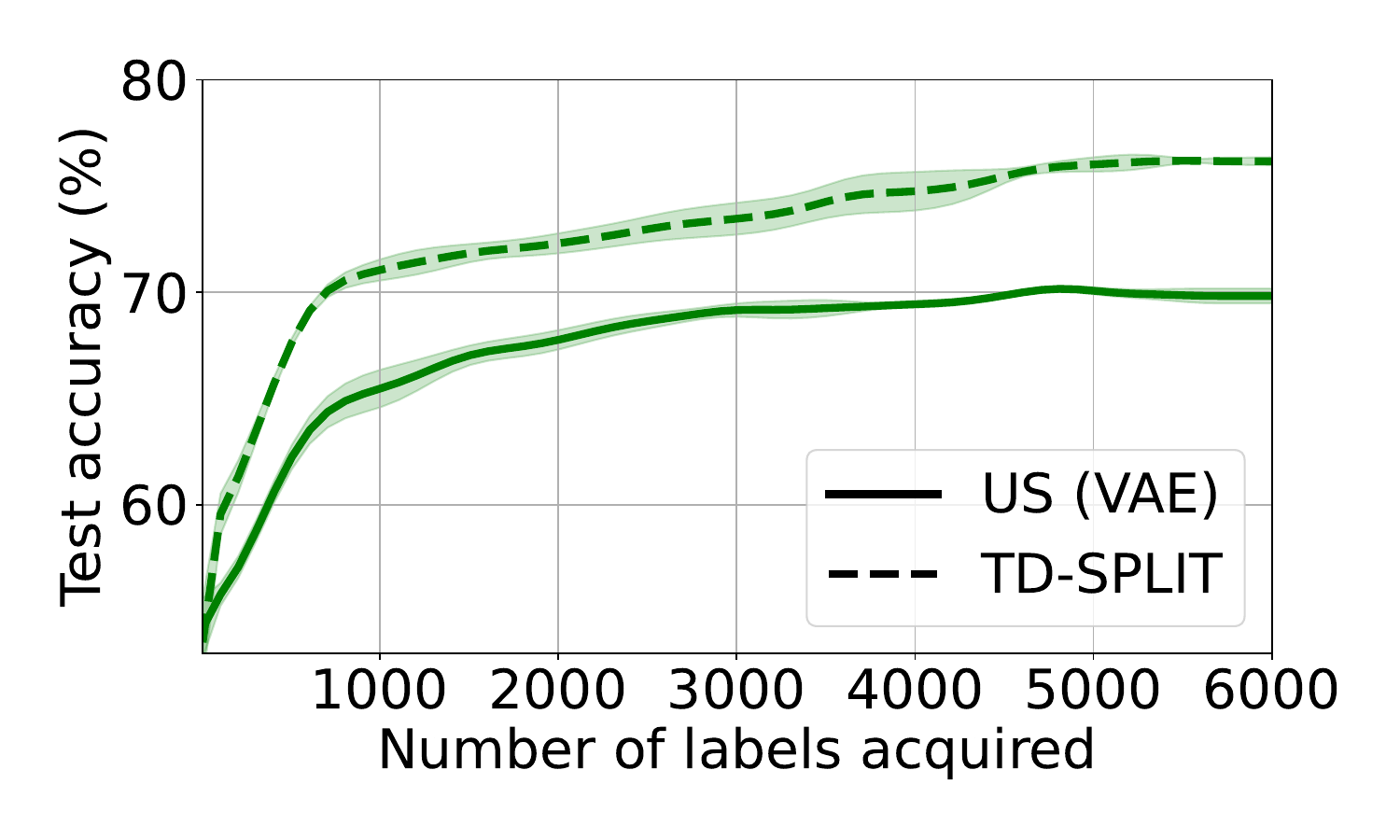}
  \caption{CS}\label{fig:sub3}
\end{subfigure}
\begin{subfigure}[b]{0.32\textwidth}
  \centering\includegraphics[width=\linewidth]{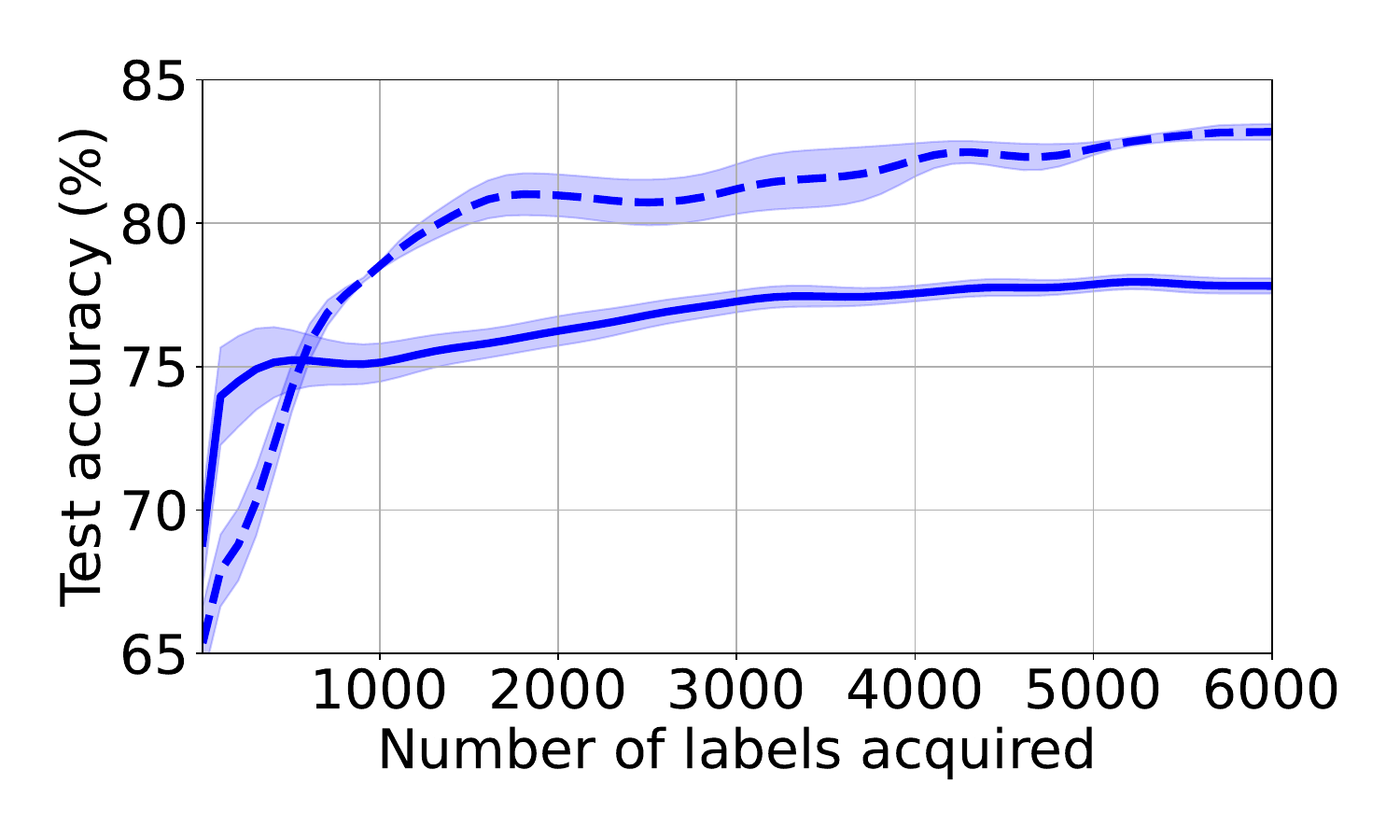}
  \caption{EPIG}\label{fig:sub4}
\end{subfigure}\hfill
\begin{subfigure}[b]{0.32\textwidth}
  \centering\includegraphics[width=\linewidth]{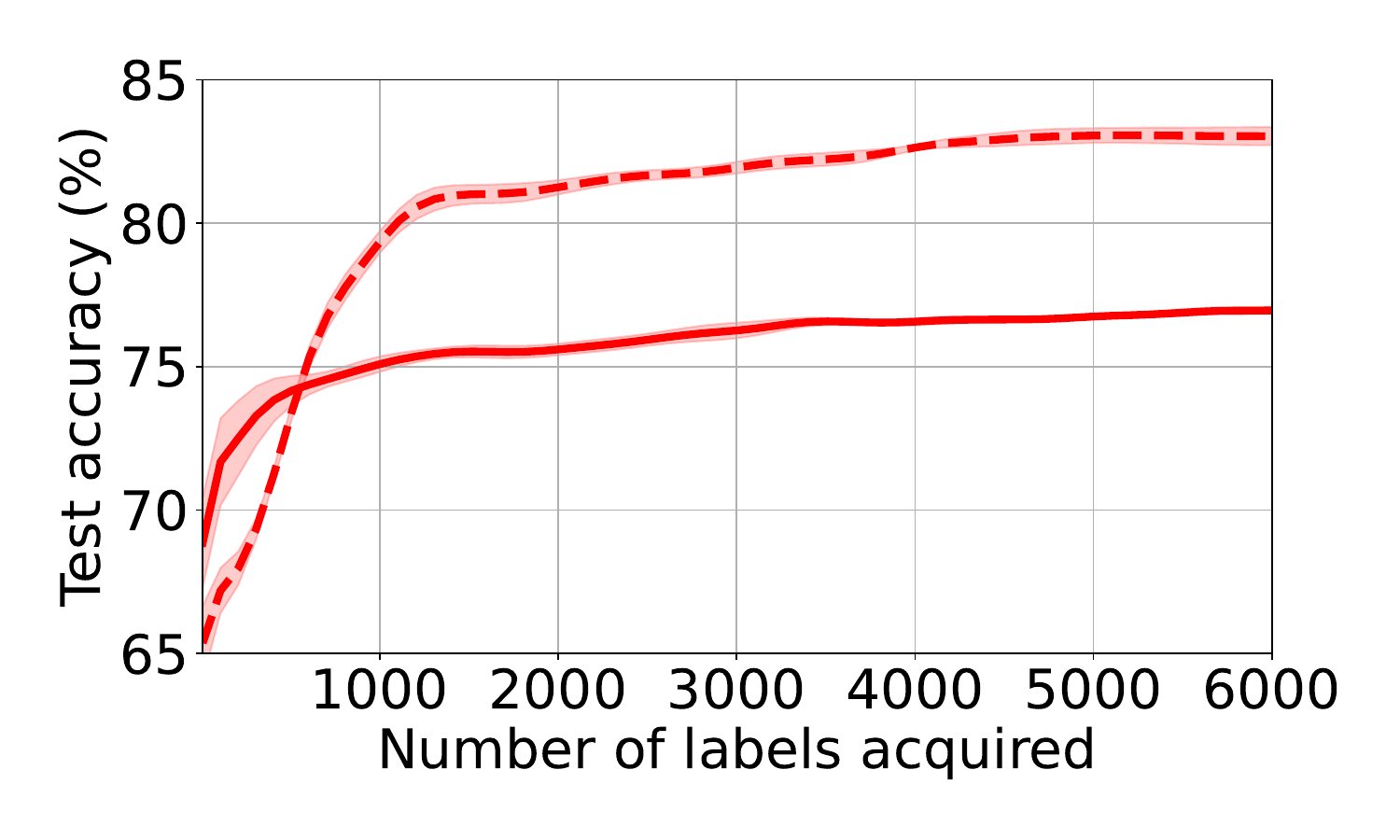}
  \caption{BALD}\label{fig:sub5}
\end{subfigure}\hfill
\begin{subfigure}[b]{0.32\textwidth}
  \centering\includegraphics[width=\linewidth]{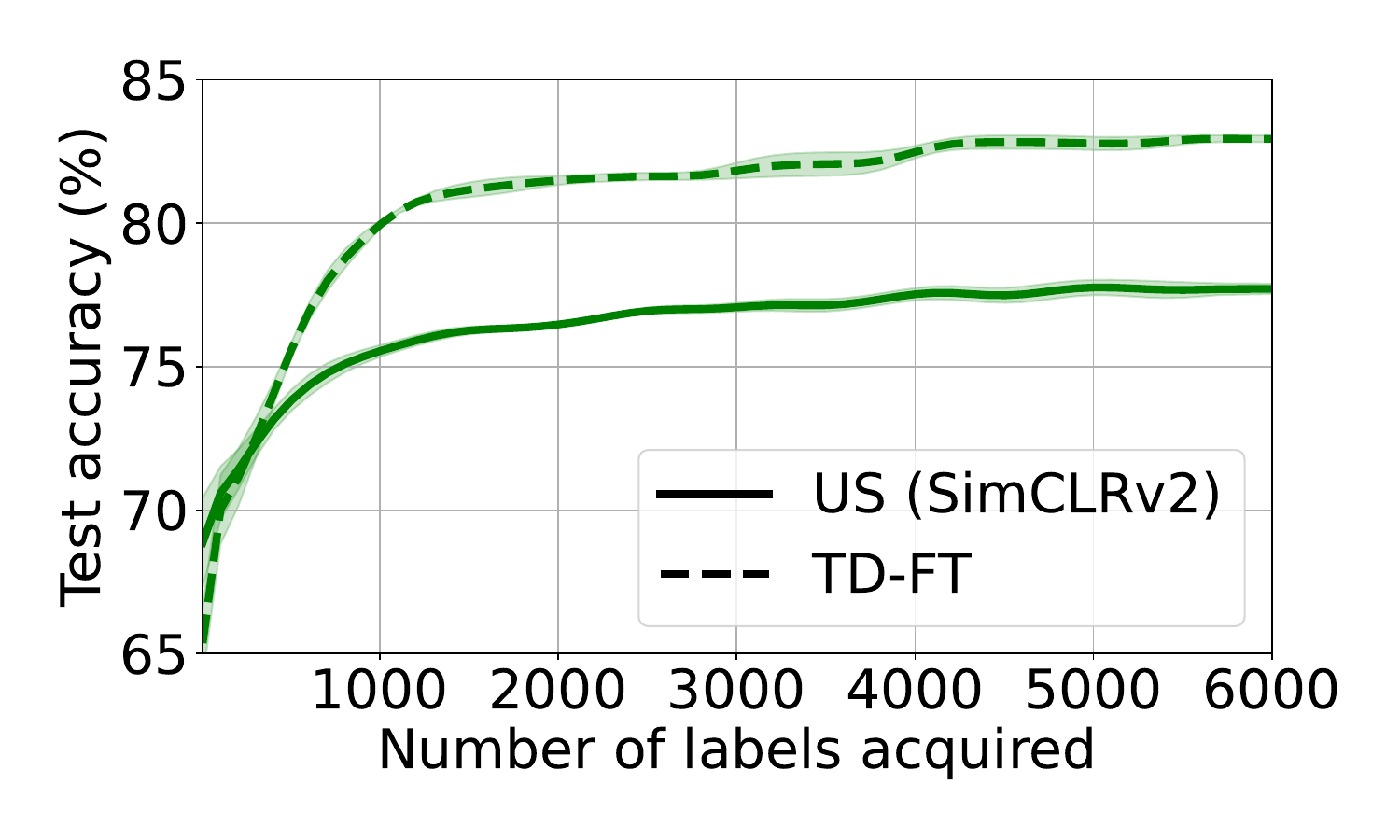}
  \caption{CS}\label{fig:sub6}
\end{subfigure}
\end{minipage}
}
\caption{Test accuracy on \textbf{CheXpert} using the \textbf{US} approach and our task--driven approach.
Top and bottom row respectively show the results using VAE--based encoders and SimCLRv2 encoders.  
Solid line shows mean and shading $\pm 1$ standard error across 4 seeds.
}
\label{fig:chexpert}
\endgroup
\end{figure*}

\subsection{Acquisition Counts for Different Approaches}
\label{sec:acq_counts_different_approaches}
\begin{table}[b!]
\vspace{-0.2cm}
\captionsetup{font=footnotesize}
\caption{Number of target classes that have been acquired at the end of active learning for different approaches on the \textbf{F+MNIST} and \textbf{CIFAR-10+100} datasets. We report the mean $\pm1$ standard error over 4 seeds. We omit the citations here for brevity.
}
\vspace{-4pt}
\label{tab:acq_counts_main}
\centering
\footnotesize
\setlength{\tabcolsep}{2pt} 
\begin{tabular}{c|c|c}
\hline
\textbf{Method} & \textbf{F+MNIST} & \textbf{CIFAR-10+100} \\
\hline
\textbf{SIMILAR} & $486 \pm 6$ & $920 \pm 237$ \\
\textbf{GALAXY} & $118 \pm 13$ & $1806 \pm 32$ \\
\textbf{Cluster Margin} & $32 \pm 1$ & $1287 \pm 53$ \\
\textbf{US+EPIG (SimCLRv2)} & $119 \pm 8$ & $936 \pm 35$  \\
\textbf{US+EPIG (VAE)} & $118 \pm 4$ & $529 \pm 17$  \\
\hline
\hline
\textbf{Random} & $31 \pm 2$ & $609 \pm 13$  \\
\textbf{TD-SPLIT (Ours)} & $141 \pm 14$ & $1651 \pm 25$  \\
\textbf{TD-FT (Ours)} & $206 \pm 38$ & $2215 \pm 38$ \\
\end{tabular}
\end{table}

\looseness=-1
Table \ref{tab:acq_counts_main} shows the number of acquisitions that have been made for the target classes at the end of active learning for different approaches. We first note that using our approach instead of unsupervised representations improves the count of the target classes by a large margin. This suggests that the gains displayed in Table \ref{tab:improving_baselines} are not merely from using a better model, but also from making better acquisitions (see also Table \ref{tab:aq_counts}). Secondly, we note that the best performing method (\textbf{TD-FT}) does \emph{not} always have the largest amount of target classes acquired. This suggests that the precise data point acquired is important, not just its class \citep{yang2023not}.

\subsection{Pool-Based Task-Driven Representations Outperform Transfer Learning Based AL}
\label{sec:foundation_models}
\looseness=-1
A common setup in AL papers that deal with messy pools is a transfer--learning approach where a fully supervised model initialised from a large pretrained/foundation model is fine--tuned to the task \citep{GALAXY, zhang2023algorithm, deepOpenActiveLearning, nuggehalli2023direct, foundation_models, gupte2024revisiting}. Our \textbf{TD-FT} framework can leverage these models by initialising the encoder, $g$, with their pretrained weights instead of using an unsupervised representation of the pool. We test this on the \textbf{CIFAR-10+100} dataset by comparing our standard \textbf{TD-FT} method against initializing the encoder using both publicly available pre-trained models---namely ResNet-50 and ResNet-101 pre--trained with supervision on ImageNet \citep{russakovsky2015imagenet}---and foundation models---namely two variants of the self-supervised DinoV2 model \citep{oquab2023dinov2}, pre--trained on a large, curated dataset of 142 million unlabelled images.

\looseness=-1
As shown in Figure \ref{fig:cifar-10+100}, pre--training on the unlabelled pool yields substantially better performance than fine-tuning from these general--purpose models, even when the latter includes significantly larger encoders. This further highlights the benefits of target--driven representation learning: features learned from a data source that is well--aligned with the target distribution are more effective than those from a more general--purpose model. Indeed, the ResNet models, pre--trained on ImageNet, prove more effective than the DinoV2 models. Although the latter are trained on a vastly larger and more varied dataset, ImageNet's focus on natural object classification is more closely aligned with the CIFAR target, leading to more transferable representations and ultimately better performance.

\vspace{-0.2cm}
\subsection{Alternative Acquisition Strategies}
\label{sec:ssl_improve_bal}
As an ablation to demonstrate the broader benefits of our task--driven approach with other acquisition strategies, we apply our \textbf{TD-FT} and \textbf{TD-SPLIT} approaches to BALD \citep{BALD} and Confidence Sampling \citep{confSampling}, comparing to analogous unsupervised (\textbf{US}) setups.

\looseness=-1
In Figures \ref{fig:fmnist}, \ref{fig:chexpert}, we see that both our approaches improve performance compared to \textbf{US} across all datasets and all choices of acquisition function. These observations suggest that using task--driven representations allows our model to better assess the utility of data points for our downstream task, and in turn make better downstream predictions, irrespective of our choice of acquisition function. 

\looseness=-1
Separately, we note that the EPIG acquisition function still achieves the best final accuracy compared to all other acquisition methods tested. This highlights the complementary nature of EPIG to our representation strategy, suggesting the importance of considering \emph{both} the model and the acquisition function.

\section{Conclusions}
\label{conclusions}
\looseness=-1
We have shown that effective active learning in presence of messy pools requires careful consideration of not only the acquisition function, but the model setup as well.
In particular, we have shown that using unsupervised representations can break down in the presence of messy pools, which is exactly the scenario where active learning has the most to gain compared to random acquisition.
To address this, we have proposed the use of \emph{task-driven representations} that explicitly incorporate the task information we aim to capture. Empirically, we have shown that this leads to more effective acquisitions and improved model performance.

\clearpage
\section*{Acknowledgements}
The authors are grateful to Freddie Bickford-Smith for helpful discussions. KA is supported
by the EPSRC CDT in Modern Statistics and Statistical Machine Learning (EP/S023151/1)
and TR is supported by the UK EPSRC grant EP/Y037200/1.

\bibliography{references}
\bibliographystyle{icml2026}

\clearpage
\appendix
\onecolumn
\cleardoublepage
\phantomsection

\addcontentsline{toc}{chapter}{Appendix Table of Contents}
\renewcommand{\contentsname}{\textcolor{Bittersweet}{Appendix Table of Contents}}

\setcounter{tocdepth}{3}      
\etocsetnexttocdepth{3}
\localtableofcontents
\cleardoublepage
\etocsettocstyle{\chapter*{Appendix Table of Contents}}{}

\section{Experimental Details}
\label{app:setup}
\subsection{Details for Sections \ref{sub:comparisons}, \ref{sec:ssl_improve_bal}}
\subsubsection{Datasets}
\label{appsub:datasets}

\looseness=-1
\paragraph{F+MNIST:} We used the \textbf{F+MNIST} as an example of a dataset with redundant classes and class imbalance by combining existing benchmarks. Specifically, we used the digits ``5'' and ``6'' from MNIST \citep{deng2012mnist} as the target classes for active learning while the entire FashionMNIST dataset \citep{xiao2017fashion} was included as the redundant data. We used an imbalance ratio of 10 for our pool, where the minority classes were chosen to be our target classes.

\looseness=-1
\paragraph{CIFAR-10+100:} We used the \textbf{CIFAR-10+100} as another example of a dataset with redundant classes and class imbalance by combining existing benchmarks. Specifically, we used the first 5 classes of CIFAR-10 \citep{CIFAR10} as our target classes for active learning while the entire CIFAR-100 dataset \citep{CIFAR10} was included as the redundant data. We again used an imbalance ratio of 10 in our pool, where the minority classes were chosen to be our target classes.

\looseness=-1
\paragraph{CheXpert:} We used the \textbf{CheXpert}~\citep{CheXpert} dataset as an example of a real--world dataset with redundant information and existing class imbalance.  \textbf{CheXpert} comprises of chest X--rays taken from a variety of patients from different angles. We considered the binary classification task of identifying \emph{pleural effusion}, i.e.~fluid in the corner of the lungs. We filtered out observations with ``NA'' as the response for the pleural effusion task. The imbalance ratio in our pool was 2.5.
\subsubsection{Representation learning}
\label{appsub:representations}

\begin{table*}[!ht]
\caption{Size of the latent dimension, $z$, size of $z_c$ and value of $\alpha$ for the \textbf{TD-SPLIT} and \textbf{TD-FT} approaches. We chose the first $|z_c|$ coordinates of $z$ to represent $z_c$ for the \textbf{TD-SPLIT} method.}
\vspace{-4pt}
\label{tab:td_split_latent_dimension}
\centering
\footnotesize
\begin{tabular}{lccc|c}
\toprule

\multirow{2}{*}{\textbf{Dataset}} & \multicolumn{3}{c}{\textbf{TD-SPLIT}} & \multicolumn{1}{c}{\textbf{TD-FT}} \\
\cmidrule(lr){2-4} \cmidrule(lr){5-5}
& \textbf{$|z|$} & \textbf{$|z_c|$} & \textbf{$\alpha$} & \textbf{$|z|$} \\
\midrule
\textbf{F+MNIST} & 10 & 3 & 20 & 512 \\ 
\textbf{CheXpert} & 45 & 5 & 20 & 512 \\
\textbf{CIFAR-10+100} & 200 & 50 & 40 & 2048 \\
\bottomrule
\end{tabular}
\end{table*}

\looseness=-1
For all datasets, we used VAE \citep{VAE} and SimCLRv2 encoders, pairing them with \textbf{TD-SPLIT}, \textbf{TD-FT} respectively. We also used their unsupervised variants for the \textbf{US} approach \citep{semi_epig}, where we first encoded our data points into the latent space of the unsupervised encoder, then performed active learning on the latent space with a prediction head. Below, we describe the details of the representation learning methods used.

\looseness=-1
\paragraph{VAE--based representations:}For \textbf{TD-SPLIT}, we followed the CCVAE \citep{ccvae} approach described in Section \ref{sub:split_representation_approach} and optimised objective \eqref{eqn:ccvae}. We chose $\alpha$ so that the labelled loss roughly matched the scale of the unlabelled loss. We performed the optimisation using stochastic gradient ascent where updates with the labelled and unlabelled data were conducted in separate batches. As the labelled dataset contained far fewer data points, we evenly spaced the labelled batches between the unlabelled ones. We trained our model for 500 epochs and, following \citet{ccvae}, we used a batch size of $200$ for both the unlabelled and labelled data, the Adam optimiser \citep{Adam} and a learning rate of $2 \times 10^{-4}$. Table \ref{tab:td_split_latent_dimension} shows the  sizes for $z$, $z_c$ and values of $\alpha$. 

\looseness=-1
For the \textbf{US} approach, we followed \citet{burgess2017understanding} and optimised their ELBO objective with $\beta=1$ (this amounts to the standard VAE objective). We used the Adam optimiser, a learning rate of $5 \times 10^{-4}$, a batch size of $200$ and $\text{KL}$ annealing as in \citet{burgess2017understanding}. For all datasets, we used the same latent dimensions as for \textbf{TD-SPLIT} and trained for 500 epochs.

\looseness=-1
\paragraph{SimCLRv2--based representations:}For \textbf{TD-FT}, we followed the approach in Section \ref{sub:finetuning_approach} by finetuning representations according to \citet{big_self_ssl}. Specifically, we first pretrained an unsupervised encoder following their pretraining setup (detailed below) then finetuned the encoder for 500 epochs from the first layer of the projection head. We used a batch size of 512, learning rate of 0.11, learning rate warmup and the LARS \citep{you2017large} optimiser with a momentum of 0.9.

\looseness=-1
For the \textbf{US} approach, we finetuned according to the pretraining setup in \citet{big_self_ssl} for CIFAR-10 as it was more reasonable for our datasets. Specifically, we pretrained for 500 epochs using a batch size of 512, learning rate of 2.26 which was linearly increased for the first 5\% of epochs and subsequently decayed with the cosine decay schedule. We again used the LARS optimiser with a momentum of 0.9 and a weight decay of $1 \times 10^{-4}$. We used a 1--layer projection head and a temperature of 0.2 for the contrastive loss. For our pretraining augmentations, we used random resized crop (area sampled uniformly from 8\% to 100\%) and color distortion (jitter and random grayscale) with strength 0.5.

\looseness=-1
\paragraph{Data augmentations:}For all datasets, we performed data augmentations on our labelled set during the training of the semi--supervised encoders. We used random rotations between $-20^{\circ}$ and $20^{\circ}$, random horizontal flips with probability 0.5, and random affine transformations with scale between 0.65 and 1.

\looseness=-1
\paragraph{Re--training frequency:}We re--trained our semi--supervised encoders every 5 acquisition rounds and ablate with different re--training periods in Section \ref{app:diff_retraining}. 

\subsubsection{Models}
\label{appsub:models}

Below, we describe the encoders/decoders we used for our VAE--based representations and SimCLRv2--based representations.

\paragraph{Encoders for VAE--based representations:}For \textbf{F+MNIST}, we used the encoder and decoder from \citet{burgess2017understanding}; for \textbf{CheXpert} we used the encoder and decoder from \citet{higgins2017betavae}; for \textbf{CIFAR-10+100} we used the ResNet--VAE used for the CIFAR-10 dataset in \citet{kingma2016improved}.

\paragraph{Encoders for SimCLRv2--based representations:} For \textbf{F+MNIST} and \textbf{CheXpert} we used the ResNet18 architecture \citep{resnet18}; for \textbf{CIFAR-10+100} we used the ResNet50 architecture \citep{resnet18}.

\paragraph{Prediction heads:} For all our experiments we used a random forest prediction head with 250 trees for \textbf{F+MNIST} and \textbf{CheXpert} and 1000 trees for \textbf{CIFAR-10+100}. We ablate with different prediction heads in Section \ref{app:diff_ph}.
\subsubsection{Active learning}
\label{appsub:active_learning}

\paragraph{Initial labelled set:}To create our initial labelled set, we randomly sampled 2 labels from ``0'', ``1'', and ``2'' for \textbf{F+MNIST}; for \textbf{CheXpert} we randomly sampled 4 labels from ``0'' and ``1''; for \textbf{CIFAR-10+100} we randomly sampled 2 classes for each of our target classes and ``0'' class.

\paragraph{Labelling budget:} We chose our labelling budget based on the number of data points at which our approach plateaued. For \textbf{F+MNIST} we chose a budget of 500 labels; for \textbf{CheXpert} we chose a budget of 6,000 labels; for \textbf{CIFAR-10+100} we chose a budget of 10,000 labels.
\subsubsection{Acquisition strategies}
\label{appsub:acquisition_strategies}

\looseness=-1
\paragraph{Acquisition strategies:}For our acquisition strategies, we used {EPIG}, {BALD}, {Confidence Sampling} and the acquisition strategies from our baselines methods which we describe below. To estimate BALD and EPIG, we used the same setup as \citet{epig}: we used 100 realisations of
$\theta_h$ (for random forests this meant using the individual trees; otherwise this meant sampling from the parameter distribution). For EPIG, we used $M$ samples of $x_*$ from a finite set of unlabelled inputs representative of the downstream task, where $M=500$ for \textbf{CheXpert} and \textbf{F+MNIST}, and $M=1000$ for \textbf{CIFAR-10+100}.

\looseness=-1
\paragraph{Batch acquisition:}  We used batch acquisitions for all datasets, with a batch size of 10 for \textbf{F+MNIST} and 100 for the others. We used the 
``power'' batch acquisition strategy from~\cite{SupervisedAL6} with $\beta=4$ for \textbf{F+MNIST} and $\beta=8$ otherwise. We make this choice as this strategy is both highly scalable and has been shown to give performance comparable to more sophisticated batch acquisition strategies.

\subsubsection{Baselines}
\label{appsub:baselines}

\looseness=-1
We used baselines which were specifically designed for scenarios with class imbalance and redundancy (\textbf{GALAXY} \citep{GALAXY}, \textbf{SIMILAR} \citep{SIMILAR}) or which have shown strong performance in such settings (\textbf{Cluster Margin} \citep{ClusterMargin} from the results in \citet{GALAXY}). We note that \textbf{Confidence Sampling} \citep{confSampling}, though not treated as a baseline, has also shown strong performance in this setting (as per the  results in \citet{GALAXY}).

\looseness=-1
\paragraph{Acquisition strategies:} For \textbf{SIMILAR}, we used the FLQMI relaxation of the submodular mutual information (SMI) due to memory constraints posed by the FLCMI relaxation. For \textbf{Cluster Margin}, we first experimented with their original hyperparameters ($\epsilon$ such that we have at least 10 clusters and $k_m=10k_t$). We found this to result in poor performance and so instead adopted the hyperparameters used in \citet{GALAXY}, i.e. choosing $\epsilon$ such that we have exactly 50 clusters and $k_m=1.25k_t$, where $k_t$ is our acquisition batch size.

\looseness=-1
\paragraph{Models:}For a fair comparison with our approach, we used a ResNet18 model for all the baselines on \textbf{F+MNIST} and \textbf{CheXpert}, and a ResNet50 for \textbf{CIFAR-10+100}. We added a final fully--connected hidden layer of 128 hidden units on top of the ResNet models and trained them in a fully supervised fashion. For \textbf{Cluster Margin}, we followed \citet{ClusterMargin} and warm--started the models by training them on a validation set (0.5\% of our pool size) that was evenly balanced between the target classes and ``other'' class;\footnote{Note that \textbf{Cluster Margin} still performs worse than our approach despite the unrealistic assumption of a validation set.} for \textbf{SIMILAR}, we followed \citet{SIMILAR} and trained the models fully from scratch; for \textbf{GALAXY}, we followed \citet{GALAXY} and initialised the ResNet backbones with weights pretrained on ImageNet in a fully supervised fashion. We trained the models using a batch size of 200, the Adam optimiser, and a learning rate of 0.001 for \textbf{Cluster Margin}, \textbf{SIMILAR} and 0.0001 for \textbf{GALAXY}.

\subsection{Details for Section \ref{sec:foundation_models}}
\label{apsub:foundation_models}

\looseness=-1
For the \textbf{TRANSFER} approach, we replaced our SimCLRv2 encoders pretrained on the pool with: ResNet50 and ResNet101 \citep{resnet18} pretrained on ImageNet in a fully supervised fashion, and DinoV2--Small and DinoV2--Big \citep{oquab2023dinov2} pretrained in a self--supervised fashion on the curated dataset discussed in \citet{oquab2023dinov2}\footnote{The ResNet models were accessed through TorchVision and the DinoV2 models were accessed from the PyTorch Hub \citep{torch}. Specifically, DinoV2--Small corresponds to \url{facebookresearch/dinov2/dinov2_vits14} and DinoV2--Big corresponds to \url{facebookresearch/dinov2/dinov2_vitb14}.}.

\looseness=-1
\paragraph{ResNet models:}For the pretrained ResNet models, we adopted the same finetuning setup as Section \ref{appsub:representations}, changing only the learning rate to 0.0008 as this resulted in more stable training. We used the same augmentation as in Section \ref{appsub:representations}.

\looseness=-1
\paragraph{DinoV2 models:} Similar to \cite{oquab2023dinov2}, we followed a similar finetuning setup to the one in \citet{touvron2022deit}. Specifically, we used a batch size of 512, the Adam optimiser, learning rate of $3 \times 10^{-4}$ with 5 epochs warmup and cosine decay, and label smoothing with level 0.1. We used the same augmentations in Section \ref{appsub:representations}.

\subsection{Details for Section \ref{sec:shortfals}}
\label{apsub:shortfall}

\begin{table*}[!ht]
\caption{Imbalance ratio corresponding to the different levels of messiness used in Section \ref{sec:shortfals}.
}
\vspace{-4pt}
\label{tab:messiness_levels}
\centering
\footnotesize
\begin{tabular}{c|c}
\hline
\textbf{Dataset} & Imbalance ratio \\
\hline
\text{Low messiness} & 2  \\
\text{Medium messiness} & 10  \\
\text{High messiness} & 150  \\
\hline
\end{tabular}
\end{table*}

\looseness=-1
To demonstrate that unsupervised representations can break down in the presence of progressively messier pools, we first pretrained unsupervised encoders on pools with different levels of messiness, then, using the representations from these encoders, performed active learning on a pool with the same target/redundant classes and no class imbalance. We do not include imbalance in the pool used for active learning as this allows us to fairly judge the effect of unsupervised representations on active learning performance. Below, we describe the different levels of messiness used for the different datasets and the unsupervised encoders.

\looseness=-1
\paragraph{Datasets:}We used the same pools described in Section \ref{appsub:datasets} for \textbf{F+MNIST} and \textbf{CIFAR-10+100}. We varied the level of messiness by changing the amount of imbalance present in our pool between our target and redundant classes. The imbalance ratios are shown in Table \ref{tab:messiness_levels}.

\looseness=-1
\paragraph{Unsupervised encoders:}We followed Sections \ref{appsub:representations}, \ref{appsub:models} and trained unsupervised VAE encoders for \textbf{F+MNIST} and unsupervised SimCLRv2 encoders for \textbf{CIFAR-10+100}. 

\section{Additional Plots}
\label{app:additional_plots}

\subsection{Full Active Learning Curves for Table \ref{tab:comparisons}}
\label{subapp:tab_1_fig}
\begin{figure*}[!ht] 
\centering 
\includegraphics[width=0.7\textwidth]{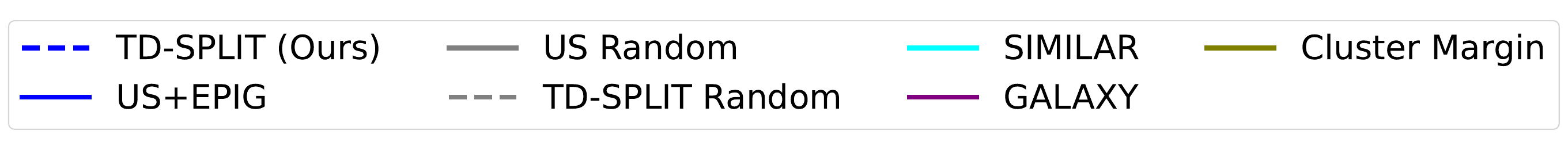}
\begin{subfigure}[b]{0.32\textwidth} 
\centering \includegraphics[width=\linewidth]{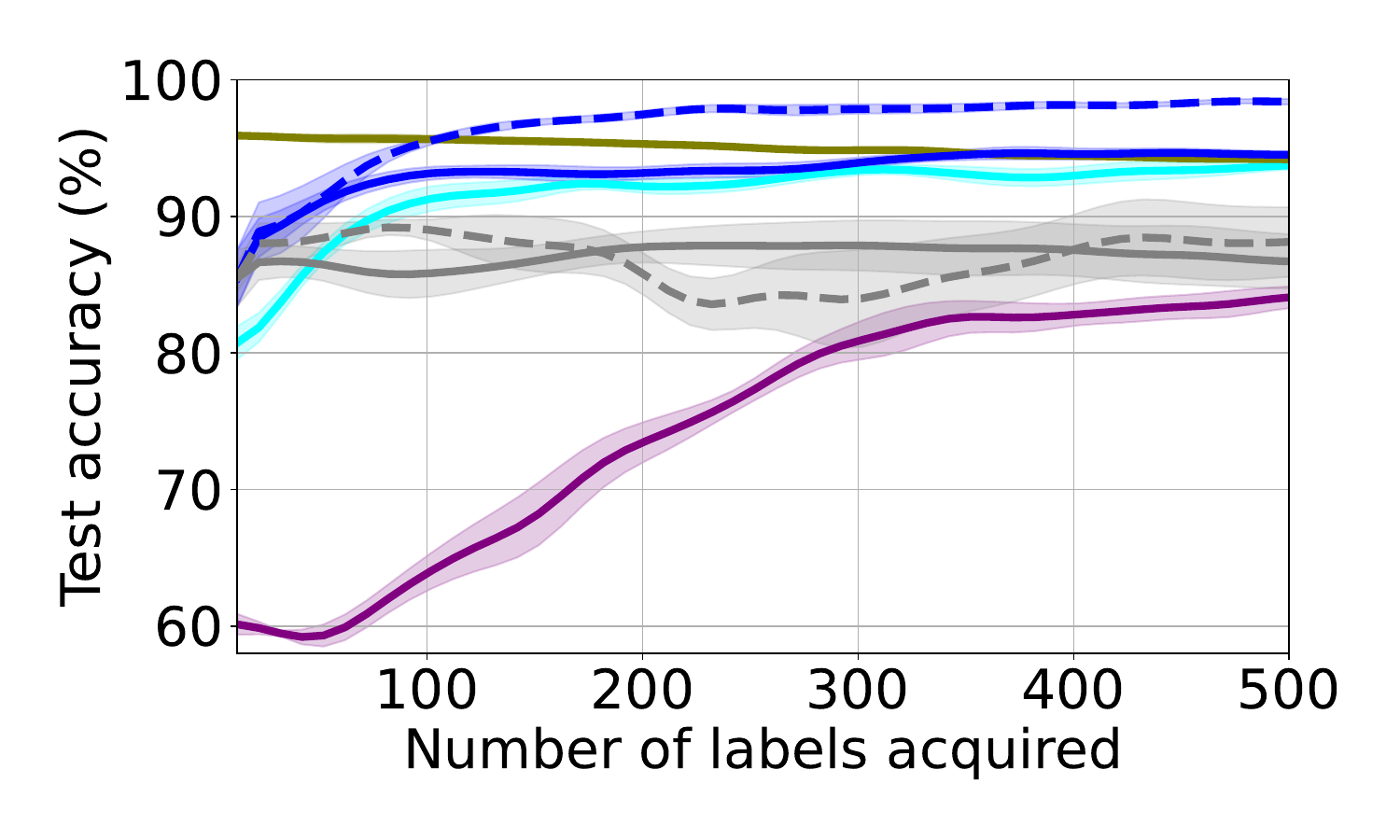} 
\caption{\textbf{F+MNIST}} \label{fig:fmnist_split_baselines} 
\end{subfigure} 
\begin{subfigure}[b]{0.32\textwidth} 
\centering 
\includegraphics[width=\linewidth]{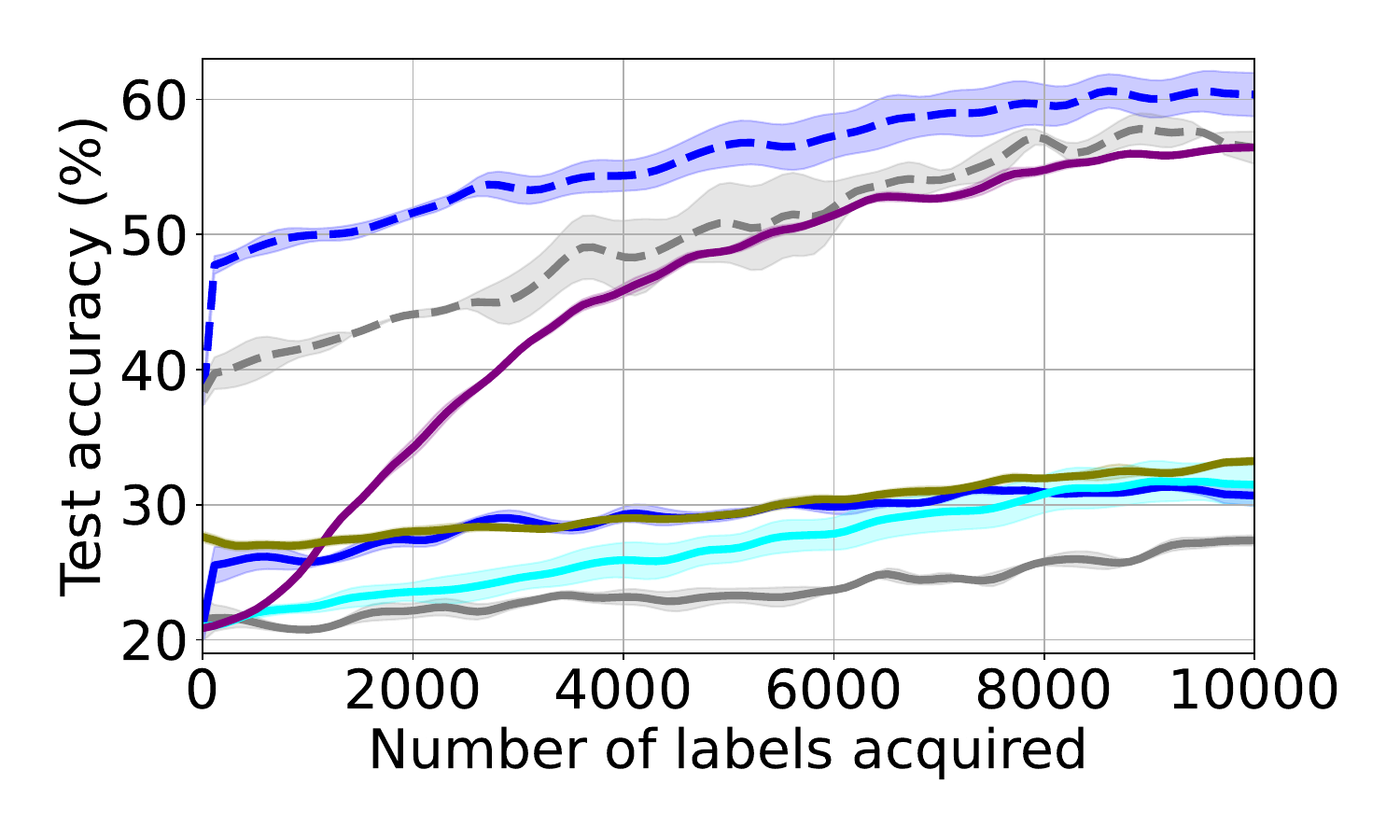} 
\caption{\textbf{CIFAR-10+100}} \label{fig:cifar_split_baselines} 
\end{subfigure} 
\begin{subfigure}[b]{0.32\textwidth} 
\centering \includegraphics[width=\linewidth]{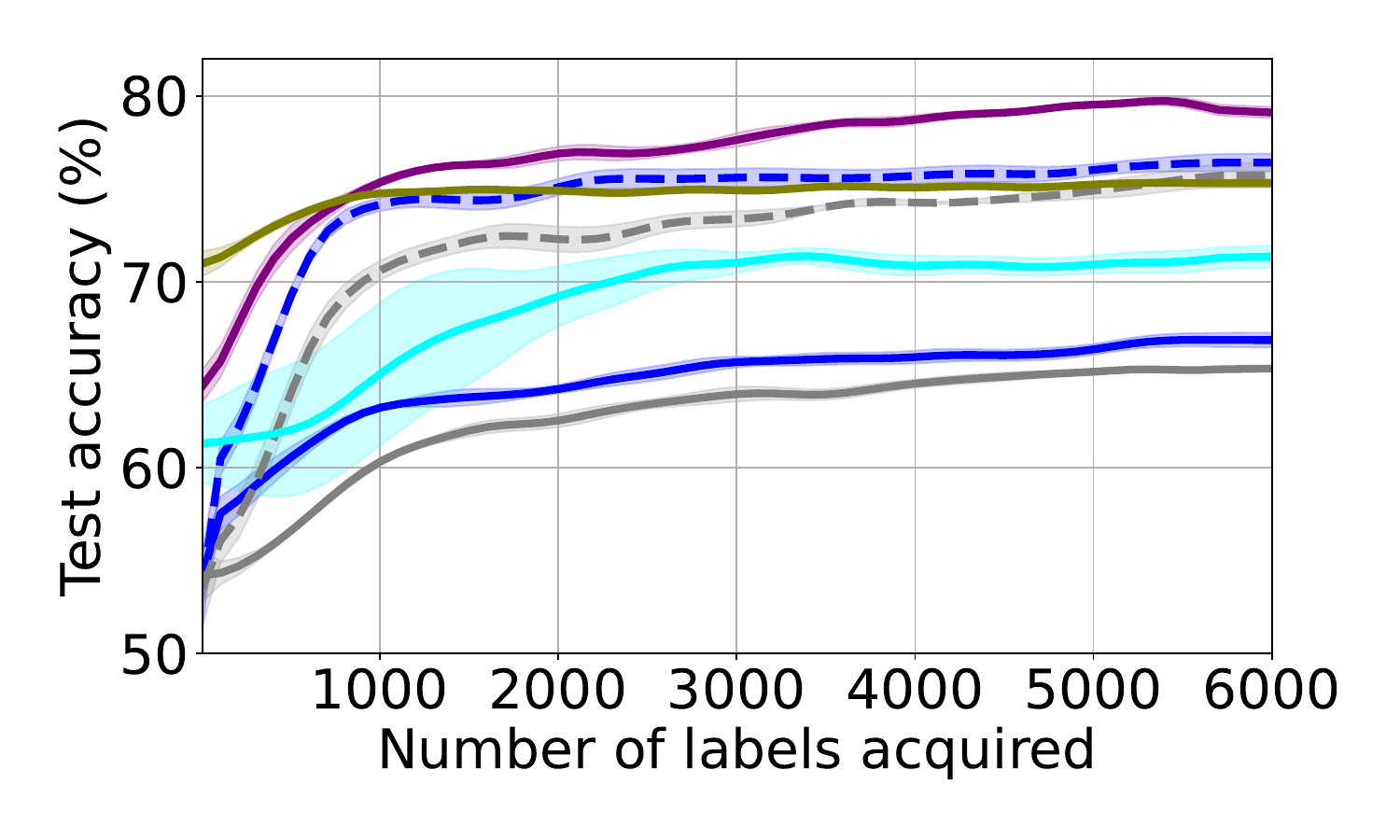} 
\caption{\textbf{CheXpert}} \label{fig:chex_split_baselines} 
\end{subfigure} \vspace{1em} 
\vspace{-0.6cm}
\caption{Test accuracy for our \
textbf{TD-SPLIT} approach on \textbf{F+MNIST}, \textbf{CIFAR-10+100}, \textbf{CheXpert} and the baselines considered in Table \ref{tab:comparisons}. All experiments were run for 4 seeds. Solid line shows mean and shading $\pm 1$ standard error.}
\label{fig:split_baselines} 
\end{figure*}

\begin{figure*}[!ht] 
\centering 
\includegraphics[width=0.7\textwidth]{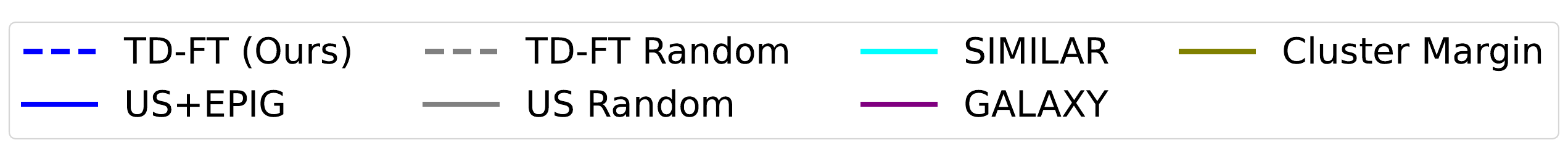}
\begin{subfigure}[b]{0.32\textwidth} 
\centering \includegraphics[width=\linewidth]{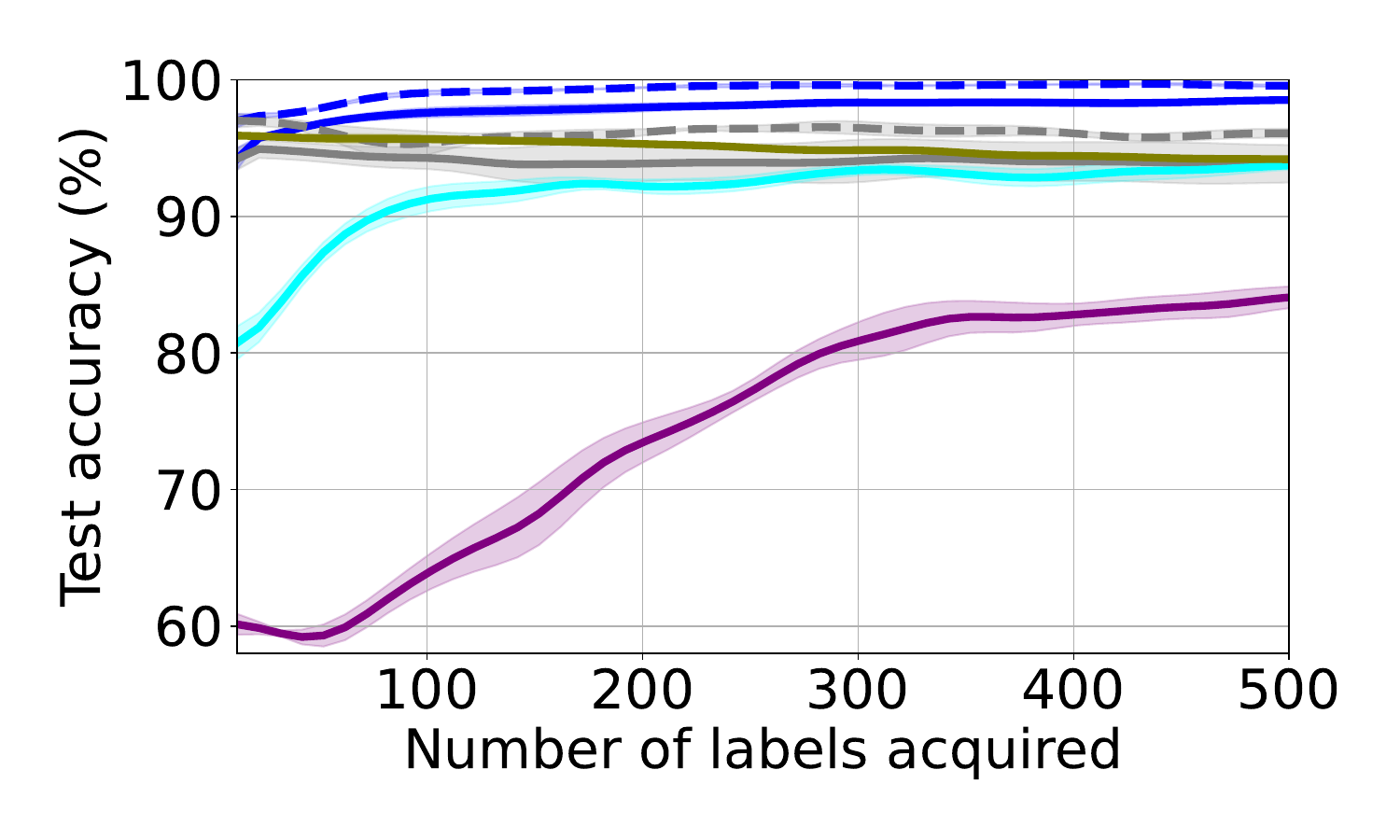} 
\caption{\textbf{F+MNIST}} \label{fig:fmnist_ft_baselines} 
\end{subfigure} 
\begin{subfigure}[b]{0.32\textwidth} 
\centering 
\includegraphics[width=\linewidth]{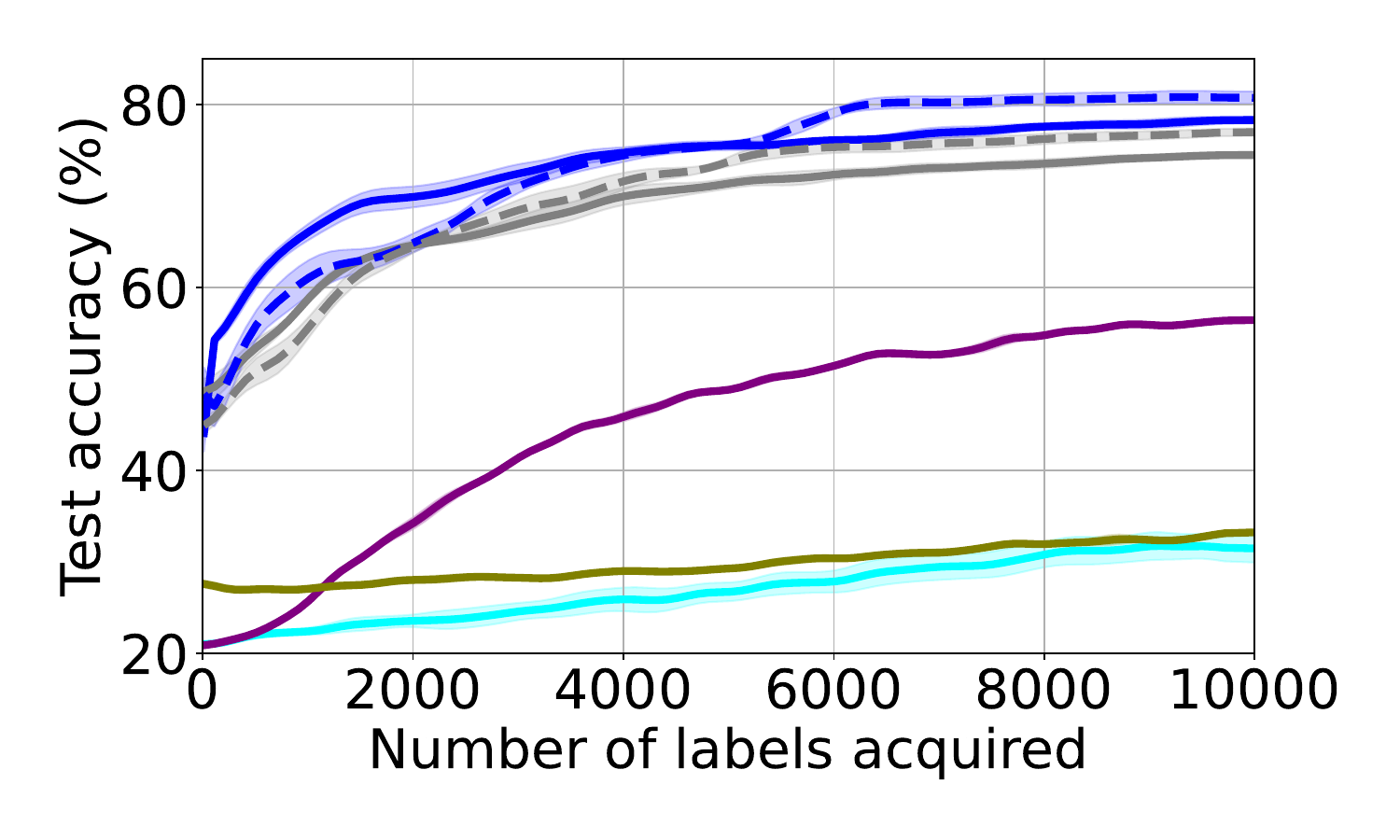} 
\caption{\textbf{CIFAR-10+100}} \label{fig:cifar_ft_baselines} 
\end{subfigure} 
\begin{subfigure}[b]{0.32\textwidth} 
\centering \includegraphics[width=\linewidth]{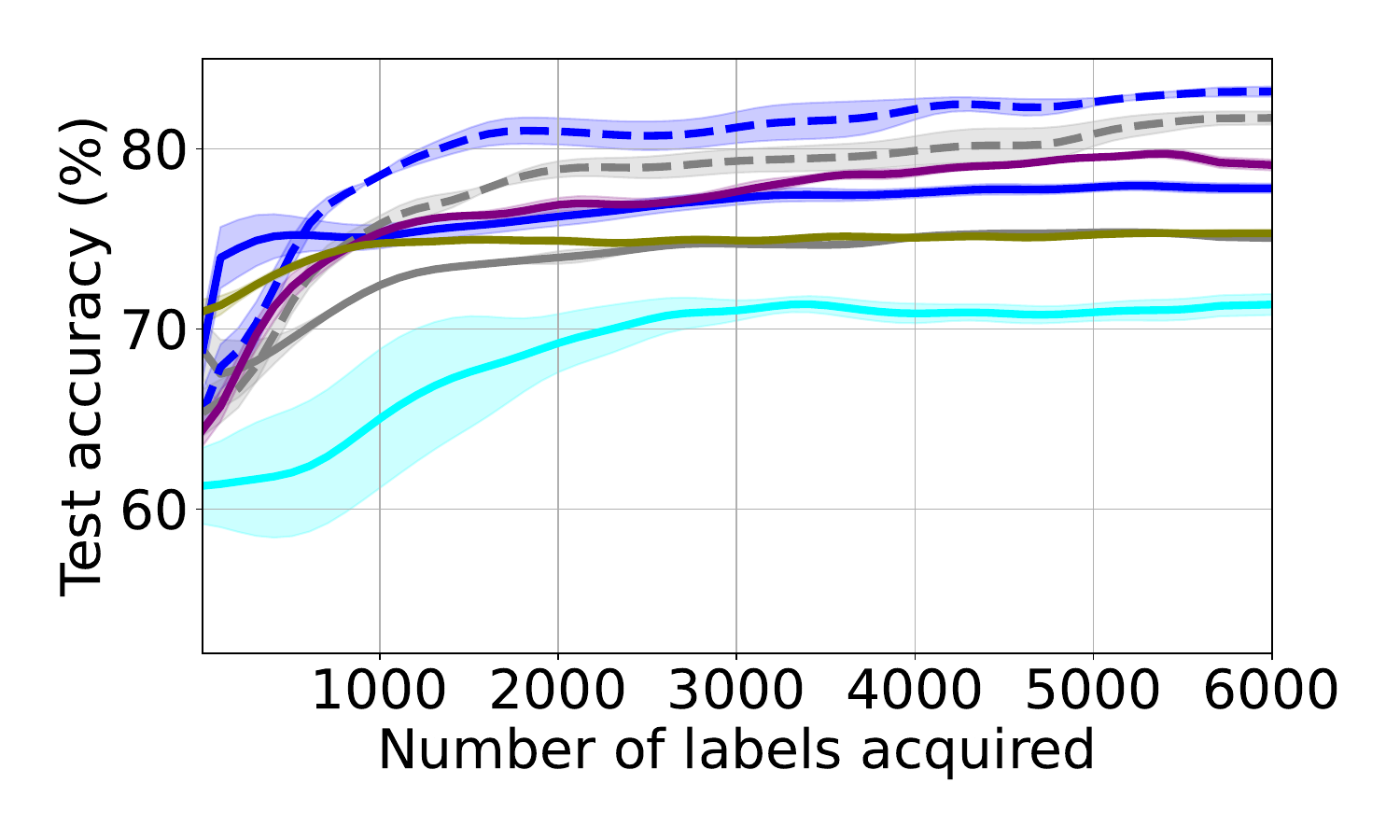} 
\caption{\textbf{CheXpert}} \label{fig:chex_ft_baselines} 
\end{subfigure} \vspace{1em} 
\vspace{-0.6cm}
\caption{Test accuracy for our \textbf{TD-FT} approach on \textbf{F+MNIST}, \textbf{CIFAR-10+100}, \textbf{CheXpert} and the baselines considered in Table \ref{tab:comparisons}. All experiments were run for 4 seeds. Solid line shows mean and shading $\pm 1$ standard error.}
\label{fig:ft_baselines} 
\end{figure*}

Figure \ref{fig:split_baselines} shows the full active learning curves for our \textbf{TD-SPLIT} approach and the baselines considered in Table \ref{tab:comparisons}. Similarly, Figure \ref{fig:ft_baselines} shows the curves for our \textbf{TD-FT} approach.

\section{Ablations}
\label{app:ablations}

We ran all of our ablations on the \textbf{F+MNIST} dataset, focusing on the {EPIG} acquisition function and comparing our \textbf{TD-SPLIT} and \textbf{TD-FT} approach to the \textbf{US} approach. We ablated with different levels of messiness, different retraining periods and different prediction heads.

\subsection{Different Levels of Messiness}
\label{app:diff_level_messiness}

We investigated the performance of our approach for different levels of messiness by varying a) the amount of imbalance between our target class and redundant classes b) varying the proportion of target to redundant classes.

\subsubsection{Different Levels of Imbalance}
\label{app:diff_level_imb}

\begin{figure*}[!ht]
\centering
\begingroup
\captionsetup{skip=2pt}
\captionsetup[sub]{skip=2pt}
\resizebox{0.85\textwidth}{!}{
\begin{minipage}{\textwidth}
\begin{subfigure}[b]{0.32\textwidth}
  \centering\includegraphics[width=\linewidth]{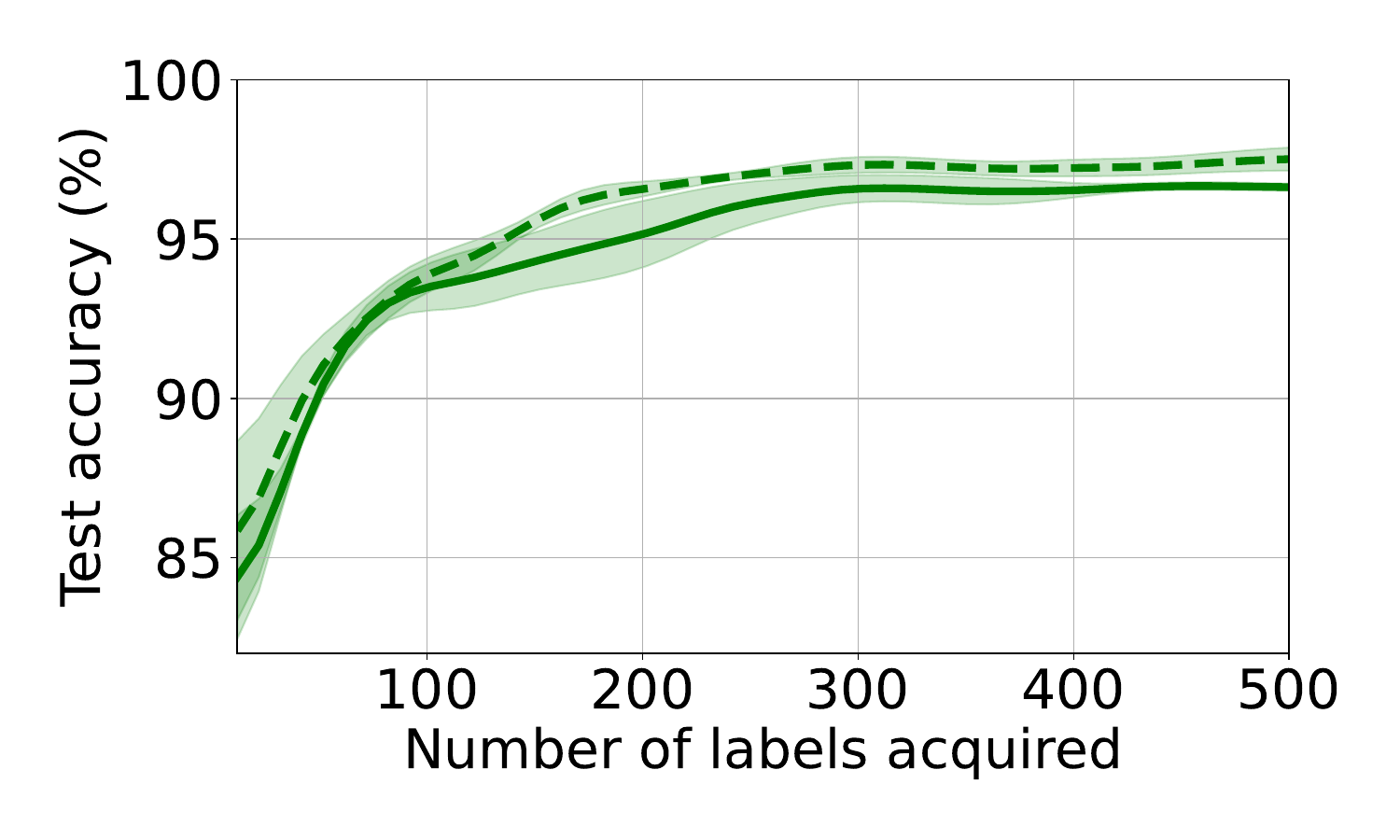}
  \caption{IR=2}\label{fig:sub1}
\end{subfigure}\hfill
\begin{subfigure}[b]{0.32\textwidth}
  \centering\includegraphics[width=\linewidth]{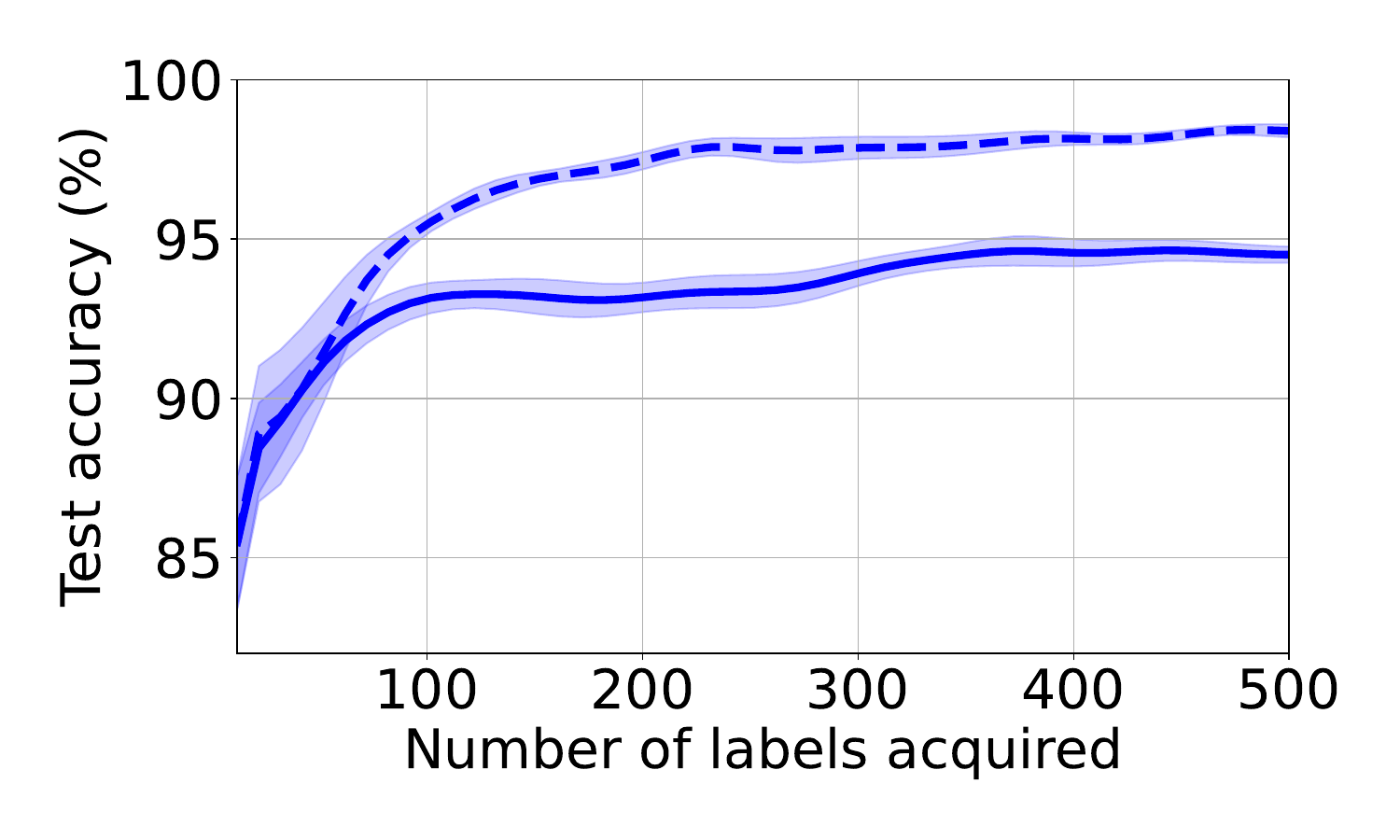}
  \caption{IR=10}\label{fig:sub2}
\end{subfigure}\hfill
\begin{subfigure}[b]{0.32\textwidth}
  \centering\includegraphics[width=\linewidth]{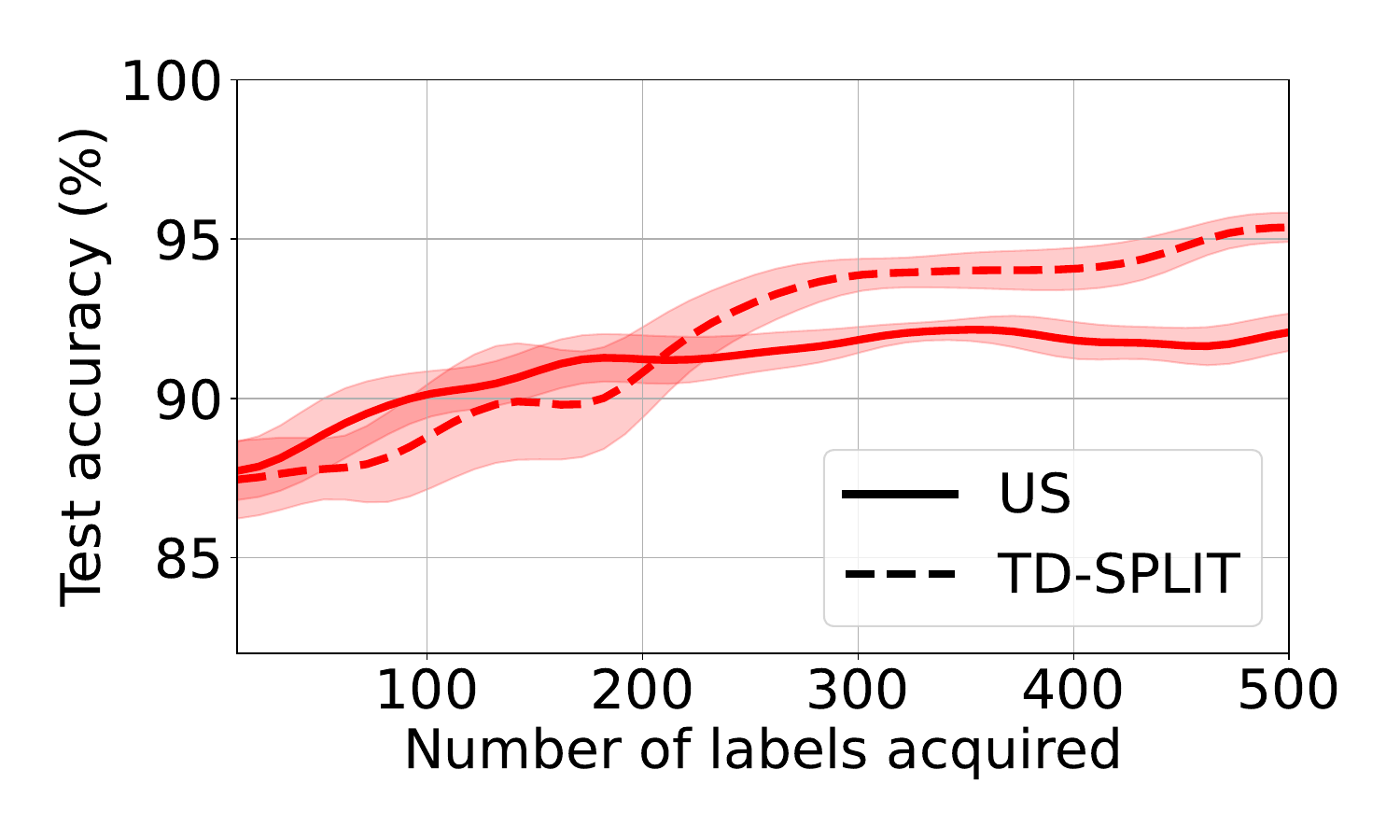}
  \caption{IR=150}\label{fig:sub3}
\end{subfigure}
\begin{subfigure}[b]{0.32\textwidth}
  \centering\includegraphics[width=\linewidth]{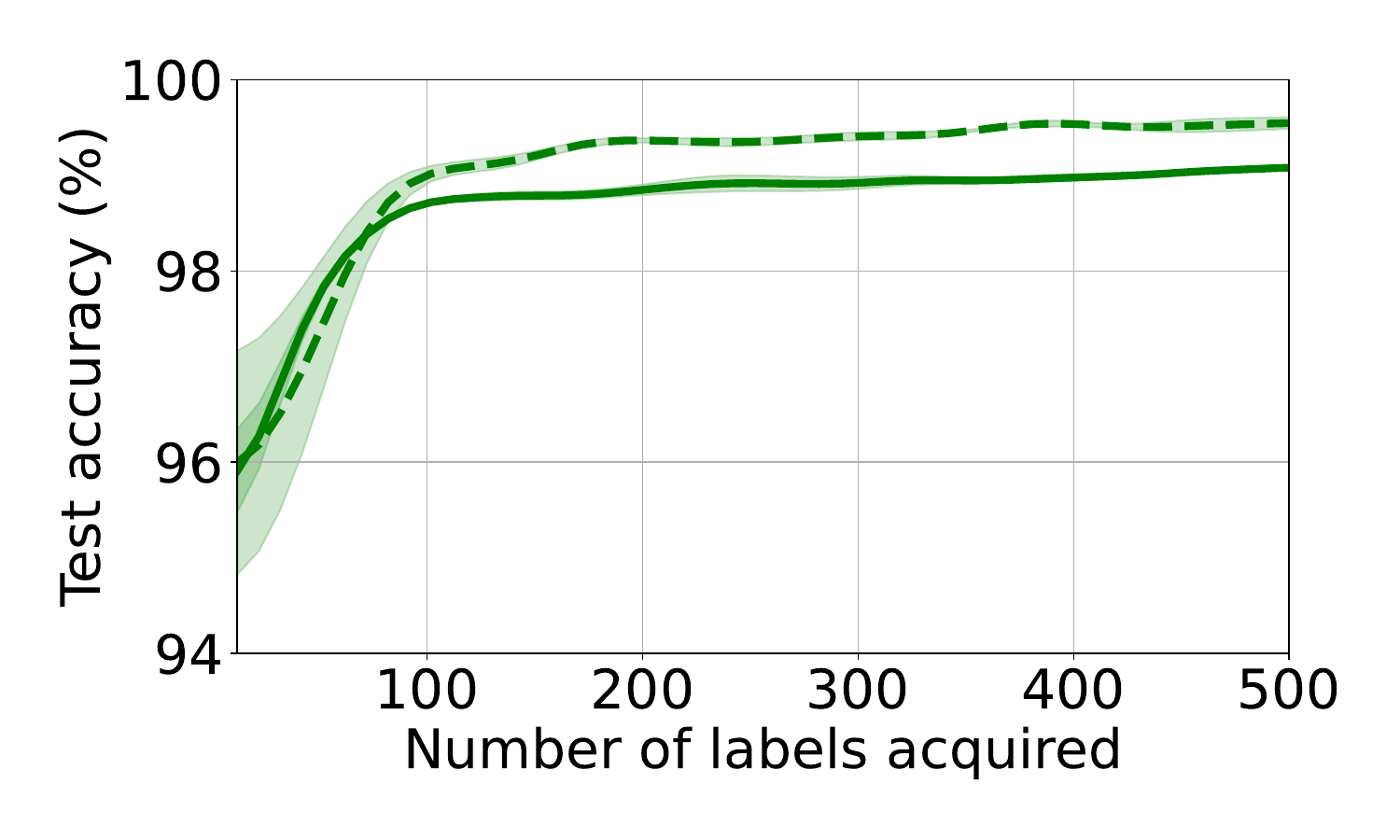}
  \caption{IR=2}\label{fig:sub4}
\end{subfigure}\hfill
\begin{subfigure}[b]{0.32\textwidth}
  \centering\includegraphics[width=\linewidth]{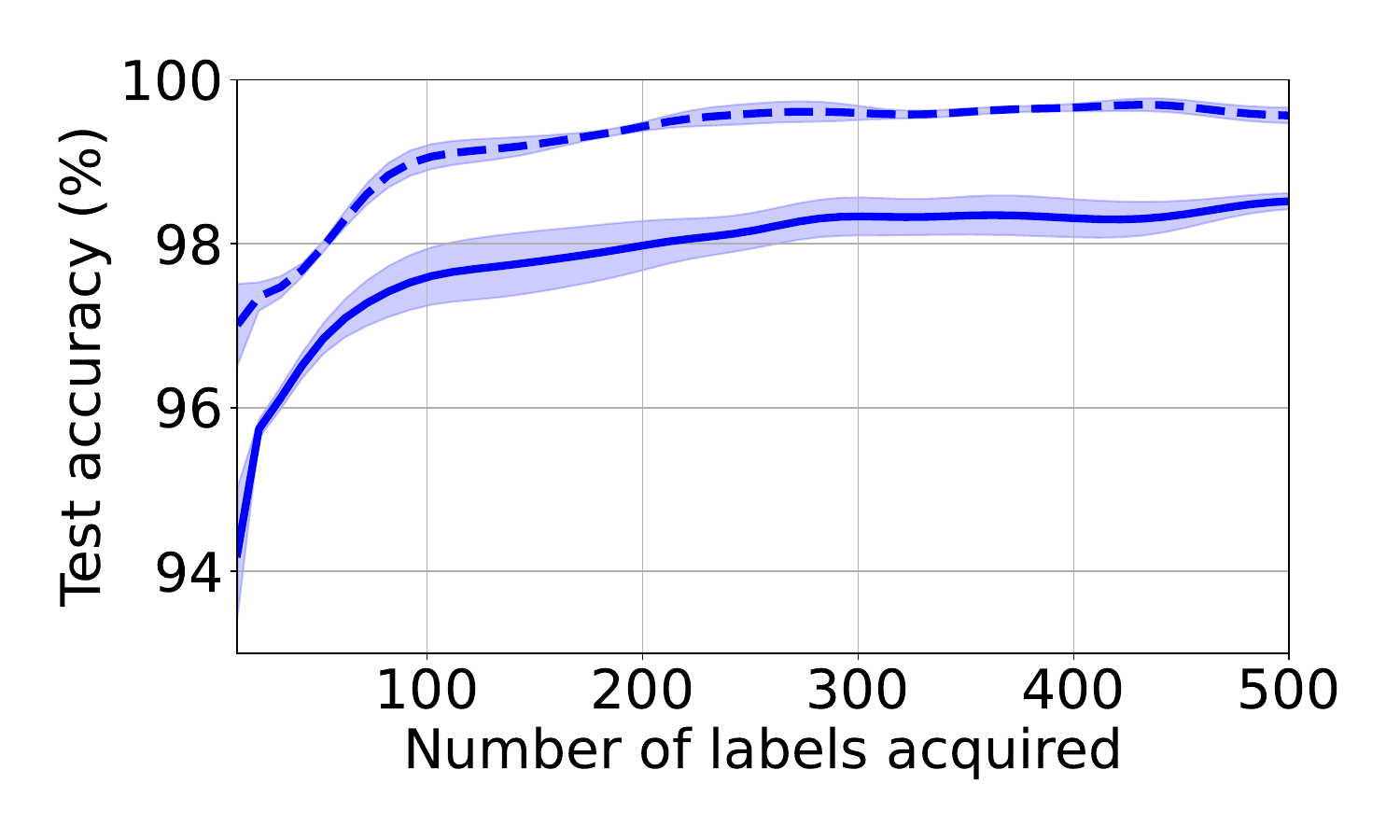}
  \caption{IR=10}\label{fig:sub5}
\end{subfigure}\hfill
\begin{subfigure}[b]{0.32\textwidth}
  \centering\includegraphics[width=\linewidth]{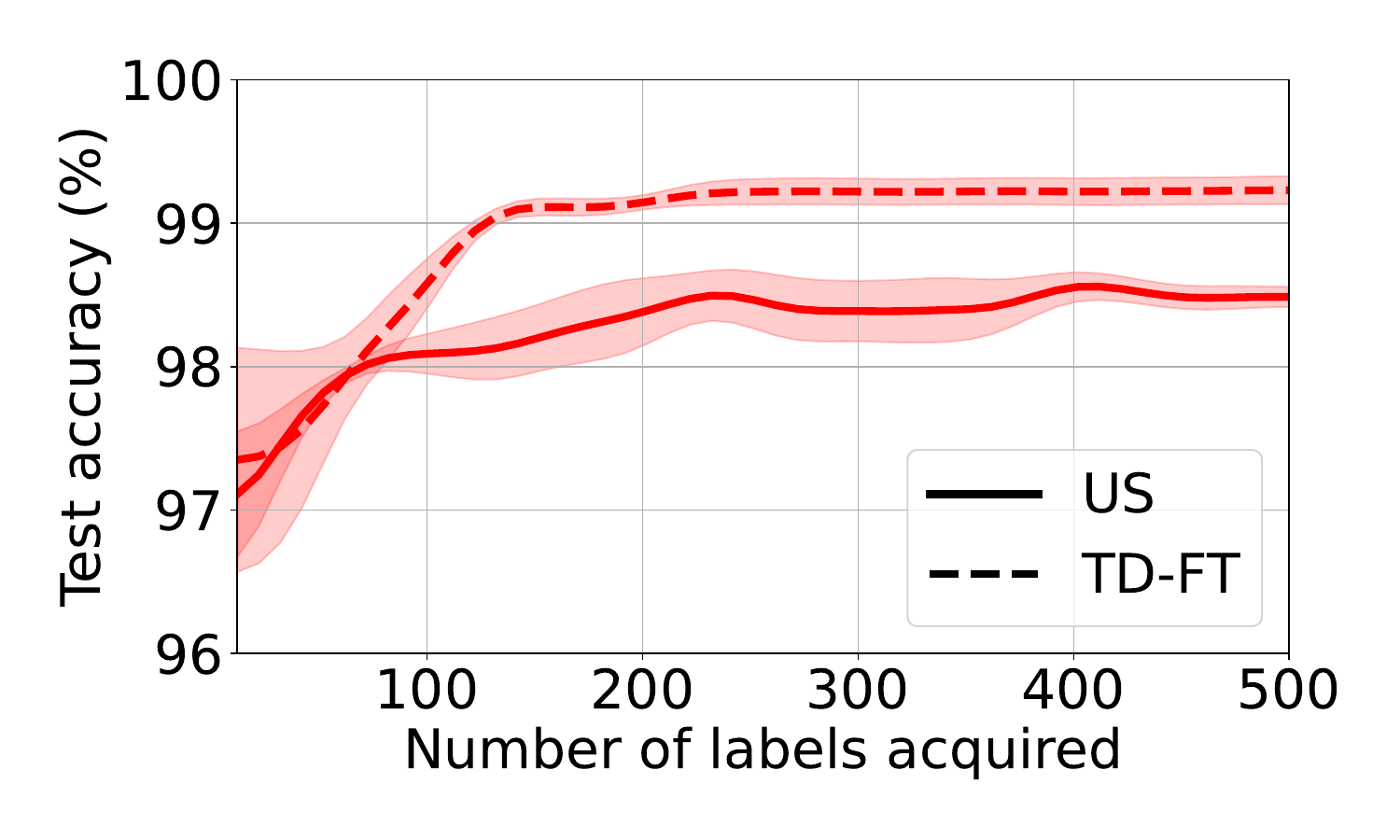}
  \caption{IR=150}\label{fig:sub6}
\end{subfigure}
\end{minipage}
}

\caption{Test accuracy on \textbf{F+MNIST} using the \textbf{US} approach and our task--driven approach for different imbalance ratios (IR) in the pool.
Top row shows the results for \textbf{TD-SPLIT} with VAE encoders and the bottom row shows the results for \textbf{TD-FT} with SimCLRv2 encoders. 
Experiments run for 4 seeds. Solid line shows mean and shading $\pm 1$ standard error.
\vspace{-8pt}}
\label{fig:imb_messiness}
\endgroup
\end{figure*}

Figure \ref{fig:imb_messiness} shows the results of our approach and \textbf{US} for varying levels of imbalance in the pool. We see that our \textbf{TD-SPLIT} and \textbf{TD-FT} approach is robust to different levels of imbalance in the pool when compared with \textbf{US}. In particular, we note that the differences are more significant at higher messiness levels. This is intuitive as this is when we expect to lose the most information about our task from unsupervised representations.

\subsubsection{Different Levels of Redundancy}
\label{app:diff_level_red}

\begin{figure*}[!ht]
\centering
\begingroup
\captionsetup{skip=2pt}
\captionsetup[sub]{skip=2pt}
\resizebox{0.85\textwidth}{!}{
\begin{minipage}{\textwidth}
\begin{subfigure}[b]{0.32\textwidth}
  \centering\includegraphics[width=\linewidth]{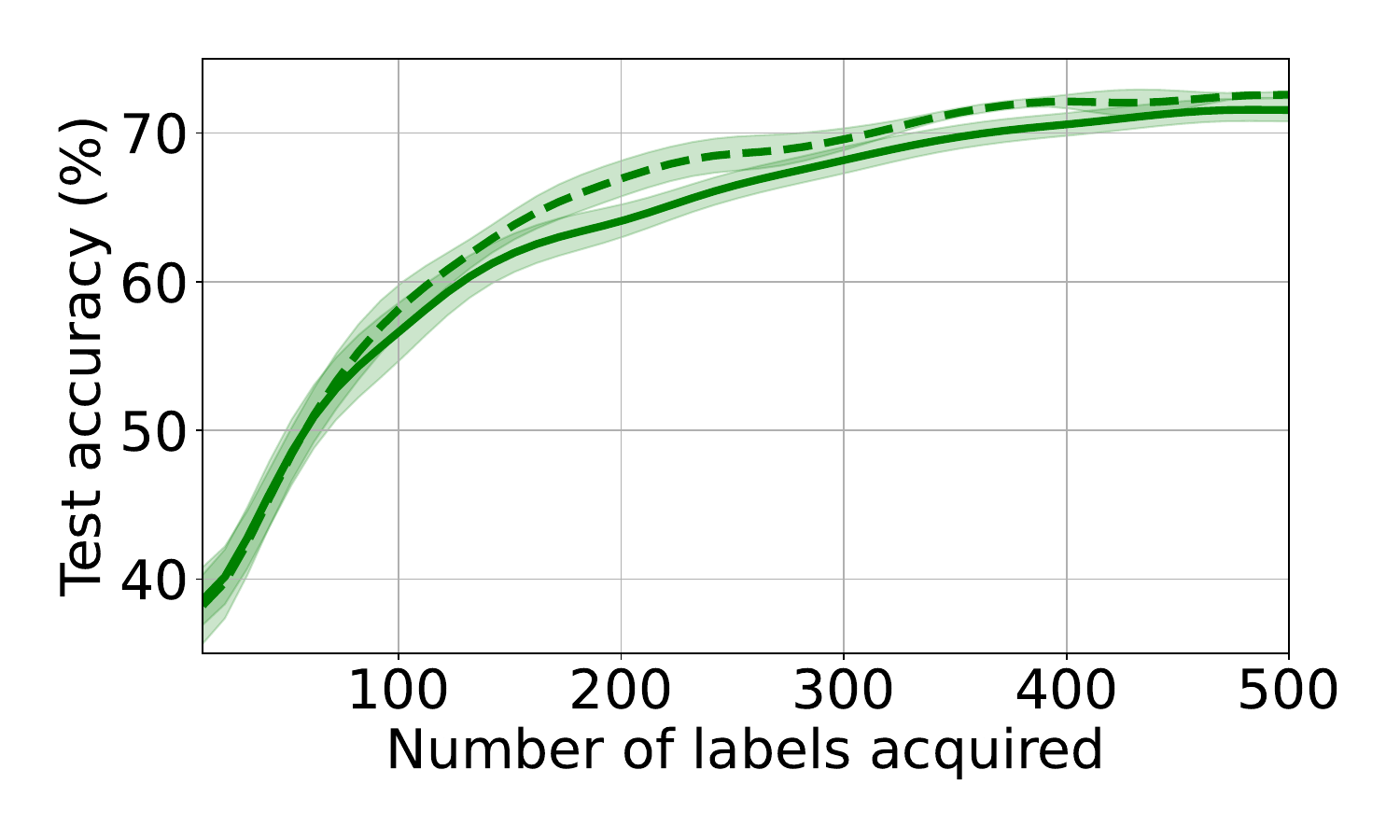}
  \caption{RR=1}\label{fig:_sub1_}
\end{subfigure}\hfill
\begin{subfigure}[b]{0.32\textwidth}
  \centering\includegraphics[width=\linewidth]{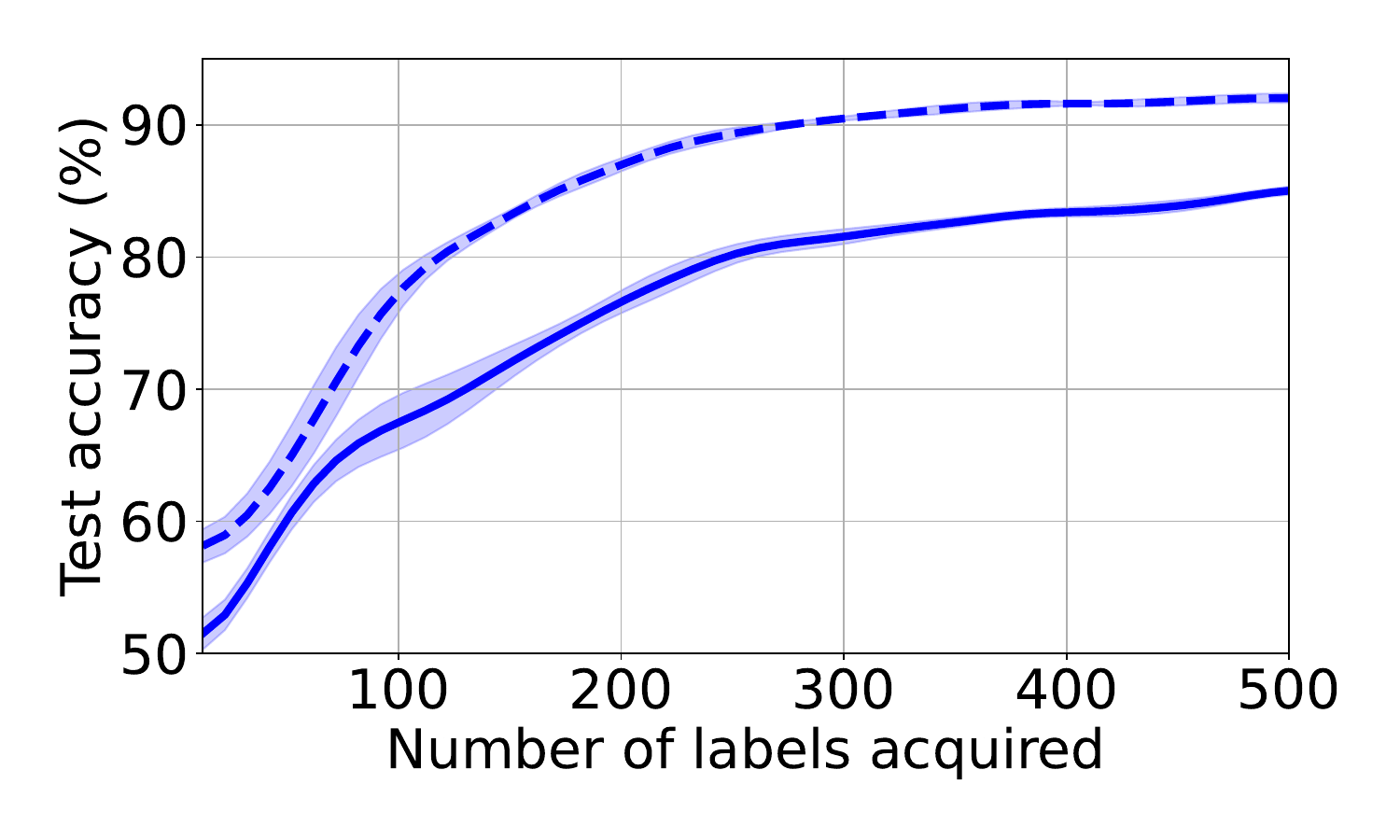}
  \caption{RR=0.5}\label{fig:_sub2_}
\end{subfigure}\hfill
\begin{subfigure}[b]{0.32\textwidth}
  \centering\includegraphics[width=\linewidth]{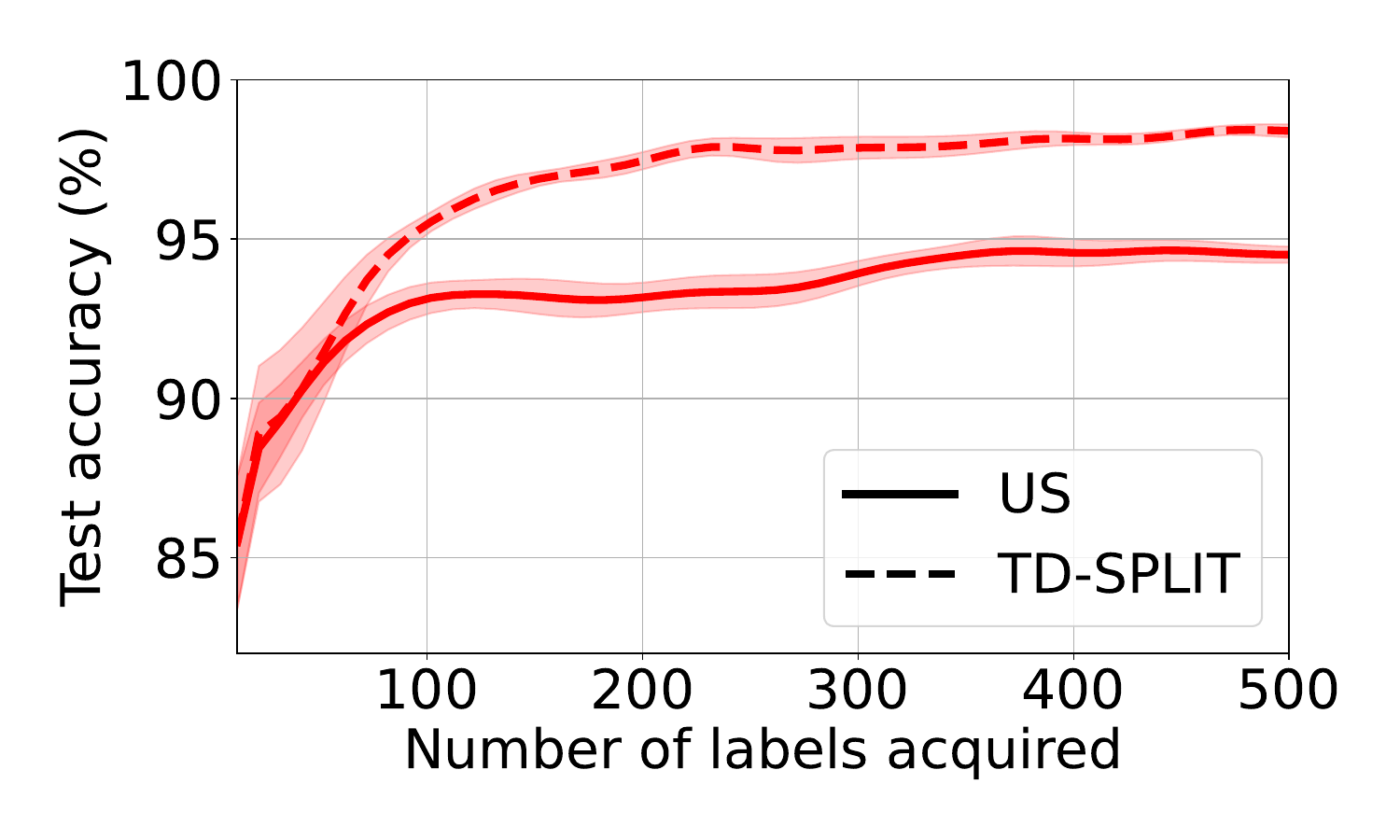}
  \caption{RR=0.17}\label{fig:_sub3_}
\end{subfigure}
\begin{subfigure}[b]{0.32\textwidth}
  \centering\includegraphics[width=\linewidth]{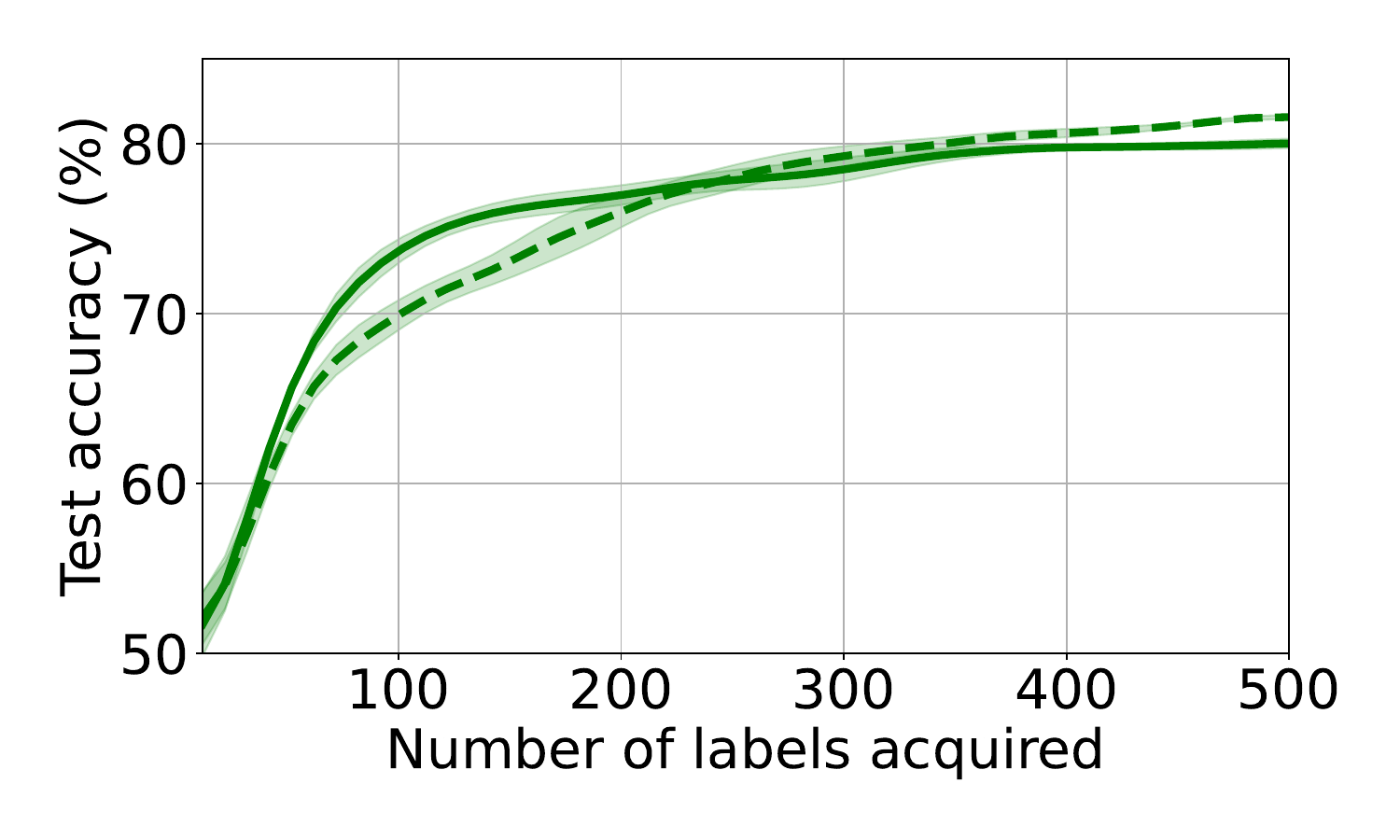}
  \caption{RR=1}\label{fig:_sub4_}
\end{subfigure}\hfill
\begin{subfigure}[b]{0.32\textwidth}
  \centering\includegraphics[width=\linewidth]{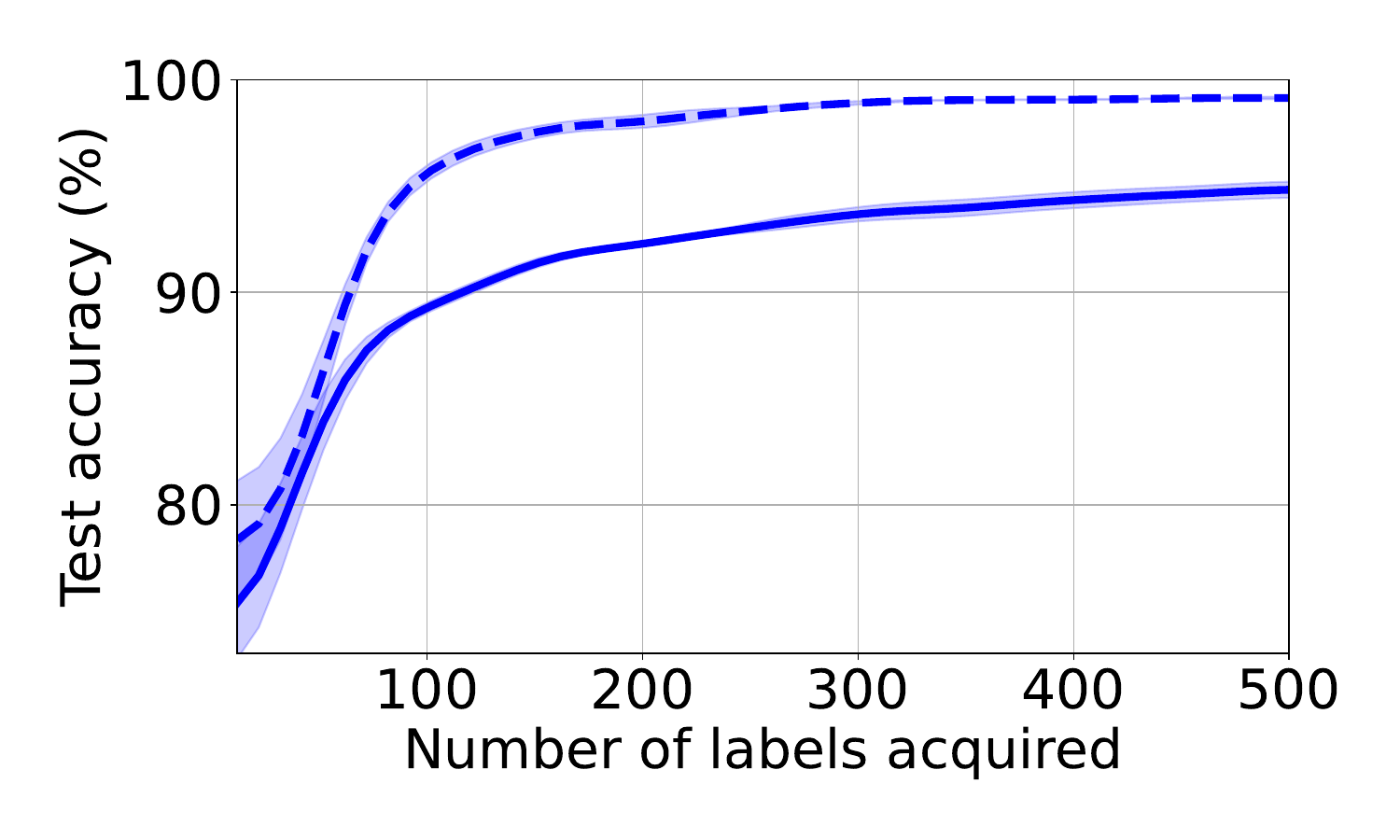}
  \caption{RR=0.5}\label{fig:_sub5_}
\end{subfigure}\hfill
\begin{subfigure}[b]{0.32\textwidth}
  \centering\includegraphics[width=\linewidth]{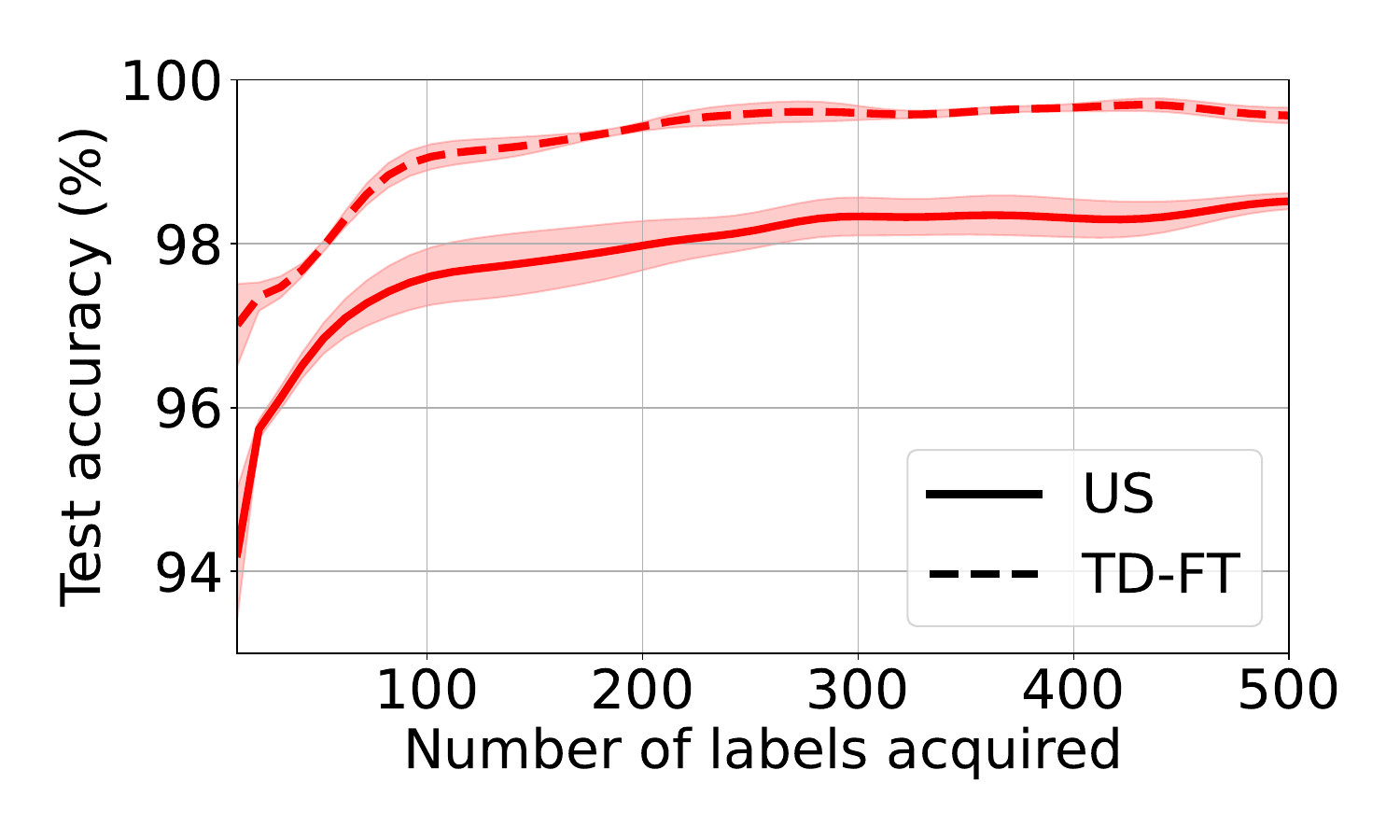}
  \caption{RR=0.17}\label{fig:_sub6_}
\end{subfigure}
\end{minipage}
}

\caption{Test accuracy on \textbf{F+MNIST} using the \textbf{US} approach and our task--driven approach for different redundant ratios (RR). Top row shows the results for \textbf{TD-SPLIT} using VAE encoders and the bottom row shows the results for \textbf{TD-SPLIT} with SimCLRv2 encoders. 
Experiments run for 4 seeds. Solid line shows mean and shading $\pm 1$ standard error.
\vspace{-8pt}}
\label{fig:class_redundancy_messiness}
\endgroup
\end{figure*}

To investigate the impact of redundant classes, we varied the number of target classes by including more/less classes from MNIST. Figure \ref{fig:class_redundancy_messiness} shows the test accuracies for three different redundant ratios (RR), defined as:
\begin{equation}
    \text{RR} = \frac{\text{number of target classes}}{\text{total number of classes}}
\end{equation}

Again, we see that our \textbf{TD-SPLIT} and \textbf{TD-FT} approach is robust to different levels of imbalance in the pool when compared with \textbf{US}, with the differences being more significant at lower RRs.

\subsection{Different Retraining Periods}
\label{app:diff_retraining}

\begin{figure*}[!ht]
\centering
\captionsetup{skip=2pt}
\captionsetup[sub]{skip=2pt}

\begin{subfigure}[b]{0.35\textwidth}
  \centering
  \includegraphics[width=\linewidth]{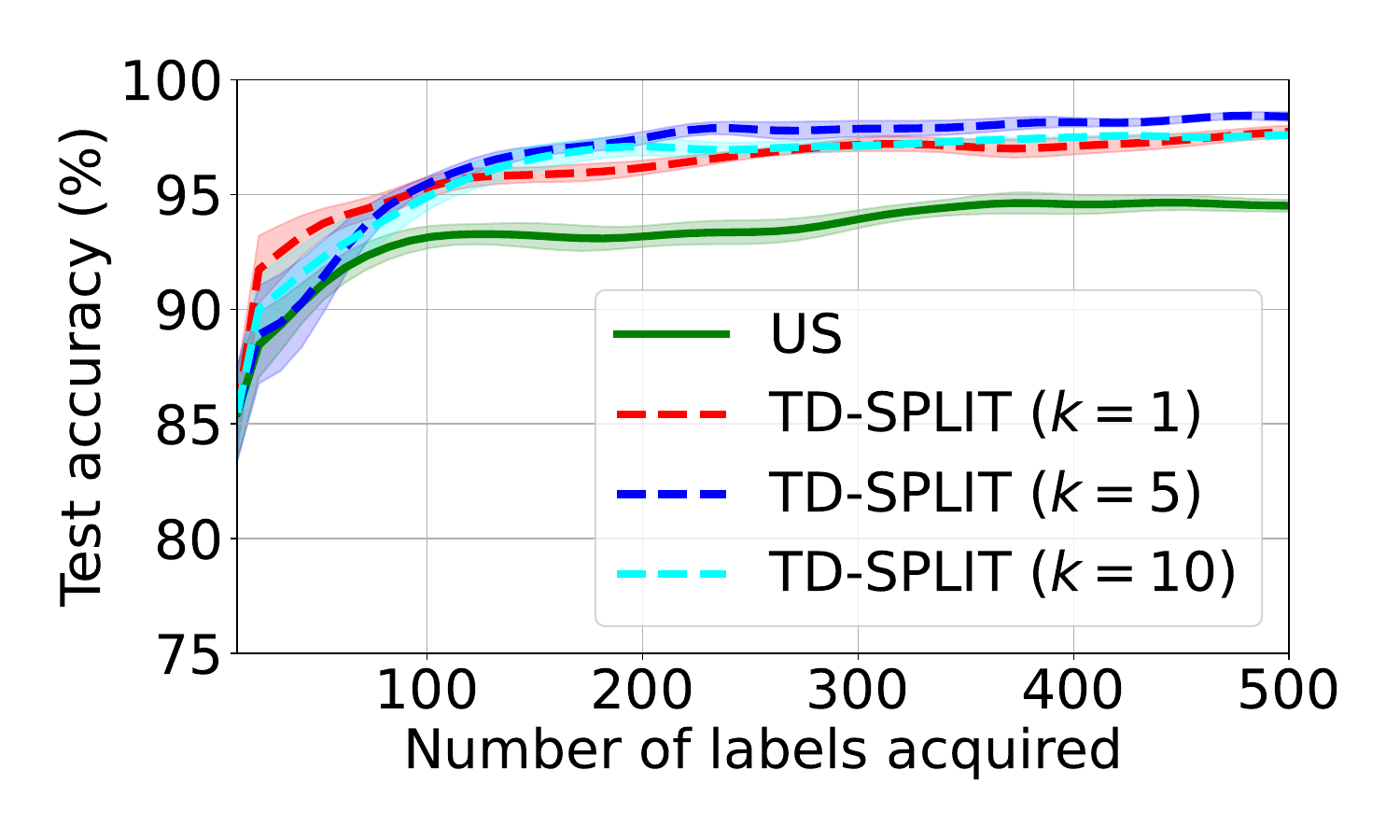}
\end{subfigure}
\begin{subfigure}[b]{0.35\textwidth}
  \centering
  \includegraphics[width=\linewidth]{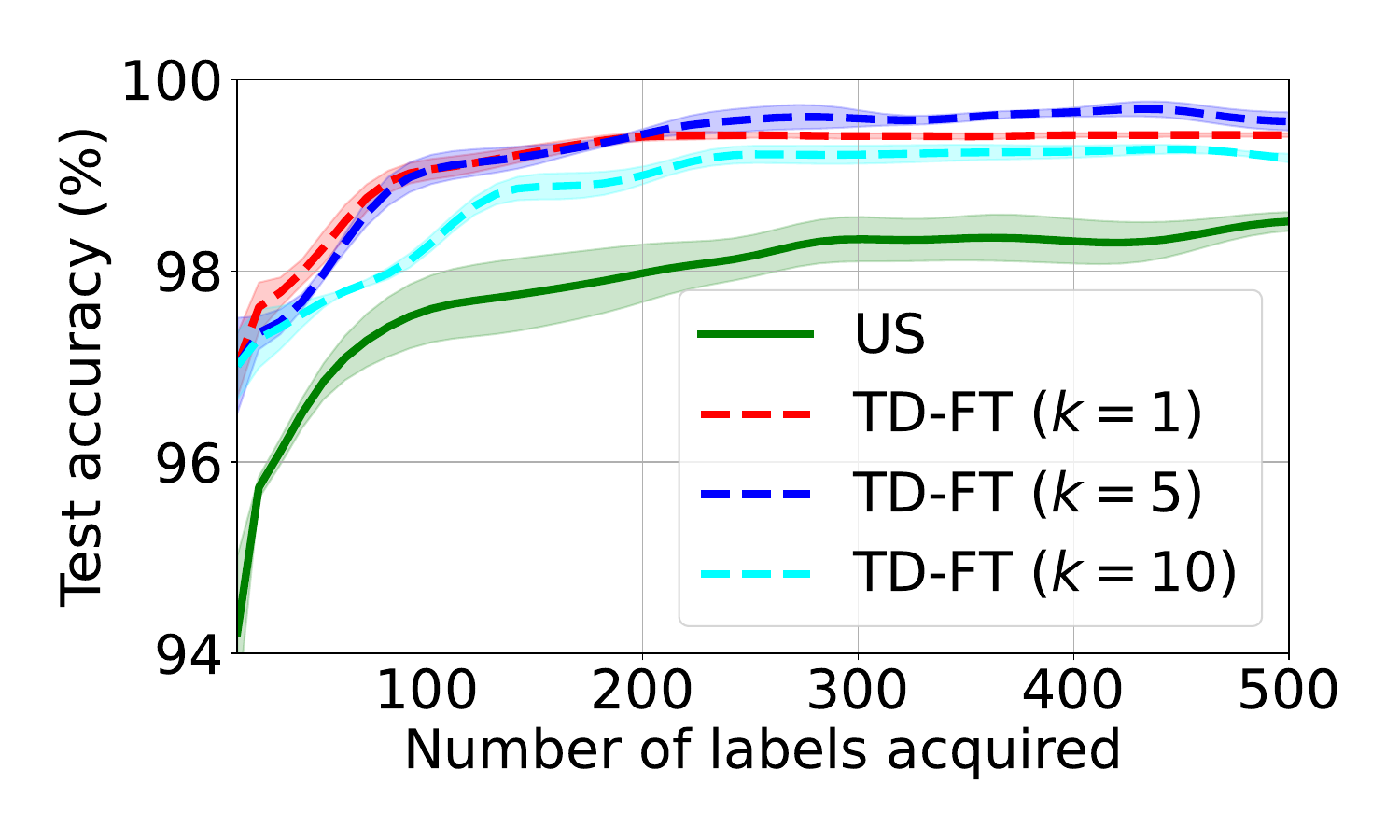}
\end{subfigure}

\caption{Test accuracy on \textbf{F+MNIST} using the \textbf{US} approach and our task--driven approaches for different retraining periods $k$. Left shows the results for \textbf{TD-SPLIT} with VAE encoders and right shows the results for \textbf{TD-FT} with SimCLRv2 encoders. Experiments run for 4 seeds. Solid line shows mean and shading $\pm 1$ standard error.
\vspace{-8pt}}
\label{fig:different_retraining_periods}
\end{figure*}

Figure \ref{fig:different_retraining_periods} shows test accuracies for different retraining periods $k$, where $k$ is the number of acquisition rounds we take before updating our semi--supervised encoder. We see that our approach is also robust to $k$ when compared with \textbf{US}. In particular, we observe that too frequent updates ($k=1$) and too few updates ($k=10$) result in suboptimal performance. This is intuitive as updating too infrequently fails to incorporate information regularly enough from acquired labels to boost later acquisitions, whereas updating too frequently can create a significant mismatch between assumed and actual model updates (see Section \ref{sub:finetuning_approach}), resulting in suboptimal acquisitions.

\subsection{Different Prediction Heads}
\label{app:diff_ph}

To show that our approach is compatible with different prediction heads,  we replaced the random forest prediction head with a 1 layer neural network with 128 hidden units. We used the Laplace approximation \cite{MacKay1992BayesianI} to infer the parameter distribution, where we used a standard Gaussian prior, $\mathcal{N}(0, I)$, and a diagonal, tempered posterior \citep{aitchison2020statistical}, with tempering implemented by raising the likelihood term to a power of $\text{dim}(\theta_h)$ (i.e. the parameter count of the prediction head).

\begin{figure*}[!ht]
\centering
\captionsetup{skip=2pt}
\captionsetup[sub]{skip=2pt}

\begin{subfigure}[b]{0.35\textwidth}
  \centering
  \includegraphics[width=\linewidth]{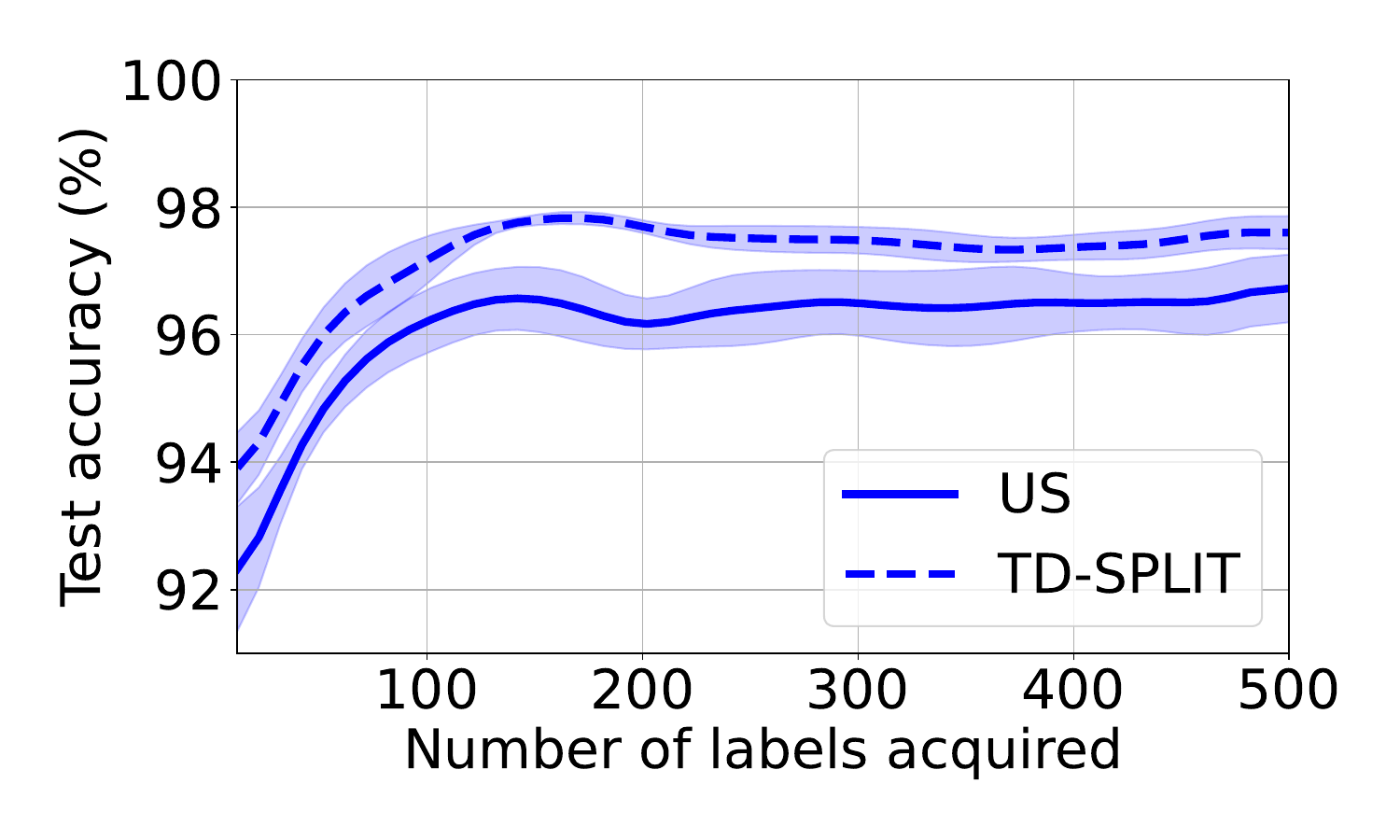}
\end{subfigure}
\begin{subfigure}[b]{0.35\textwidth}
  \centering
  \includegraphics[width=\linewidth]{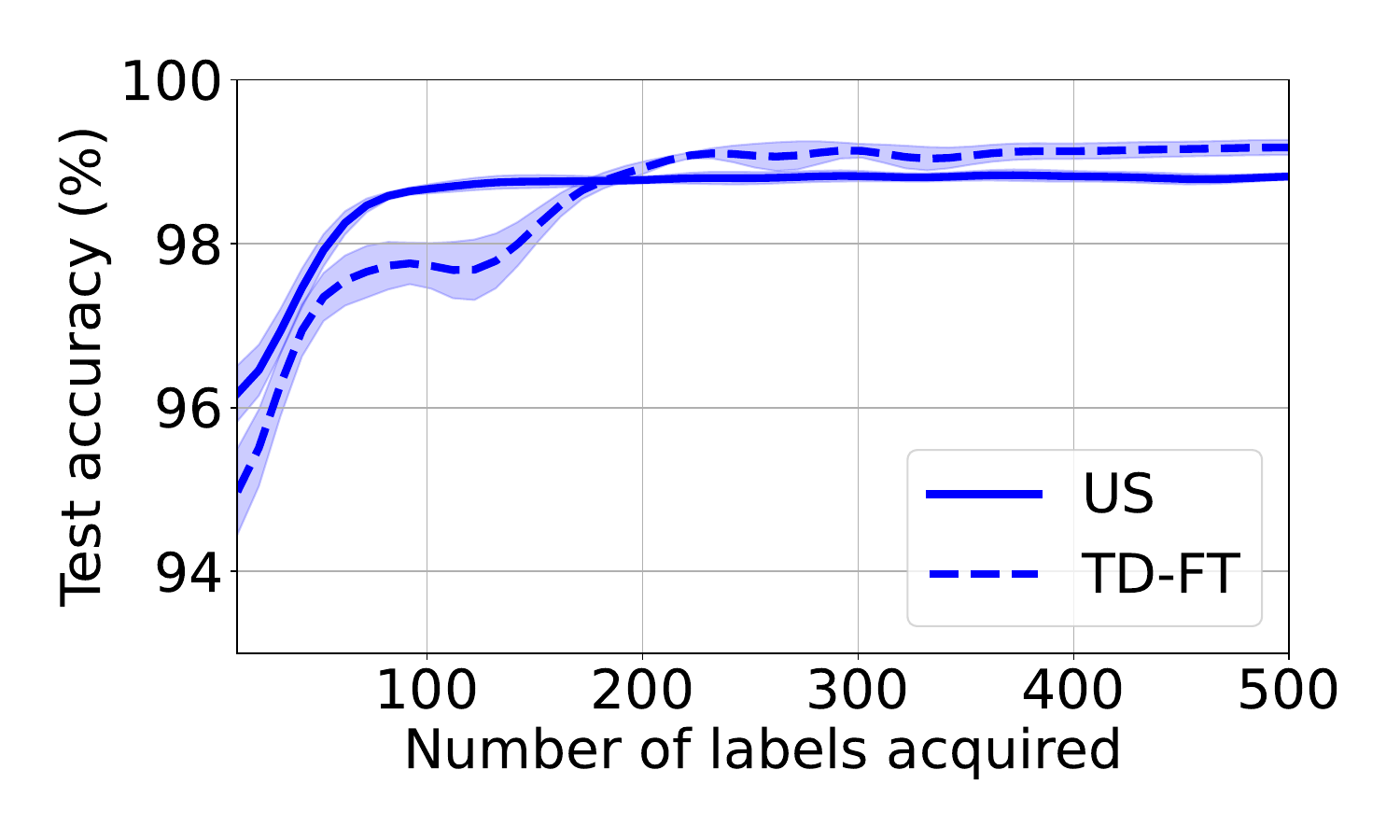}
\end{subfigure}

\caption{Test accuracy on \textbf{F+MNIST} using the \textbf{US} approach and our task--driven approaches using a neural network prediction head. Left shows the results for \textbf{TD-SPLIT} with VAE encoders and right shows the results for \textbf{TD-FT} with SimCLRv2 encoders. Experiments run for 4 seeds. Solid line shows mean and shading $\pm 1$ standard error.
\vspace{-8pt}}
\label{fig:different_PH}
\end{figure*}

From Figure \ref{fig:different_PH}, we see that our approach still outperforms the \textbf{US} approach with a neural network prediction head.

\section{Additional Results}
\label{app:additional_results}

In this section, we provide additional results about the computational cost of our approach, using our task--driven representations for the baselines, and the acquisition counts for different approaches.

\subsection{Task--Driven Representations Improve Baselines}

Table \ref{tab:improving_baselines} shows the results for different baselines, our approach, and now also the baselines using our approach. To integrate the baselines into our approach, we used the same semi--supervised encoder, prediction head and retraining period, varying only the acquisition strategy. As \textbf{SIMILAR} requires gradient embeddings for its acquisition strategy, we replaced the random forest prediction head with a fully--connected neural network with 128 hidden units. Moreover, for simplicity we used the semi--supervised encoder from \textbf{TD-FT}. We use the prefix \textbf{TD} to indicate that the baselines are used with our approach.

We see that integrating the baselines into our approach significantly boosts their performance on all datasets. We note, however, that their performance is still lagging behind our approach with the {EPIG} acquisition strategy.

\begin{table*}[!ht]
\captionsetup{font=footnotesize}
\caption{Final test accuracy of different active learning methods on the \textbf{F+MNIST}, \textbf{CIFAR-10+100} and \textbf{CheXpert} datasets. The prefix \textbf{TD} for \textbf{SIMILAR}, \textbf{GALAXY}, and \textbf{Cluster Margin} is used to indicate that they are used with our approach. We report the mean $\pm1$ standard error over 4 seeds.
}
\vspace{-4pt}
\label{tab:improving_baselines}
\centering
\footnotesize
\begin{tabular}{c|c|c|c} 
\hline
\textbf{Method} & \textbf{F+MNIST} & \textbf{CIFAR-10+100} & \textbf{CheXpert} \\
\hline
\textbf{SIMILAR \citep{SIMILAR}} & $93.82 \pm 0.18$ & $30.87 \pm 1.57$ & $71.94 \pm 0.47$ \\
\textbf{GALAXY \citep{GALAXY}} & $84.74 \pm 0.74$ & $55.28 \pm 0.42$ & $78.76 \pm 0.35$ \\
\textbf{Cluster Margin \citep{ClusterMargin}} & $94.24 \pm 0.17$ & $32.79 \pm 0.45$ & $75.41 \pm 0.29$ \\
\textbf{US+EPIG (SimCLRv2, \cite{semi_epig})} & $98.53 \pm 0.12$ & $76.19 \pm 0.42$ & $77.84 \pm 0.28$ \\
\textbf{US+EPIG (VAE, \cite{semi_epig})} & $94.50 \pm 0.34$ & $30.47 \pm 1.17$ & $66.69 \pm 0.56$ \\
\textbf{US Random (SimCLRv2)} & $92.76 \pm 2.37$ & $74.60 \pm 0.35$ & $74.70 \pm 0.07$ \\
\textbf{US Random (VAE)} & $86.50 \pm 2.24$ & $27.90 \pm 0.87$ & $65.60 \pm 0.18$ \\
\hline
\hline
\textbf{TD-SIMILAR \citep{SIMILAR}} & $98.05 \pm 1.19$ & $79.90 \pm 0.23$ & $77.26 \pm 1.06$ \\
\textbf{TD-GALAXY \citep{GALAXY}} & $99.30 \pm 0.13$ & $73.62 \pm 0.77$ & $78.07 \pm 0.73$ \\
\textbf{TD-Cluster Margin \citep{ClusterMargin}} & $99.32 \pm 0.18$ & $ 70.47\pm 2.58$ & $76.30 \pm 0.38$ \\
\hline
\hline
\textbf{TD-FT Random} & $96.23 \pm 0.37$ & $77.14 \pm 0.42$ & $81.67 \pm 0.40$ \\
\textbf{TD-SPLIT Random} & $88.19 \pm 2.62$ & $54.90 \pm 2.31$ & $75.79 \pm 0.75$ \\
\textbf{TD-SPLIT (Ours)} & $\mathbf{98.46 \pm 0.17}$ & $59.84 \pm 1.25$ & $76.47 \pm 0.27$ \\
\textbf{TD-FT (Ours)} & $\mathbf{99.56 \pm 0.10}$ & $\mathbf{80.90 \pm 0.75}$ & $\mathbf{83.23 \pm 0.38}$ \\
\end{tabular}
\vspace{-8pt}
\end{table*}

\subsection{Acquisition Counts for Different Approaches}
\label{sub:acq_counts_different_approaches}

Table \ref{tab:aq_counts} shows the number of acquisitions that have been made for the target classes at the end of active learning for different approaches. Note that we only have one \textbf{Random} acquisition strategy as this strategy does not depend on the model used.

From Table \ref{tab:aq_counts}, we note two things. First we note that using our approach, whether that is with the baselines or instead of unsupervised representations, improves the count of the target classes by a large margin. This suggests that the gains displayed in Table \ref{tab:improving_baselines} are not merely from using a better model, but also from making better acquisitions. Secondly, we note that the best performing method (\textbf{TD-FT}) does \emph{not} have the largest amount of target classes acquired. This suggests that the precise data point acquired is important, not just its class \citep{yang2023not}.

\begin{table*}[!ht]
\captionsetup{font=footnotesize}
\caption{Number of target classes that have been acquired at the end of active learning for different approaches on the \textbf{F+MNIST} and \textbf{CIFAR-10+100} datasets. We report the mean $\pm1$ standard error over 4 seeds.
}
\vspace{-4pt}
\label{tab:aq_counts}
\centering
\footnotesize
\begin{tabular}{c|c|c} 
\hline
\textbf{Method} & \textbf{F+MNIST} & \textbf{CIFAR-10+100} \\
\hline
\textbf{SIMILAR \citep{SIMILAR}} & $486 \pm 6$ & $920 \pm 237$ \\
\textbf{GALAXY \citep{GALAXY}} & $118 \pm 13$ & $1806 \pm 32$ \\
\textbf{Cluster Margin \citep{ClusterMargin}} & $32 \pm 1$ & $1287 \pm 53$ \\
\textbf{US+EPIG (SimCLRv2, \cite{semi_epig})} & $119 \pm 8$ & $936 \pm 35$  \\
\textbf{US+EPIG (VAE, \cite{semi_epig})} & $118 \pm 4$ & $529 \pm 17$  \\
\hline
\hline
\textbf{TD-SIMILAR \citep{SIMILAR}} & $493 \pm 1$ & $2231 \pm 94$ \\
\textbf{TD-GALAXY \citep{GALAXY}} & $262 \pm 5$ & $1799 \pm 125$  \\
\textbf{TD-Cluster Margin \citep{ClusterMargin}} & $268 \pm 9$ & $ 6503\pm 82$  \\
\hline
\hline
\textbf{Random} & $31 \pm 2$ & $609 \pm 13$  \\
\textbf{TD-SPLIT (Ours)} & $141 \pm 14$ & $1651 \pm 25$  \\
\textbf{TD-FT (Ours)} & $206 \pm 38$ & $2215 \pm 38$ \\
\end{tabular}
\vspace{-8pt}
\end{table*}

\subsection{Computational Cost of TD-SPLIT and TD-FT}

Table \ref{tab:comp_cost} shows the computational cost of running active learning for the \textbf{TD-SPLIT} and \textbf{TD-FT} approaches. We see that, overall, \textbf{TD-FT} is significantly cheaper. This is a result of only being required to train on the labelled data whereas the \textbf{TD-SPLIT} approach requires training on both the unlabelled and labelled data simultaneously. 

We note also that \textbf{TD-SPLIT} has a shorter wall time than \textbf{TD-FT} on the \textbf{F+MNSIT} dataset. This is a result of a using a much lower--dimensional latent space and also more lightweight encoder.

\begin{table*}[!ht]
\captionsetup{font=footnotesize}
\caption{Total wall time in minutes of \textbf{TD-SPLIT} and \textbf{TD-FT} for the \textbf{F+MNIST}, \textbf{CIFAR-10+100} and \textbf{CheXpert} datasets. We report the mean $\pm1$ standard error over 4 seeds.
}
\vspace{-4pt}
\label{tab:comp_cost}
\centering
\footnotesize
\begin{tabular}{c|c|c|c} 
\hline
\textbf{Method} & \textbf{F+MNIST} & \textbf{CIFAR-10+100} & \textbf{CheXpert} \\
\hline
\textbf{TD-SPLIT} & $43 \pm 10$ & $610 \pm 32$ & $532 \pm 47$ \\
\textbf{TD-FT} & $57 \pm 5$ & $310 \pm 21$ & $50 \pm 8$ \\
\end{tabular}
\vspace{-8pt}
\end{table*}

\subsection{Final Test Accuracies for Figures \ref{fig:fmnist}, \ref{fig:chexpert}}

Table \ref{tab:final_acc_fig_3_4} shows the final test accuracies for Figure \ref{fig:fmnist}, \ref{fig:chexpert}. Again, we see that using our approach improves unsupervised representations across both datasets and all acquisition strategies. In particular, we note that {EPIG} still performs best across all the acquisition strategies, owing to its prediction--oriented nature.

\begin{table*}[!ht]
\caption{Final test accuracies for our \textbf{TD-SPLIT} and \textbf{TD-FT} approaches and the \textbf{US} approach on \textbf{F+MNIST} and \textbf{CheXpert} for three different acquisition strategies. We report the mean $\pm1$ standard error over 4 seeds.}
\vspace{-4pt}
\label{tab:final_acc_fig_3_4}
\centering
\footnotesize
\begin{tabular}{l|l|c|c}
\toprule
\textbf{Method} & \textbf{Strategy} & \textbf{F+MNIST} & \textbf{CheXpert} \\
\midrule

\multirow{3}{*}{\makecell{\textbf{TD-SPLIT}}}
& \textbf{EPIG} & $\mathbf{98.46 \pm 0.17}$  & $\mathbf{76.47 \pm 0.27}$ \\
& \textbf{BALD} & $98.30 \pm 0.18$ & $75.65 \pm 0.43$  \\
& \textbf{CS}   & $98.43 \pm 0.04$  & $76.11 \pm 0.21$ \\
\midrule

\multirow{3}{*}{\makecell{\textbf{~~~US}\\\textbf{~~~(VAE)}}}
& \textbf{EPIG} & $94.50 \pm 0.34$ & $66.69 \pm 0.56$ \\
& \textbf{BALD} & $97.30 \pm 0.21$ & $67.56 \pm 0.24$ \\
& \textbf{CS}   & $97.59 \pm 0.18$ & $69.91 \pm 0.30$ \\
\midrule
\midrule

\multirow{3}{*}{\makecell{\textbf{~~TD-FT}}}
& \textbf{EPIG} & $\mathbf{99.56 \pm 0.10}$ & $\mathbf{83.23 \pm 0.38}$ \\
& \textbf{BALD} & $99.32 \pm 0.07$ & $83.10 \pm 0.37$ \\
& \textbf{CS}   & $99.41 \pm 0.10$ & $82.79 \pm 0.11$ \\
\midrule

\multirow{3}{*}{\makecell{\textbf{US}\\\textbf{(SimCLRv2)}}}
& \textbf{EPIG} & $98.53 \pm 0.12$ & $77.84 \pm 0.28$ \\
& \textbf{BALD} & $98.13 \pm 0.13$ & $76.96 \pm 0.16$ \\
& \textbf{CS}   & $98.36 \pm 0.11$ & $77.71 \pm 0.06$ \\
\end{tabular}
\end{table*}
\vfill

\end{document}